\documentclass{article}

\usepackage[main, final]{neurips_2026}

\usepackage{titletoc}
\usepackage{pifont}

\usepackage[utf8]{inputenc} 
\usepackage[T1]{fontenc}    
\usepackage[dvipsnames]{xcolor}
\definecolor{MyPurple}{HTML}{6B67EE}
\usepackage[colorlinks=true, linkcolor=MyPurple, citecolor=MyPurple, urlcolor=MyPurple]{hyperref}       
\usepackage{url}            
\usepackage{booktabs}       
\usepackage{amsfonts}       
\usepackage{nicefrac}       
\usepackage{microtype}      
\usepackage{xcolor}         
\usepackage{pagecolor}
\usepackage{amsmath}
\usepackage{amssymb}
\usepackage{graphicx}
\usepackage{subcaption}
\usepackage{colortbl}
\usepackage{algorithm}
\usepackage{subcaption}
\usepackage{tabularx}
\usepackage{amsmath,amssymb,amsthm,bbm}
\usepackage{multirow}
\usepackage{adjustbox}
\usepackage{algorithm}
\usepackage{algpseudocode}
\usepackage{amsmath}
\usepackage[noEnd=false]{algpseudocodex}
\usepackage[toc,page]{appendix} 
\usepackage{tocloft} 
\usepackage{titlesec}
\usepackage{booktabs}
\usepackage{threeparttable}
\usepackage{enumitem}
\setlist[enumerate]{left=10pt}
\usepackage[font=small]{caption}
\usepackage{wrapfig}
\usepackage{listings}
\usepackage{booktabs}
\usepackage{mdframed}
\usepackage{thm-restate}

\usepackage{thmtools}
\usepackage[most]{tcolorbox}
\tcbuselibrary{theorems}

\tcolorboxenvironment{proposition}{
  colback=gray!10,
  colframe=gray!10,
  boxrule=0pt,
  arc=0pt,
  left=6pt,
  right=6pt,
  top=6pt,
  bottom=6pt,
  enhanced,
}

\tcolorboxenvironment{remark}{
  colback=gray!10,
  colframe=gray!10,
  boxrule=0pt,
  arc=0pt,
  left=6pt,
  right=6pt,
  top=6pt,
  bottom=6pt,
  enhanced,
}

\tcolorboxenvironment{definition}{
  colback=gray!10,
  colframe=gray!10,
  boxrule=0pt,
  arc=0pt,
  left=6pt,
  right=6pt,
  top=6pt,
  bottom=6pt,
  enhanced,
}

\tcolorboxenvironment{box}{
  colback=gray!10,
  colframe=gray!10,
  boxrule=0pt,
  arc=0pt,
  left=6pt,
  right=6pt,
  top=6pt,
  bottom=6pt,
  enhanced,
}

\tcolorboxenvironment{lemma}{
  colback=gray!10,
  colframe=gray!10,
  boxrule=0pt,
  arc=0pt,
  left=6pt,
  right=6pt,
  top=6pt,
  bottom=6pt,
  enhanced,
}

\tcolorboxenvironment{restatable}{
  colback=gray!10,
  colframe=gray!10,
  boxrule=0pt,
  arc=0pt,
  left=6pt,
  right=6pt,
  top=6pt,
  bottom=6pt,
  enhanced,
}

\tcolorboxenvironment{theorem}{
  colback=gray!10,
  colframe=gray!10,
  boxrule=0pt,
  arc=0pt,
  left=6pt,
  right=6pt,
  top=6pt,
  bottom=6pt,
  enhanced,
}

\declaretheorem[
  name=Definition,
  numberwithin=section,
  refname={Definition,Definition},
  Refname={Definition,Definition}
]{definition}

\declaretheorem[
  name=Proposition,
  numberwithin=section,
  refname={Proposition,Propositions},
  Refname={Proposition,Propositions}
]{proposition}

\declaretheorem[
  name=Lemma,
  numberwithin=section,
  refname={Lemma,Lemmas},
  Refname={Lemma,Lemmas}
]{lemma}

\declaretheorem[
  name=Theorem,
  numberwithin=section,
  refname={Theorem,Theorems},
  Refname={Theorem,Theorems}
]{theorem}

\declaretheorem[
  name=Remark,
  numberwithin=section,
  refname={Remark,Remarks},
  Refname={Remark,Remarks}
]{remark}

\definecolor{customblue}{HTML}{4A90E2}

\title{Entangled Schrödinger Bridge Matching}

\author{
  Sophia Tang$^{1}$, Yinuo Zhang$^{2}$, Pranam Chatterjee$^{1, 2}$ \\\\
  $^{1}$Department of Computer and Information Science, University of Pennsylvania \\
  $^{2}$Department of Bioengineering, University of Pennsylvania \\
  \\
  Correspondence to: \href{mailto:sophtang@engineering.upenn.edu}{\texttt{sophtang@engineering.upenn.edu}}, 
  \href{mailto:pranam@upenn.edu}{\texttt{pranam@upenn.edu}}
}

\begin{document}

\maketitle

\begin{abstract}
Simulating trajectories of multi-particle systems on complex energy landscapes is a central task in molecular dynamics (MD) and drug discovery, but remains challenging at scale due to computationally expensive and long simulations. Previous approaches leverage techniques such as flow or Schrödinger bridge matching to implicitly learn joint trajectories from data snapshots. However, many systems, including biomolecular systems and heterogeneous cell populations, undergo \textit{dynamic} interactions that evolve over their trajectory and cannot be captured through static snapshots. To close this gap, we introduce \textbf{Entangled Schrödinger Bridge Matching (EntangledSBM)}, a framework that learns the first- and second-order stochastic dynamics of interacting, multi-particle systems where the direction and magnitude of each particle's path depend dynamically on the paths of the other particles. We define the Entangled Schrödinger Bridge (EntangledSB) problem as solving a coupled system of bias forces that \emph{entangle} particle velocities. We show that our framework accurately simulates heterogeneous cell populations under perturbations and rare transitions in high-dimensional biomolecular systems.
\end{abstract}

\section{Introduction}
Schrödinger bridge matching (SBM) has enabled progress in problems ranging from molecular dynamics simulation \citep{holdijk2023stochastic} to cell state modeling \citep{tong24a} by learning dynamic trajectories between initial and target distributions, with extensions for state-cost minimization \citep{generalized-sb} and branching dynamics \citep{tang2025branched}. Most of these frameworks treat particles as independent or capture interactions only implicitly through static snapshots, and existing approaches for interacting particle systems rely on a mean-field assumption in which particles are exchangeable and interact only through averaged effects \citep{Liu2022, zhang2025modeling}. This assumption fails for heterogeneous dynamics such as those of conformationally dynamic proteins \citep{vincoff2025fuson, chen2023generative}, heterogeneous cell populations \citep{PiersonSmela2023}, or interacting token sequences \citep{geshkovski2023emergence}, where each particle's motion depends on the evolving configuration of the others.

Modeling such dependencies requires a coupled distribution over the joint $n$-particle phase space $\boldsymbol{X}_t = (\boldsymbol{R}_t, \boldsymbol{V}_t) \in \mathcal{X}$, transporting $\pi_\mathcal{A}(\boldsymbol{X}_0)$ to $\pi_\mathcal{B}(\boldsymbol{X}_T)$, and learning second-order dynamics in which interactions between velocity fields evolve over time -- a regime that remains largely unexplored. We introduce \textbf{Entangled Schrödinger Bridge Matching (EntangledSBM)}, a framework that learns interacting second-order dynamics of $n$-particle systems, capturing dependencies on both positions and velocities at each step.
\begin{figure*}
    \centering
    \includegraphics[width=\linewidth]{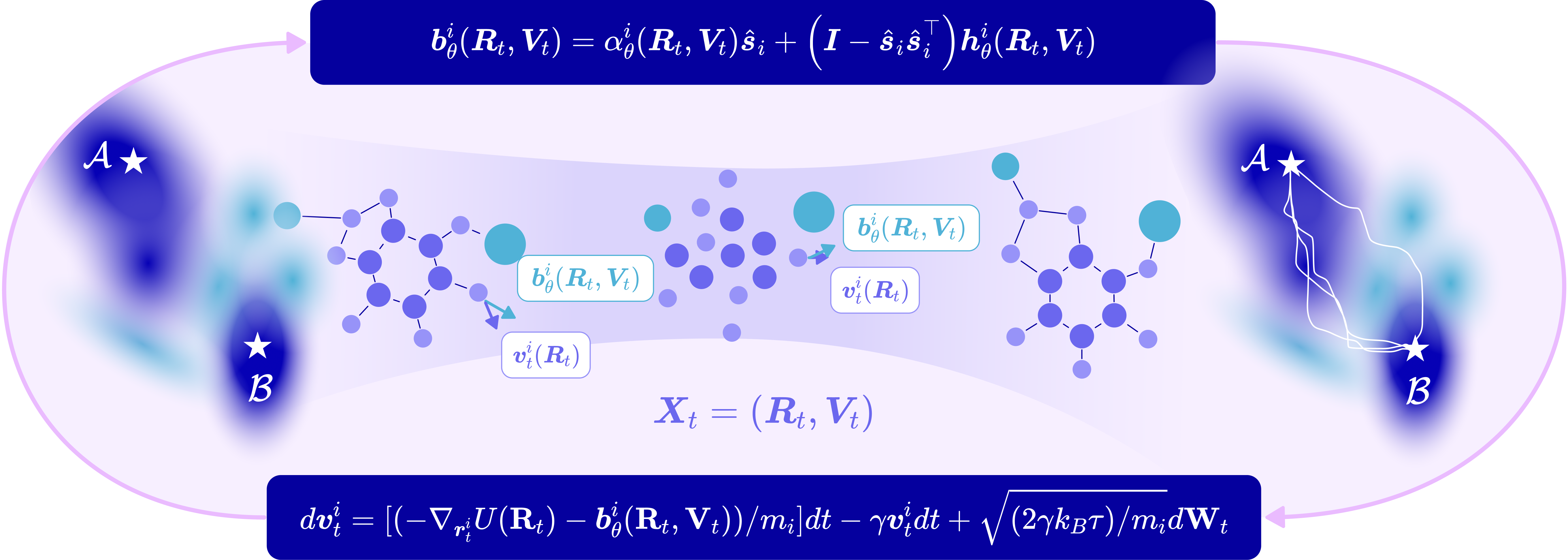}
    \caption{\textbf{Entangled Schrödinger Bridge Matching.} We consider the problem of simulating interacting multi-particle systems, where each particle's velocity depends dynamically on the velocities of the other particles in the system, and introduce \textbf{EntangledSBM}, a framework that parameterizes an entangled bias force capturing the dynamic interactions between particles.}
    \label{fig:EntangledSBM}
\end{figure*}

Our \textbf{main contributions} can be summarized in the following points: 
\begin{enumerate}
    \item  \textbf{Problem Formulation:} We formulate the Entangled Schrödinger Bridge (EntangledSB) problem over the joint phase space of positions and velocities under second-order Langevin dynamics, and reduce it to a stochastic optimal control (SOC) problem with a non-factorizable control that assumes no exchangeable, mean-field interactions (Sec~\ref{sec:Problem Formulation}). 
    \item \textbf{Entangled Schrödinger Bridge Matching:} We propose an importance-weighted cross-entropy objective that is convex in the controlled path measure, fully off-policy, and defined in discrete time, so that a replay buffer can reuse each simulated rollout over many gradient steps (Sec~\ref{sec:Cross-Entropy Objective}). 
    \item \textbf{Target-Conditioned Parameterization:} We introduce a target-conditioned bias force with a non-negative component along the target direction and an unconstrained orthogonal correction. This makes sparse-reward transition path sampling trainable and allows conditioning on unseen targets at inference time (Sec~\ref{sec:First- and Second-Order}).
\end{enumerate}
We evaluate EntangledSBM on drug-perturbed cell dynamics (Sec~\ref{experiment:Cell}), pancreatic $\beta$-cell differentiation (Sec~\ref{experiment:Beta}), and all-atom transition path sampling (Sec~\ref{experiment:TPS}), demonstrating stronger performance across baselines while modeling strictly more expressive dynamics. 

\section{Preliminaries}
\label{sec:Preliminaries}
\paragraph{Langevin Dynamics}
A time-evolving $n$-particle molecular system can be represented as $\boldsymbol{X}_t=(\boldsymbol{R}_t, \boldsymbol{V}_t)$, where $\boldsymbol{R}_t=\{\boldsymbol{r}^i_t\in \mathbb{R}^{d}\}_{i=1}^n$ denotes the set of coordinates and $\boldsymbol{V}_t=\{\boldsymbol{v}^i_t\in \mathbb{R}^{d}\}_{i=1}^n$ denotes the set of velocities of each particle $i$. The evolution of the positions and velocities of the system, given a potential energy function $U:\mathcal{X}\to \mathbb{R}$, can be modeled with \textit{Langevin dynamics}, which effectively captures the motion of particles under \textbf{conservative forces} between particles and \textbf{stochastic collisions} with the surrounding environment \citep{bussi2007accurate, yang2006effective} using a pair of stochastic differential equations (SDEs) defined as
\begin{small}
\begin{align}
\begin{cases}
    d\boldsymbol{r}^i_t&=\boldsymbol{v}^i_tdt\\
    d\boldsymbol{v}^i_t&=\frac{-\nabla_{\boldsymbol{r}^i_t}U(\boldsymbol{R}_t)}{m_i}dt-\gamma \boldsymbol{v}^i_tdt+\sqrt{\frac{2\gamma k_B\tau}{m_i}}d\boldsymbol{W}^i_t
\end{cases}\label{eq:langevin-dynamics}
\end{align}
\end{small}

where $m_i$ is the mass of each particle, $\gamma$ is the friction coefficient, $k_B$ is the Boltzmann constant, $\tau$ is the temperature, and $d\boldsymbol{W}_t$ is standard Brownian motion. In molecular dynamics (MD) simulations of biomolecules, the particles undergo \textit{underdamped} Langevin dynamics with small $\gamma$, where \textit{inertia} is not negligible.

Many biological systems, including cell clusters for cell-state trajectory simulation, can be modeled with \textit{overdamped} Langevin dynamics, where inertia is negligible but the system still undergoes external forces that define its motion. This can be represented with the first-order SDE given by
\begin{small}
\begin{align}
    d\boldsymbol{r}^i_t=\frac{-\nabla_{\boldsymbol{r}^i_t}U(\boldsymbol{R}_t)}{\gamma}dt+\sqrt{\frac{2 k_B\tau}{\gamma}}d\boldsymbol{W}_t^i\label{eq:langevin-dynamics-2}
\end{align}
\end{small}

\paragraph{Schrödinger Bridge Matching}
Tasks such as simulating cell state trajectories and transition path sampling aim to simulate the Langevin dynamics from an initial distribution to a desired target state or distribution. Given an initial distribution $\pi_\mathcal{A}$ and a target distribution $\pi_\mathcal{B}$, we define the distribution of paths $\boldsymbol{X}_{0:T}:=(\boldsymbol{X}_t)_{t\in [0,T]}$ satisfying the endpoint constraints $\boldsymbol{R}_0\sim \pi_\mathcal{A}$ and $\boldsymbol{R}_T\sim \pi_\mathcal{B}$ as the \textit{optimal bridge distribution} $\mathbb{P}^\star(\boldsymbol{X}_{0:T})$ defined as
\begin{align}
    \mathbb{P}^\star(\boldsymbol{X}_{0:T})&=\frac{1}{Z(\boldsymbol{X}_0)} \mathbb{P}^0(\boldsymbol{X}_{0:T})\pi_\mathcal{B}(\boldsymbol{R}_T), \quad Z(\boldsymbol{X}_0)=\mathbb{E}_{\mathbb{P}^0}\left[\pi_\mathcal{B}(\boldsymbol{R}_T)\mid \boldsymbol{X}_0\right]
\end{align}

where $\mathbb{P}^0(\boldsymbol{X}_{0:T})$ is the base path distribution generated from the SDEs (\ref{eq:langevin-dynamics}) or (\ref{eq:langevin-dynamics-2}) and $Z$ is the normalizing constant (see App~\ref{app:sb_soc_connection} for how this soft-terminal bridge relates to the standard two-marginal SB problem). Schrödinger Bridge Matching (SBM) aims to parameterize a \textbf{control} or \textbf{bias force} $\boldsymbol{b}_\theta$ that tilts the path distribution so that $\mathbb{P}^{b_\theta}(\boldsymbol{X}_{0:T})$ minimizes the KL-divergence from the bridge path distribution $\mathbb{P}^\star$ given by
\begin{small}
\begin{align}
    &\boldsymbol{b}^\star=\arg\min_{\boldsymbol{b}_\theta}D_{\text{KL}}\left(\mathbb{P}^{b_\theta}\|\mathbb{P}^\star\right)\quad \text{s.t.}\;\;\begin{cases}
        d\boldsymbol{r}^i_t=\boldsymbol{v}^i_tdt\\
        d\boldsymbol{v}^i_t=\frac{-\nabla_{\boldsymbol{r}^i_t}U(\boldsymbol{R}_t)+\boldsymbol{b}^i_\theta(\boldsymbol{R}_t)}{m_i}dt-\gamma \boldsymbol{v}^i_tdt+\sqrt{\frac{2\gamma k_B\tau}{m_i}}d\boldsymbol{W}^i_t
    \end{cases}\label{eq:control-langevin}
\end{align}
\end{small}
In the case of transition path sampling (TPS) where there is a single initial state $\boldsymbol{R}_\mathcal{A}$ and target state $\boldsymbol{R}_\mathcal{B}$, we define the target distribution as the relaxed indicator function $\pi_\mathcal{B}(\boldsymbol{R}_T)=\boldsymbol{1}_{\mathcal{B}}(\boldsymbol{R}_T)$ centered around $\boldsymbol{R}_\mathcal{B}$.

\section{Learning Interacting Multi-Particle Dynamics}
\label{sec:Problem Formulation}
The \textbf{key challenge} with simulating the dynamics of multi-particle systems lies in the question: \textit{how can we simulate dynamic trajectories from static snapshot data?} The emergence of flow and Schrödinger bridge matching frameworks has effectively approached this problem by defining a \textbf{parameterized velocity field} that learns feasible trajectories from data snapshots. However, current strategies remain limited in their ability to model \textit{interacting} multi-particle systems where the velocities carry inherent dependencies that cannot be captured via static snapshots of the system. Prior attempts to capture interactions between particles rely on the \textit{mean-field assumption} \citep{generalized-sb, Liu2022, zhang2025modeling}, where each particle interacts with the others only through their averaged effect, which \textbf{does not hold for heterogeneous biomolecular systems} \citep{illarionov2023combining}.

To address this challenge, we propose a framework that introduces an additional degree of freedom through a \textbf{bias force} that implicitly captures the dependencies between \textit{both} the static positions and dynamic velocities of each particle in the system to control the joint trajectories. We leverage a Transformer architecture and treat each particle as an individual token with features that attend to the features of all other tokens. Crucially, our approach requires no handcrafted features, making it scalable to high-complexity systems. 

In Sec \ref{sec:EntangledSB Problem}, we formalize this problem as the \textbf{Entangled Schrödinger Bridge (EntangledSB)} problem, which aims to determine the optimal set of trajectories to a target state while capturing the dynamic interactions between particles. We then provide a tractable approach to solve the EntangledSB problem with stochastic optimal control (SOC) theory in Sec \ref{sec:Solving EntangledSB with SOC}. We illustrate the high-level framework of learning \textbf{entangled} bias forces in Alg \ref{alg:Framework} and detail our specific implementation and parameterization in Sec \ref{sec:EntangledSBM}.

\subsection{Entangled Schrödinger Bridge Problem}
\label{sec:EntangledSB Problem}
Here, we formalize the \textbf{Entangled Schrödinger Bridge (EntangledSB) problem}, which aims to find a set of optimal bias forces for each particle in the system that depends \textit{dynamically} on the positions and velocities across the entire multi-particle system to guide the Schrödinger bridge trajectory to a target distribution. Specifically, we consider an $n$-particle system $\boldsymbol{X}_t=(\boldsymbol{R}_t, \boldsymbol{V}_t)$ with $\boldsymbol{R}_t=\{\boldsymbol{r}^i_t\in \mathbb{R}^d\}_{i=1}^n$ and $\boldsymbol{V}_t=\{\boldsymbol{v}^i_t\in \mathbb{R}^d\}_{i=1}^n$, where each particle evolves over the time horizon $t\in [0, T]$ via the \textbf{bias-controlled} SDE given by
\begin{small}
\begin{align}
    d\boldsymbol{r}^i_t=\frac{1}{\gamma}\left(-\nabla_{\boldsymbol{r}^i_t}U(\boldsymbol{R}_t)+\boldsymbol{b}^i(\boldsymbol{R}_t, \boldsymbol{V}_t)\right)dt+\sqrt{\frac{2 k_B\tau}{\gamma}}d\boldsymbol{W}^i_t
\end{align}
\end{small}

The velocities follow the potential energy landscape defined over the joint coordinates of the system $U(\boldsymbol{R}_t)$ with an \textit{entangled} bias force $\boldsymbol{b}^i(\boldsymbol{R}_t, \boldsymbol{V}_t)$ that depends dynamically on the position and velocity of \textit{all} particles in the system. We can further extend this framework to \textbf{second-order} Langevin dynamics via the pair of SDEs

\begin{align}
    d\boldsymbol{r}_t^i&=\boldsymbol{v}_t^idt, \quad d\boldsymbol{v}^i_t=\frac{-\nabla_{\boldsymbol{r}_t ^i}U(\boldsymbol{R}_t)+\boldsymbol{b}^i(\boldsymbol{R}_t, \boldsymbol{V}_t)}{m_i}dt-\gamma \boldsymbol{v}_t^idt+\sqrt{\frac{2\gamma k_B\tau}{m_i}}d\boldsymbol{W}^i_t\label{eq:multi-langevin}
\end{align}

Formally, we seek the optimal $\boldsymbol{b}^\star(\boldsymbol{R}_t, \boldsymbol{V}_t):=\{\boldsymbol{b}^i(\boldsymbol{R}_t, \boldsymbol{V}_t)\}_{i=1}^n$ that solves the EntangledSB problem defined below. 

\begin{definition}[Entangled Schrödinger Bridge Problem]\label{def:EntangledSB}
    Given an initial distribution $\pi_\mathcal{A}$, a target distribution $\pi_\mathcal{B}$, the EntangledSB problem aims to determine the set of optimal bias forces $\boldsymbol{b}^\star:=\{\boldsymbol{b}^i(\boldsymbol{R}_t, \boldsymbol{V}_t)\}_{i=1}^n$ for each particle $i$ in the system that solves the optimization problem
    \begin{small}
    \begin{align}
        &\boldsymbol{b}^\star=\arg\min_{\boldsymbol{b}_\theta}D_\text{KL}\left(\mathbb{P}^{b_\theta}\|\mathbb{P}^\star\right)\;\text{s.t.}\;
        \begin{cases}
            \mathbb{P}^\star=\frac{1}{Z}\mathbb{P}^0\pi_\mathcal{B}(\boldsymbol{R}_T)\\
            \mathbb{P}^\star_0=\pi_\mathcal{A}(\boldsymbol{X}_0)
        \end{cases}\nonumber
    \end{align}
    \end{small}
    where $\mathbb{P}^0$ is the base path measure with joint dynamics that follow the SDEs in (\ref{eq:multi-langevin}).
\end{definition}

\subsection{Solving EntangledSB with Stochastic Optimal Control}
\label{sec:Solving EntangledSB with SOC}
To solve the EntangledSB problem, we leverage stochastic optimal control (SOC) theory (see App ~\ref{app:sb_soc_connection} for the formal connection between Schrödinger bridges and SOC), where we aim to find the set of optimal control drifts $\boldsymbol{b}^\star(\boldsymbol{R}_t, \boldsymbol{V}_t)$ given a target distribution $\pi_\mathcal{B}$.

\begin{restatable}[Solving EntangledSB with Stochastic Optimal Control]{proposition}{entangledsoc}\label{prop:EntangledSOC}
    We can solve the EntangledSB problem with the stochastic optimal control (SOC) objective given by
    \begin{small}
    \begin{align}
        &\boldsymbol{b}^\star=\arg\min_{\boldsymbol{b}_\theta}\mathbb{E}_{\boldsymbol{X}_{0:T}\sim \mathbb{P}^{b_\theta}}\left[\int_0^T\frac{1}{2}\|\boldsymbol{b}_\theta\|^2dt-r(\boldsymbol{X}_T)\right]\nonumber
    \end{align}
    \end{small}
    such that (\ref{eq:multi-langevin}) where $\boldsymbol{b}_\theta\equiv \boldsymbol{b}_\theta(\boldsymbol{R}_t, \boldsymbol{V}_t)$ and $r(\boldsymbol{X}_T):=\log \pi_\mathcal{B}(\boldsymbol{R}_T)$ is the terminal reward that measures the log-probability under the target distribution.
\end{restatable}
The proof is provided in App \ref{appendix-prop:EntangledSOC}. The SOC problem is optimized over \textit{unconstrained} trajectories. The terminal reward acts as a soft penalty rather than a hard constraint, and the terminal distribution of $\mathbb{P}^{b_\theta}$ matches that of $\mathbb{P}^\star$ only at the optimum. In Sec \ref{sec:Cross-Entropy Objective}, we introduce an importance-weighted cross-entropy objective that shares this optimum and can be optimized off-policy.

\begin{algorithm}[t]
\caption{Framework for \textbf{Entangled Schrödinger Bridge Matching}}\label{alg:Framework}
    \begin{algorithmic}[1]
        \State \textbf{Input:} Parameterized bias force network $\boldsymbol{b}_\theta(\boldsymbol{R}_t, \boldsymbol{V}_t):\mathbb{R}^{n\times d}\times \mathbb{R}^{n\times d}\to \mathbb{R}^{n\times d}$, initial and target joint distributions of the system $\pi_\mathcal{A},\pi_\mathcal{B}$, energy function $U(\boldsymbol{R}_t): \mathbb{R}^{n\times d}\to \mathbb{R}$, objective function in path space $\mathcal{L}(\mathbb{P}^\star,\mathbb{P}^{b_\theta})$ where $\mathbb{P}^\star=\arg\min_{\mathbb{P}^{\boldsymbol{b}_\theta}} \mathcal{L}(\mathbb{P}^\star,\mathbb{P}^{b_\theta})$
        \While{not converged}
            \State \texttt{Sample} initial positions $\boldsymbol{R}_0\sim \pi_\mathcal{A}$
            \For{$t=0$ to $T-1$}
            \State \texttt{Evaluate} unconditional velocities $\boldsymbol{V}_t(\boldsymbol{R}_t)$
            \State \texttt{Compute} entangled bias force $\boldsymbol{b}_\theta(\boldsymbol{R}_t, \boldsymbol{V}_t)$
            \State \texttt{Simulate} step over a discrete time step $\Delta t$ to get the updated state $(\boldsymbol{R}_{t+1}, \boldsymbol{V}_{t+1})$
            \EndFor
            \State \texttt{Compute} reward on terminal states $r(\boldsymbol{X}_T)$
            \State \texttt{Evaluate} path objective $\mathcal{L}(\mathbb{P}^\star, \mathbb{P}^{b_\theta})$ on simulated paths
            \State \texttt{Update} $\theta$ using the gradient $\nabla_\theta\mathcal{L}(\mathbb{P}^\star, \mathbb{P}^{b_\theta})$
        \EndWhile
    \end{algorithmic}
\end{algorithm}

\section{Entangled Schrödinger Bridge Matching}
\label{sec:EntangledSBM}
In this section, we introduce \textbf{Entangled Schrödinger Bridge Matching} (EntangledSBM), a novel framework for learning to simulate trajectories of $n$-particle systems with a \textbf{bias force that dynamically depends on the positions and velocities of the other particles} in the system and can generalize to \textit{unseen} target distributions without further training. Our parameterization ensures that a bias-only update does not increase the distance to some target state for small enough $\Delta t$, and its orthogonal component remains unconstrained (Sec \ref{sec:Sampling Direction}). We introduce a \textbf{weighted cross-entropy} objective that enables efficient \textit{off-policy learning} with a replay buffer of simulated trajectories (Sec \ref{sec:Cross-Entropy Objective}). 

\subsection{Parameterizing the Entangled Bias Force}
\label{sec:First- and Second-Order}
\paragraph{Bias Force Parameterization}
\label{sec:Sampling Direction}
To ensure that the bias force does not increase the distance from the target distribution $\pi_\mathcal{B}$, we ensure that the projection of the predicted bias force onto the direction of the gradient of the target distribution is non-negative for all particles $i$ in the system
\begin{align}
    \langle\boldsymbol{b}_\theta^i(\boldsymbol{R}_t, \boldsymbol{V}_t), \nabla_{\boldsymbol{r}_t^i}\log \pi_\mathcal{B}\rangle\geq 0\label{eq:positivity}
\end{align}
Let $\boldsymbol{s}_i=\nabla_{\boldsymbol{r}_t^i}\log \pi_\mathcal{B}\in \mathbb{R}^d$ denote the direction towards the target state. To ensure that the bias force is within the cone surrounding $\boldsymbol{s}_i$, we use the following parameterization
\begin{small}
\begin{align}
    \boldsymbol{b}_\theta^i(\boldsymbol{R}_t,\boldsymbol{V}_t)&=\underbrace{\alpha^i_\theta(\boldsymbol{R}_t, \boldsymbol{V}_t)\hat{\boldsymbol{s}}_i}_{\text{parallel component}}+\underbrace{\left(\boldsymbol{I}-\hat{\boldsymbol{s}}_i \hat{\boldsymbol{s}}_i^\top\right)\boldsymbol{h}^i_\theta(\boldsymbol{R}_t, \boldsymbol{V}_t)}_{\text{orthogonal component}}\label{eq:bias-force}
\end{align}
\end{small}
where $\hat{\boldsymbol{s}}_i=\boldsymbol{s}_i/\|\boldsymbol{s}_i\|$ and $\alpha^i_\theta(\boldsymbol{R}_t, \boldsymbol{V}_t):=\texttt{softplus}(a^i_\theta(\boldsymbol{R}_t, \boldsymbol{V}_t))\geq 0$, with raw network output $a^i_\theta$, is a scaling factor for the unit vector $\hat{\boldsymbol{s}}_i$ and $\boldsymbol{h}^i_\theta(\boldsymbol{R}_t, \boldsymbol{V}_t)\in \mathbb{R}^d$ is the per-atom correction vector that is projected onto the plane orthogonal to $\hat{\boldsymbol{s}}_i$, both parameterized with neural networks $\theta$. Since the orthogonal component does not affect the dot product with $\hat{\boldsymbol{s}}_i$, the non-negativity constraint $\langle\boldsymbol{b}_\theta^i(\boldsymbol{R}_t, \boldsymbol{V}_t), \nabla_{\boldsymbol{r}_t^i}\log \pi_\mathcal{B}\rangle\geq 0$ in (\ref{eq:positivity}) is guaranteed by the first term. Intuitively, the orthogonal component enables the bias force to have greater flexibility in moving sideways (e.g., to avoid infeasible regions or add rotational/collective effects) while the distance from some target state stays non-increasing under the bias-only update in Proposition~\ref{prop:Non-Increasing Distance}.

\begin{restatable}[Non-Increasing Distance from Target Distribution]{proposition}{nonincreasing}\label{prop:Non-Increasing Distance}
    For small enough $\Delta t$, the distance from some target state $\boldsymbol{R}_\mathcal{B}\in\mathrm{supp}(\pi_{\mathcal{B}})$ after an update with only the bias force $\boldsymbol{b}_\theta^i(\boldsymbol{R}_t,\boldsymbol{V}_t)$ defined in (\ref{eq:bias-force}) is non-increasing, such that
    \begin{align}
        \exists \boldsymbol{R}_\mathcal{B}\in\mathrm{supp}(\pi_\mathcal{B})\;\;\text{s.t.}\;\;\|\boldsymbol{R}_{t+\Delta t}-\boldsymbol{R}_\mathcal{B}\|\leq \|\boldsymbol{R}_{t}-\boldsymbol{R}_\mathcal{B}\|\nonumber
    \end{align}
    where $\boldsymbol{R}_{t+\Delta t}$ collects the per-particle updates $\boldsymbol{r}^i_{t+\Delta t}=\boldsymbol{r}^i_{t}+(\boldsymbol{b}_\theta^i(\boldsymbol{R}_t, \boldsymbol{V}_t)/m_i)\Delta t$, which omit the potential, friction, and noise terms. 
\end{restatable}

The proof is provided in App \ref{appendix-prop:Non-Increasing Distance}. In contrast to previous works that constrain the bias force to point strictly in the direction of a fixed target position, our approach allows greater flexibility in the direction of the orthogonal component. 

\paragraph{Model Architecture}
To integrate dependencies on the positions and velocities across $n$ particles, we leverage a Transformer-based architecture where each particle has input features \texttt{Cat}$(\boldsymbol{r}^i_t, \boldsymbol{v}^i_t)\in \mathbb{R}^{2d}$ and the full system features are \texttt{Cat}$(\boldsymbol{R}_t, \boldsymbol{V}_t)\in \mathbb{R}^{n\times 2d}$, where \texttt{Cat} denotes concatenation. Each particle additionally receives its target-relative offset $\boldsymbol{r}^i_\mathcal{B}-\boldsymbol{r}^i_t$ and distance $\|\boldsymbol{r}^i_\mathcal{B}-\boldsymbol{r}^i_t\|$. Attention couples these features across all $n$ particles, so the bias force is conditioned on the full target-relative configuration of the system, acting as a goal-conditioned correction rather than a fixed directional field. This is what allows conditioning on unseen targets at inference. The Transformer encoder enables efficient and expressive propagation of feature information across all pairs of particles in the system to generate context-aware embeddings for bias force parameterization. For MD systems, we ensure invariance of the coordinate frame using the Kabsch algorithm \citep{kabsch1976solution}, which aligns the position $\boldsymbol{R}_t$ with the target position $\boldsymbol{R}_\mathcal{B}$ before input into the model. Further details are provided in App \ref{app:cell-experiment} and \ref{app:tps-experiment}.

\subsection{Off-Policy Learning with Weighted Cross-Entropy}
\label{sec:Cross-Entropy Objective}
\paragraph{Log-Variance Objective}
To train the bias force to match the optimal $\boldsymbol{b}^\star$, we can adapt the \textit{log-variance} (LV) divergence \citep{seong2025transition, nusken2021solving} defined as
\begin{small}
\begin{align}
    \mathcal{L}_{\text{LV}}&(\mathbb{P}^\star, \mathbb{P}^{b_\theta})=\text{Var}_{\mathbb{P}^v}\left[\log \frac{\mathrm{d}\mathbb{P}^\star}{\mathrm{d}\mathbb{P}^{b_\theta}}\right]=\mathbb{E}_{\mathbb{P}^v}\left[\left(\log \frac{\mathrm{d}\mathbb{P}^\star}{\mathrm{d}\mathbb{P}^{b_\theta}}-\mathbb{E}_{\mathbb{P}^v}\left[\log \frac{\mathrm{d}\mathbb{P}^\star}{\mathrm{d}\mathbb{P}^{b_\theta}}\right]\right)^2\right]
\end{align}
\end{small}
where $\boldsymbol{v}$ is an arbitrary control that enables \textit{off-policy learning} from trajectories that need not be generated by the current bias force $\boldsymbol{b}_\theta$. While the optimal solution to the LV objective is exactly the optimal bias force $\boldsymbol{b}^\star$, the LV objective is non-convex with respect to the path measure $\mathbb{P}^{b_\theta}$, unlike the cross-entropy objective below. We provide more details in App \ref{app:Extended Preliminaries}.

\paragraph{Cross-Entropy Objective}
To achieve a more theoretically favorable optimization problem, we propose a cross-entropy objective that is \textit{convex} with respect to the biased path measure $\mathbb{P}^{b_\theta}$ given by
\begin{small}
\begin{align}
    \mathcal{L}_{\text{CE}}(\mathbb{P}^\star, \mathbb{P}^{b_\theta}):=D_{\text{KL}}(\mathbb{P}^\star\|\mathbb{P}^{b_\theta})=\mathbb{E}_{\mathbb{P}^\star}\left[\log\frac{\mathrm{d}\mathbb{P}^\star}{\mathrm{d}\mathbb{P}^{b_\theta}}\right]=\mathbb{E}_{\mathbb{P}^{v}}\left[\frac{\mathrm{d}\mathbb{P}^\star}{\mathrm{d}\mathbb{P}^{v}}\log\frac{\mathrm{d}\mathbb{P}^\star}{\mathrm{d}\mathbb{P}^{b_\theta}}\right]\label{eq:CE-loss}
\end{align}
\end{small}

which equals the path cross-entropy $\mathbb{E}_{\mathbb{P}^\star}[-\log \frac{\mathrm{d}\mathbb{P}^{b_\theta}}{\mathrm{d}\mathbb{P}^0}]$ up to a constant independent of $\theta$, where $\mathbb{P}^\star$ is the target bridge measure and $\mathbb{P}^v$ is the path measure generated with an arbitrary control $\boldsymbol{v}$. To avoid taking the expectation with respect to the unknown path measure $\mathbb{P}^\star$, we define an importance weight $w^\star(\boldsymbol{X}_{0:T}):=\frac{\mathrm{d}\mathbb{P}^\star}{\mathrm{d}\mathbb{P}^{v}}(\boldsymbol{X}_{0:T})$ independent of $\boldsymbol{b}_\theta$ that enables optimization using trajectories generated from an arbitrary control $\boldsymbol{v}$. We note that similar CE objectives have been adopted in earlier work for different applications \citep{kappen2016adaptive, zhu2025mdns, tang2025tr2}, which we discuss in App \ref{app:Related Works}.

\begin{restatable}[Convexity and Uniqueness of Cross-Entropy Objective]{proposition}{crossentropy}\label{prop:Cross-Entropy}
    The cross-entropy objective $\mathcal{L}_{\text{CE}}$ is convex in $\mathbb{P}^{b_\theta}$, and there exists a unique minimizer $\boldsymbol{b}^\star$ that is the solution to the EntangledSOC problem in Proposition \ref{prop:EntangledSOC}.
\end{restatable}

The proof is given in App \ref{appendix-prop:Cross-Entropy}. To amortize the cost of simulation and reinforce high-reward trajectories, we define the arbitrary control as the biased path measure from the previous iteration $\boldsymbol{v}:=\bar{\boldsymbol{b}}=\texttt{stopgrad}(\boldsymbol{b}_\theta)$, which allows us to reuse the trajectories over multiple training steps by maintaining a replay buffer $\mathcal{R}$ that contains the trajectories $\boldsymbol{X}_{0:T}$ and their importance weights.

\begin{restatable}[Equivalence of Variational and Path Integral Objectives]{proposition}{crossentropytwo}
\label{prop:Cross-Entropy-2}
    The cross-entropy objective can be expressed in path-integral form as
    \begin{small}
    \begin{align}
        &\mathcal{L}_{\text{CE}}(\theta)=\mathbb{E}_{\mathbb{P}^{v}}\left[w^\star(\boldsymbol{X}_{0:T})\mathcal{F}_{\boldsymbol{b}_\theta, \boldsymbol{v}}(\boldsymbol{X}_{0:T})\right]\quad \text{s.t.}\quad \begin{cases}
            w^\star(\boldsymbol{X}_{0:T})=\frac{\mathrm{d}\mathbb{P}^\star}{\mathrm{d}\mathbb{P}^v}(\boldsymbol{X}_{0:T})=\frac{e^{r(\boldsymbol{X}_T)}}{Z}\frac{\mathrm{d}\mathbb{P}^0}{\mathrm{d}\mathbb{P}^v}(\boldsymbol{X}_{0:T})\\
        \mathcal{F}_{\boldsymbol{b}_\theta, \boldsymbol{v}}(\boldsymbol{X}_{0:T})=\frac{1}{2}\int_0^T\|\boldsymbol{b}_\theta(\boldsymbol{R}_t, \boldsymbol{V}_t)\|^2dt\\
        -\int_0^T(\boldsymbol{b}_\theta^\top \boldsymbol{v})(\boldsymbol{R}_t, \boldsymbol{V}_t)dt-\int_0^T\boldsymbol{b}_\theta(\boldsymbol{R}_t, \boldsymbol{V}_t)^\top d\boldsymbol{W}_t
        \end{cases}\nonumber
    \end{align}
    \end{small}
    where the off-policy control drift $\boldsymbol{v}=\bar{\boldsymbol{b}}:=\texttt{stopgrad}(\boldsymbol{b}_\theta)$ comes from the previous iteration, and we drop the $\theta$-independent term $w^\star(\boldsymbol{X}_{0:T})(r(\boldsymbol{X}_T)-\log Z)$. 
\end{restatable}

The proof is given in App \ref{appendix-prop:Cross-Entropy-2}. Since $Z$ is intractable, we use the self-normalized estimate $w^\star(\boldsymbol{X}^{(j)}_{0:T})\approx\texttt{softmax}_j\big(r(\boldsymbol{X}^{(j)}_T)+\log p^0(\boldsymbol{X}^{(j)}_{0:T})-\log p^{\bar{b}}(\boldsymbol{X}^{(j)}_{0:T})\big)$ over the $N$ trajectories in a batch. This estimate is consistent but biased at order $\mathcal{O}(1/N)$. The replay buffer $\mathcal{R}$ retains the highest-reward trajectories. When a batch contains a single trajectory (i.e., for large systems), the self-normalized weight is trivially $1$. We therefore draw trajectories from $\mathcal{R}$ in the TPS experiments with probability proportional to $\exp(r(\boldsymbol{X}_T))$, so that the reweighting by the terminal reward happens at the sampling stage. Setting $\boldsymbol{v}=\bar{\boldsymbol{b}}$ keeps the proposal close to the current model. 

\paragraph{Discrete Time Objective}
Since we want to train on off-policy trajectories from previous iterations to dynamically update the learned bias force given the velocities of the remaining particles, we require storing the simulated trajectories in a discretized form $\boldsymbol{X}_{0:K}:=(\boldsymbol{X}_k)_{k\in \{0, \dots, K\}}$ with step size $\Delta t$. To compute the term $\mathcal{F}_{\boldsymbol{b}_\theta, \bar{\boldsymbol{b}}}$ in the CE loss, we discretize it as
\begin{align}
    \hat{\mathcal{F}}_{\boldsymbol{b}_\theta, \bar{\boldsymbol{b}}}&(\boldsymbol{X}_{0:K})=\frac{1}{2}\sum_{k=0}^{K-1}\|\boldsymbol{b}_\theta(\boldsymbol{X}_k)\|^2\Delta t-\sum_{k=0}^{K-1}(\boldsymbol{b}_\theta\cdot\bar{\boldsymbol{b}})(\boldsymbol{X}_k)\Delta t-\sum_{k=0}^{K-1}\boldsymbol{b}_\theta(\boldsymbol{X}_k)\cdot \Delta\boldsymbol{W}_k\label{eq:discretized-F}
\end{align}

where $\Delta\boldsymbol{W}_k=\boldsymbol{\Sigma}^{-1}[\boldsymbol{X}_{k+1}-\boldsymbol{X}_k-(\boldsymbol{f}(\boldsymbol{X}_k)+\boldsymbol{\Sigma}\bar{\boldsymbol{b}}(\boldsymbol{X}_k))\Delta t]$. Now, we can establish a simplified version of the cross-entropy objective in Prop \ref{prop:Cross-Entropy-2}. 

\begin{restatable}[Discretized Cross-Entropy]{proposition}{discretized}\label{prop:discretized}
    Given the discretized $\hat{\mathcal{F}}_{\boldsymbol{b}_\theta, \bar{\boldsymbol{b}}}(\boldsymbol{X}_{0:K})$ in (\ref{eq:discretized-F}), we can derive a simplified loss function as
    \begin{align}
        \hat{\mathcal{L}}_{\text{CE}}(\theta)&=\mathbb{E}_{\boldsymbol{X}_{0:K}\sim \mathbb{P}^{\bar{b}}}\bigg[\underbrace{\frac{\mathrm{d}\mathbb{P}^\star}{\mathrm{d}\mathbb{P}^{\bar{b}}}(\boldsymbol{X}_{0:K})}_{w^\star(\boldsymbol{X}_{0:K})}\log \left(\frac{p^0(\boldsymbol{X}_{0:K})\exp(r(\boldsymbol{X}_K))}{p^{b_\theta}(\boldsymbol{X}_{0:K})}\right)\bigg]
    \end{align}
    where $w^\star(\boldsymbol{X}_{0:K})$ is the importance weight of the discrete time trajectory $\boldsymbol{X}_{0:K}$.
\end{restatable}

The proof is provided in App \ref{appendix-prop:discretized}. This allows us to train by sampling trajectories, tracking their log-probabilities under the biased and base path measures, computing the terminal reward $r(\boldsymbol{X}_K)$, storing the values in the replay buffer $\mathcal{R}$, and reusing the trajectories over $N_{\text{epochs}}$ training iterations.

\section{Related Works} 

\paragraph{Schrödinger Bridges} 
Bridge-matching methods such as DSBM \citep{Shi2023} learn first-order dynamics by repeatedly sampling the reference process pinned at both endpoints. Closed-form pinned bridges exist for linear references, including second-order phase-space dynamics \citep{stochastic-bridges-chen2016, chen2024phase}, but not for underdamped Langevin dynamics under a nonlinear force field $-\nabla U$. We therefore train on simulated trajectories with an off-policy SOC objective. Interacting-particle extensions use mean-field couplings among exchangeable particles \citep{Liu2022, zhang2025modeling}, whereas we make no exchangeability assumption. 

\paragraph{Transition Path Sampling} ML-based TPS methods include PIPS \citep{holdijk2023stochastic} and TPS-DPS \citep{seong2025transition}, which learn bias forces with SOC objectives, and Doob's Lagrangian \citep{du2024doobs}. Steered MD and PIPS bias along a fixed direction, and TPS-DPS uses a rank-one bias parallel to the target direction. Our parameterization contains the latter as a special case and adds $d-1$ orthogonal degrees of freedom (Sec~\ref{sec:Sampling Direction}).

An extended discussion of related works is in App~\ref{app:Related Works}.

\begin{table*}[t]
\caption{\textbf{Results for simulating cell cluster dynamics under Clonidine and Trametinib perturbation with EntangledSBM.} Best values for each PC dimension are \textbf{bolded}. We report RBF-MMD for all PCs and $\mathcal{W}_1$ and $\mathcal{W}_2$ distances of the top 2 PCs between ground truth and reconstructed clusters after $100$ simulation steps and cluster size set to $n=16$. The mean and standard deviation of metrics from 5 independent simulations are reported. We simulate PCs $d=\{50, 100, 150\}$ for Clonidine and $d=50$ for Trametinib to cells sampled from the training target distribution and unseen target distributions, and compare against the learned bias force with no velocity conditioning.}
\label{table:cell-metrics}
\begin{center}
\begin{small}
\resizebox{\linewidth}{!}{
\begin{tabular}{@{}lcccccc@{}}
\toprule
\textbf{Clonidine Perturbation} & \multicolumn{3}{c}{\textbf{Trained Target Distribution}} &\multicolumn{3}{c}{\textbf{Unseen Target Distribution}} \\
\textbf{Model} & RBF-MMD ($\downarrow$) & $\mathcal{W}_1$ ($\downarrow$) & $\mathcal{W}_2$ ($\downarrow$)  & RBF-MMD ($\downarrow$) & $\mathcal{W}_1$ ($\downarrow$) & $\mathcal{W}_2$ ($\downarrow$) \\
\midrule
\textbf{Base Dynamics} (50 PCs) & $0.677_{\pm 0.001}$ & $5.947_{\pm 0.005}$ & $6.015_{\pm 0.005}$ & $0.784_{\pm 0.001}$ & $8.217_{\pm 0.005}$ & $8.384_{\pm 0.005}$ \\
\midrule
\textbf{EntangledSBM w/o Velocity Conditioning}  & & &  & & & \\
50 PCs  & $0.440_{\pm 0.000}$ & $1.741_{\pm 0.003}$ & $1.857_{\pm 0.004}$ & $0.478_{\pm 0.000}$ & $2.907_{\pm 0.006}$ & $3.022_{\pm 0.006}$ \\
100 PCs & $0.494_{\pm 0.000}$ & $2.315_{\pm 0.004}$ & $2.423_{\pm 0.004}$ & $0.539_{\pm 0.000}$ & $4.110_{\pm 0.004}$ & $4.249_{\pm 0.003}$ \\
150 PCs & $0.510_{\pm 0.000}$ & $2.497_{\pm 0.006}$ & $2.620_{\pm 0.006}$ & $0.560_{\pm 0.000}$ & $4.573_{\pm 0.006}$ & $4.716_{\pm 0.007}$ \\
\midrule
\textbf{EntangledSBM w/ CE} &  & &   & & & \\
50 PCs  & $\mathbf{0.401}_{\pm 0.000}$ & $\mathbf{0.342}_{\pm 0.002}$ & $\mathbf{0.400}_{\pm 0.001}$ & $\mathbf{0.419}_{\pm 0.000}$ & $\mathbf{0.538}_{\pm 0.013}$ & $\mathbf{0.705}_{\pm 0.030}$ \\
100 PCs & $\mathbf{0.455}_{\pm 0.000}$ & $\mathbf{0.953}_{\pm 0.025}$ & $\mathbf{1.015}_{\pm 0.025}$ & $\mathbf{0.500}_{\pm 0.001}$ & $\mathbf{0.899}_{\pm 0.006}$ & $\mathbf{1.055}_{\pm 0.008}$ \\
150 PCs & $\mathbf{0.478}_{\pm 0.000}$ & $\mathbf{0.753}_{\pm 0.008}$ & $\mathbf{0.826}_{\pm 0.007}$ & $\mathbf{0.506}_{\pm 0.000}$ & $\mathbf{0.700}_{\pm 0.009}$ & $\mathbf{0.811}_{\pm 0.011}$  \\
\bottomrule
\end{tabular}
}
\end{small}
\end{center}

\begin{center}
\begin{small}
\resizebox{\linewidth}{!}{
\begin{tabular}{@{}lccccccccc@{}}
\toprule
\textbf{Trametinib Perturbation} & \multicolumn{3}{c}{\textbf{Trained Target Distribution}} &\multicolumn{3}{c}{\textbf{Unseen Target Distribution 1}} &\multicolumn{3}{c}{\textbf{Unseen Target Distribution 2}} \\
\textbf{Method} & RBF-MMD ($\downarrow$) & $\mathcal{W}_1$ ($\downarrow$) & $\mathcal{W}_2$ ($\downarrow$)  & RBF-MMD ($\downarrow$) & $\mathcal{W}_1$ ($\downarrow$) & $\mathcal{W}_2$ ($\downarrow$) & RBF-MMD ($\downarrow$) & $\mathcal{W}_1$ ($\downarrow$) & $\mathcal{W}_2$ ($\downarrow$)\\
\midrule
\textbf{Base Dynamics} & $0.938_{\pm 0.001}$ & $7.637_{\pm 0.005}$ & $7.653_{\pm 0.006}$ & $0.900_{\pm 0.000}$ & $7.766_{\pm 0.009}$ & $7.877_{\pm 0.009}$ & $0.754_{\pm 0.001}$ & $1.201_{\pm 0.009}$ & $1.455_{\pm 0.012}$ \\
\midrule
\textbf{EntangledSBM w/o Velocity Conditioning} & $0.449_{\pm 0.000}$ & $1.506_{\pm 0.005}$ & $1.544_{\pm 0.005}$ & $0.476_{\pm 0.000}$ & $2.116_{\pm 0.005}$ & $2.197_{\pm 0.005}$ & $0.480_{\pm 0.000}$ & $0.505_{\pm 0.004}$ & $0.627_{\pm 0.005}$ \\
\textbf{EntangledSBM w/ CE} & $\mathbf{0.428}_{\pm 0.000}$ & $\mathbf{0.392}_{\pm 0.005}$ & $\mathbf{0.434}_{\pm 0.006}$ & $\mathbf{0.409}_{\pm 0.000}$ & $\mathbf{0.453}_{\pm 0.008}$ & $\mathbf{0.561}_{\pm 0.009}$ & $\mathbf{0.451}_{\pm 0.000}$ & $\mathbf{0.394}_{\pm 0.003}$ & $\mathbf{0.469}_{\pm 0.004}$ \\
\bottomrule
\end{tabular}
}
\end{small}
\end{center}
\end{table*}

\section{Experiments}
\label{sec:Experiments}

We evaluate \textbf{EntangledSBM} on trajectory simulation tasks involving $n$-particle systems with interacting dynamics: drug-perturbed cell cluster dynamics (Sec~\ref{experiment:Cell}), pancreatic $\beta$-cell differentiation (Sec~\ref{experiment:Beta}), and transition path sampling (TPS) for all-atom protein dynamics (Sec~\ref{experiment:TPS}).

\subsection{Simulating Interacting Cell Dynamics Under Perturbation}
\label{experiment:Cell}
Cell populations undergo dynamic signaling interactions that shift their state under drug perturbation, a process central to drug discovery. We use \textbf{EntangledSBM} to parameterize an \textit{entangled} bias force that learns \textbf{dynamic interactions between cells}, guiding a batch of cells toward the perturbed state. Beyond reconstructing in-distribution targets along the data manifold, EntangledSBM generates paths to \textit{unseen} target distributions, demonstrating its potential to \textbf{scale to diverse perturbations and cell types under sparse data}.

\paragraph{Setup and Baselines} 
We use two perturbations from Tahoe-100M~\citep{tahoe-100m}, Clonidine and Trametinib (both at $5\mu M$), reduced to the top 2000 HVGs, followed by PCA. To probe generalization to \textit{unseen} targets, we partition the perturbed cells into disjoint clusters (two for Clonidine, three for Trametinib; Fig.~\ref{fig:tahoe-datasets}), train on one, and hold out the rest. The held-out clusters are spatially disjoint from the training cluster but lie on the same gene-expression manifold. This setting therefore tests in-manifold generalization, not zero-shot generalization to unseen perturbations. The bias force is trained on batches of $n=16$ cells with positions and velocities $\boldsymbol{X}_t=(\boldsymbol{R}_t, \boldsymbol{V}_t)$ over $d$ PCs, where the energy landscape places low energy at high data density (App~\ref{app:data-manifold}). After 100 simulation steps, we report RBF-MMD across all PCs and Wasserstein distances ($\mathcal{W}_1$, $\mathcal{W}_2$) over the top two PCs. We ablate velocity conditioning by training on positions alone, and compare the cross-entropy objective $\mathcal{L}_{\text{CE}}$ against the log-variance divergence $\mathcal{L}_{\text{LV}}$ (App~\ref{app:LV-comparison},~\ref{app:cell-experiment}).

\paragraph{Clonidine}
EntangledSBM accurately steers base dynamics -- which otherwise stay in the initial high-density region -- to the perturbed target across $d\in\{50, 100, 150\}$ (Fig.~\ref{fig:clonidine-full}; Table~\ref{table:cell-metrics}), and \textbf{generalizes to unseen perturbed clusters} that diverge from the training distribution by conditioning on the target. Masking the velocity input while keeping the architecture fixed degrades reconstruction on every metric. The gap is largest on unseen targets ($\mathcal{W}_1$ $0.538\to2.907$), where a position-only model must extrapolate (Table~\ref{table:cell-metrics}). The two objectives also behave differently. $\mathcal{L}_{\text{LV}}$ attains lower terminal error but produces near-straight, abrupt paths, whereas $\mathcal{L}_{\text{CE}}$ traces smooth trajectories along the data manifold (Table~\ref{table:clonidine-metrics-full}; Fig.~\ref{fig:clonidine-full}; App~\ref{app:LV-comparison}).

\paragraph{Trametinib}
Trametinib induces three distinct perturbed distributions (Fig.~\ref{fig:tahoe-datasets}B). Trained on only one, EntangledSBM with $\mathcal{L}_{\text{CE}}$ accurately reconstructs the remaining two \textit{without} retraining and substantially outperforms the velocity-ablated variant (Table~\ref{table:cell-metrics}; Fig.~\ref{fig:trametinib-full}). $\mathcal{L}_{\text{LV}}$ again fails to capture intermediate dynamics (Table~\ref{table:trametinib-metrics-full}; Fig.~\ref{fig:trametinib-full}). 

\subsection{Pancreatic $\beta$-cell Differentiation}
\label{experiment:Beta}

Reconstructing differentiation trajectories from sparse population snapshots tests whether a model captures the structure of cell-fate dynamics or merely interpolates observed marginals. We show that EntangledSBM, supervised only on the initial and terminal distributions, improves on mean-field and unbalanced Schrödinger bridge baselines at late time points, even though both baselines are trained on every intermediate snapshot.

\paragraph{Setup and Baselines}
We compare against CytoBridge \citep{zhang2025modeling}, a mean-field Schrödinger bridge model of cell-cell interactions, and DeepRUOT \citep{Zhang2024}, which learns unbalanced path measures through regularized unbalanced optimal transport. Both baselines are trained on every intermediate snapshot $t\in\{1,\dots,6\}$ following their intended protocol, and we report their published numbers. EntangledSBM receives only $t=0$ and $t=7$ as endpoint distributions. Intermediate structure enters only through the base potential $U$, the RBF data-manifold metric of App~\ref{app:data-manifold}, fit on cells pooled across all time points without time labels. Since $\mathcal{L}_{\text{CE}}$ penalizes control cost relative to these base dynamics, paths through low-density regions are costly, so the learned bridge follows high-density intermediate regions without being trained on the intermediate marginals (App~\ref{app:pancreatic-cell}).

\paragraph{Results}
EntangledSBM is less accurate than the baselines at early time points ($t\le5$), but it is the most accurate at $t=6$ and $t=7$, reducing terminal-distribution error by $46\%$ relative to CytoBridge ($4.21$ vs.\ $7.83$ in Table~\ref{tab:beta-cell}). The terminal state corresponds to mature $\beta$-cell commitment, where fine-grained cell-cell dependencies are expected to matter most. This suggests that pairwise entangled dynamics resolve late-stage interactions that mean-field couplings average out. $\mathcal{L}_{\text{CE}}$ also outperforms $\mathcal{L}_{\text{LV}}$ on 5 of the 6 intermediate marginals, consistent with $\mathcal{L}_{\text{CE}}$ learning path measures with meaningful intermediate dynamics.
 
\begin{table*}[t]
\centering
\caption{\textbf{Results for pancreatic $\beta$-cell differentiation modeling.} 1-Wasserstein distances at each observed time point (lower is better). Baselines are trained on all intermediate snapshots; EntangledSBM is trained only on $t=0$ and $t=7$. $\dagger$ indicates values taken from \citet{zhang2025modeling}. \textbf{Bold} indicates the best result per column.}
\label{tab:beta-cell}
\small
\resizebox{\linewidth}{!}{
\begin{tabular}{lccccccc}
\toprule
\textbf{Model} & $t=1$ & $t=2$ & $t=3$ & $t=4$ & $t=5$ & $t=6$ & $t=7$ \\
\midrule
\textbf{DeepRUOT}$^{\dagger}$ & $\mathbf{8.0447}_{\pm 0.0005}$ & $8.0773_{\pm 0.0021}$ & $7.6301_{\pm 0.0032}$ & $\mathbf{8.0064}_{\pm 0.0042}$ & $\mathbf{7.9018}_{\pm 0.0117}$ & $8.3977_{\pm 0.0102}$ & $7.8346_{\pm 0.0109}$ \\
\textbf{CytoBridge}$^{\dagger}$ & $8.0448_{\pm 0.0005}$ & $\mathbf{8.0771}_{\pm 0.0021}$ & $\mathbf{7.6299}_{\pm 0.0032}$ & $8.0066_{\pm 0.0043}$ & $\mathbf{7.9018}_{\pm 0.0117}$ & $8.3974_{\pm 0.0102}$ & $7.8343_{\pm 0.0109}$ \\
\midrule
\textbf{EntangledSBM w/ LV} & $17.9857_{\pm 0.0015}$ & $16.0905_{\pm 0.0014}$ & $12.7845_{\pm 0.0019}$ & $11.1558_{\pm 0.0015}$ & $9.0821_{\pm 0.0017}$ & $8.3907_{\pm 0.0007}$ & $4.3539_{\pm 0.0006}$ \\
\textbf{EntangledSBM w/ CE} & $13.2059_{\pm 0.0083}$ & $14.2824_{\pm 0.0140}$ & $12.8030_{\pm 0.0097}$ & $10.6162_{\pm 0.0130}$ & $8.8382_{\pm 0.0138}$ & $\mathbf{8.1769}_{\pm 0.0070}$ & $\mathbf{4.2109}_{\pm 0.0052}$ \\
\bottomrule
\end{tabular}
}
\end{table*}

\subsection{Transition Path Sampling of Protein Folding Dynamics}
\label{experiment:TPS}

Simulating rare transitions across high-energy barriers in molecular dynamics (MD) is challenging due to the high dimensionality of biomolecules and long timescales involved. We show that EntangledSBM generates feasible \textit{all-atom} transition paths with higher target accuracy than baselines.

\paragraph{Setup and Baselines}
Following~\citet{seong2025transition}, we simulate positions and velocities with OpenMM~\citep{eastman2017openmm} under an annealed temperature schedule and report the Kabsch-aligned RMSD over 64 paths (\textbf{RMSD}; \r{A})~\citep{kabsch1976solution}, target hit percentage (\textbf{THP}; \%), and highest-energy transition state along successful paths (\textbf{ETS}; kJ/mol). Baselines include unbiased MD (\textbf{UMD}) with sampling temperatures at $3600$K for Alanine Dipeptide and $1200$K for the fast-folding proteins, steered MD (\textbf{SMD}~\citep{schlitter1994targeted, izrailev1999steered}) at $10$K and $20$K, path integral path sampling (\textbf{PIPS}~\citep{holdijk2023stochastic}), and TPS with diffusion path samplers (\textbf{TPS-DPS}~\citep{seong2025transition}, scale parameterization); see App~\ref{app:tps-experiment} for temperatures and hyperparameters. The terminal reward is computed over all Kabsch-aligned heavy atoms, and collective variables are used only to define hits for THP.

\begin{figure*}[t]
    \centering
    \includegraphics[width=\linewidth]{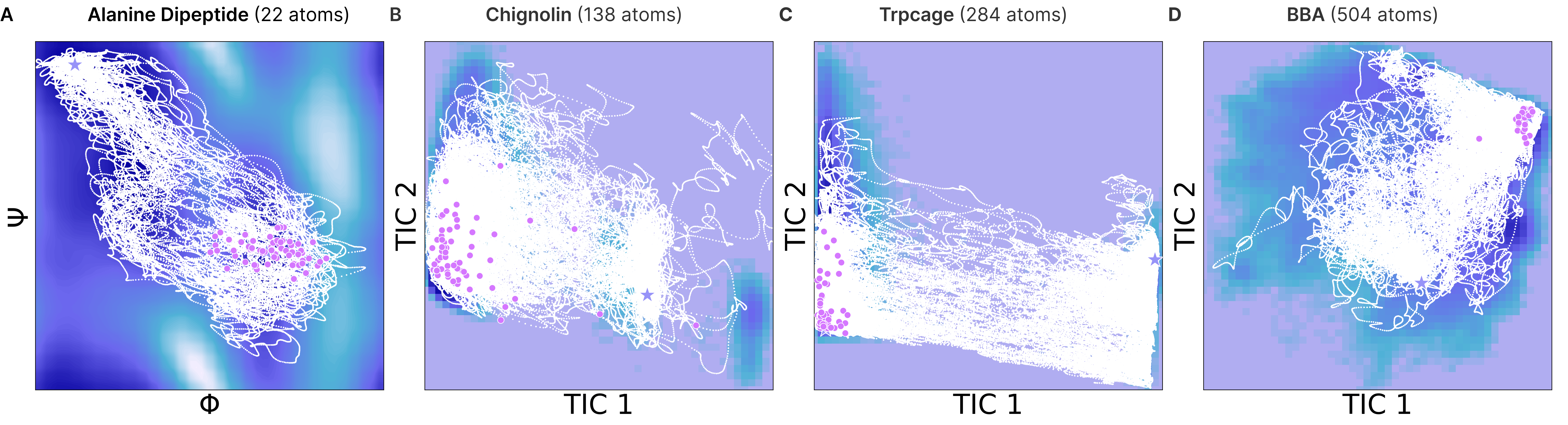}
    \caption{\textbf{Transition paths generated with EntangledSBM.} Energy landscapes (dark navy: low; light purple: high) are plotted over backbone dihedrals $(\phi, \psi)$ for Alanine Dipeptide and the top two TICA components for the fast-folding proteins. Purple stars mark the start $\boldsymbol{R}_0$ and target $\boldsymbol{R}_\mathcal{B}$ states; pink circles mark the simulated endpoints $\boldsymbol{R}_T$.}
    \label{fig:TPS-benchmark}
\end{figure*}

\paragraph{Alanine Dipeptide}
\label{experiment:ALDP}
For alanine dipeptide (22 atoms), we target $\pi_\mathcal{B}=\{\boldsymbol{R}\mid\|\xi(\boldsymbol{R})-\xi(\boldsymbol{R}_\mathcal{B})\|< 0.75\}$ defined by backbone dihedrals $\xi(\boldsymbol{R}) =(\phi, \psi)$. EntangledSBM generates feasible trajectories through \textit{both} saddle points, recovering the two reaction channels (Fig.~\ref{fig:TPS-benchmark}A), and achieves the highest THP and lowest RMSD among all baselines (Table~\ref{table:tps-benchmark}).

\begin{table*}[h]
\caption{\textbf{Transition path sampling benchmarks with EntangledSBM.} Best values are \textbf{bolded}. All metrics are averaged over 64 paths. Unless specified in brackets, paths are generated at $300$K for Alanine Dipeptide and Chignolin and $400$K for the others. $\dagger$ denotes values taken from \citet{seong2025transition}.}
\label{table:tps-benchmark}
\begin{center}
\begin{small}
\resizebox{\linewidth}{!}{
\begin{tabular}{@{}lccclccc@{}}
\toprule
\textbf{Protein} & \multicolumn{3}{c}{\textbf{Alanine Dipeptide}} &  & \multicolumn{3}{c}{\textbf{Chignolin}}  \\
\midrule
\textbf{Method} & RMSD ($\downarrow$) & THP ($\uparrow$) & ETS ($\downarrow$) & \textbf{Method} & RMSD ($\downarrow$) & THP ($\uparrow$) & ETS ($\downarrow$) \\
\midrule
UMD $\dagger$ & $1.19_{\pm 0.32}$ & $6.25$ & $812.47_{\pm 148.80}$ & UMD $\dagger$ & $7.23_{\pm 0.93}$ & $1.56$ & $388.17$ \\
SMD (10) $\dagger$  & $0.86_{\pm 0.21}$ & $7.81$ & $33.15_{\pm 6.46}$ & SMD (10k) $\dagger$  & $1.26_{\pm 0.31}$ & $6.25$ & $-527.95_{\pm 93.58}$ \\
SMD (20) $\dagger$ & $0.56_{\pm 0.27}$ & $54.69$ & $78.40_{\pm 12.76}$ & SMD (20k) $\dagger$ & $\mathbf{0.85}_{\pm 0.24}$ & $34.38$ & $179.52_{\pm 138.87}$ \\
PIPS (Force) $\dagger$ & $0.66_{\pm 0.15}$ & $43.75$ & $28.17_{\pm 10.86}$ & PIPS (Force) $\dagger$ & $4.66_{\pm 0.17}$ & $0.00$ & $-$ \\
TPS-DPS (Scale Param.) $\dagger$ & $0.25_{\pm 0.20}$ & $76.00$ & $\mathbf{22.79}_{\pm 13.57}$ & TPS-DPS (Scale Param.) $\dagger$ & $1.17_{\pm 0.66}$ & $59.38$ & $\mathbf{-780.18}_{\pm 216.93}$ \\
\midrule
\textbf{EntangledSBM (Ours)} & $\mathbf{0.18}_{\pm 0.07}$ & $\mathbf{92.19}$ & $47.91_{\pm 22.76}$ & \textbf{EntangledSBM (Ours)} & $0.92_{\pm 0.13}$ & $\mathbf{64.06}$ & $2825.61_{\pm 318.94}$ \\
\bottomrule
\end{tabular}
}
\end{small}
\end{center}
\begin{center}
\begin{small}
\resizebox{\linewidth}{!}{
\begin{tabular}{@{}lccclccc@{}}
\toprule
\textbf{Protein} & \multicolumn{3}{c}{\textbf{Trpcage}} & & \multicolumn{3}{c}{\textbf{BBA}} \\
\midrule
\textbf{Method} & RMSD ($\downarrow$) & THP ($\uparrow$) & ETS ($\downarrow$) & \textbf{Method} & RMSD ($\downarrow$) & THP ($\uparrow$) & ETS ($\downarrow$)  \\
\midrule
UMD $\dagger$ & $8.27_{\pm 1.13}$ & $0.00$ & $-$ & UMD $\dagger$ & $10.81_{\pm 1.05}$ & $0.00$ & - \\
SMD (10k) $\dagger$ & $1.68_{\pm 0.23}$ & $3.12$ & $-312.54_{\pm 20.67}$ & SMD (10k) $\dagger$ & $2.89_{\pm 0.32}$ & $0.00$ & -  \\
SMD (20k) $\dagger$ & $1.20_{\pm 0.20}$ & $42.19$ & $-226.40_{\pm 85.59}$ & SMD (20k) $\dagger$  & $1.66_{\pm 0.30}$ & $26.56$ & $-3104.95_{\pm 97.57}$  \\
PIPS (Force) $\dagger$ & $7.47_{\pm 0.19}$ & $0.00$ & - & PIPS (Force) $\dagger$ & $9.84_{\pm 0.18}$ & $0.00$ & -  \\
TPS-DPS (Scale Param.) $\dagger$ & $\mathbf{0.76}_{\pm 0.12}$ & $81.25$ & $\mathbf{-317.61}_{\pm 140.89}$ & TPS-DPS (Scale Param.) $\dagger$ & $1.21_{\pm 0.09}$ & $84.38$ & $\mathbf{-3801.68}_{\pm 139.38}$  \\
\midrule
\textbf{EntangledSBM} (Ours) & $1.04_{\pm 0.22}$ & $\mathbf{82.81}$ & $765.74_{\pm 155.28}$ & \textbf{EntangledSBM} (Ours) & $\mathbf{0.84}_{\pm 0.08}$ & $\mathbf{96.88}$ & $1453.80_{\pm 367.84}$ \\
\bottomrule
\end{tabular}
}
\end{small}
\end{center}
\end{table*}

\paragraph{Fast-Folding Proteins}
\label{experiment:Fast-Folding Proteins}
We scale to all-atom transition paths for three fast folders: \textbf{Chignolin} (10 aa, 138 atoms), \textbf{Trpcage} (20 aa, 284 atoms), and \textbf{BBA} (28 aa, 504 atoms). A hit reaches $\pi_\mathcal{B}$ defined over the top two TICA components~\citep{perez2013identification}. EntangledSBM produces \textit{diverse} transition paths over high-energy barriers (Fig.~\ref{fig:TPS-benchmark}), achieving the best THP across all proteins and competitive or lower RMSD (Table~\ref{table:tps-benchmark}). The higher average ETS reflects the larger volume and diversity of successful paths. We verify in App \ref{app:diversity} that this corresponds to mechanistically distinct folding routes rather than energy-landscape mismatch, and in App \ref{app:physics_eval} that the resulting trajectories remain physically feasible. Given that base dynamics $\mathbb{P}^0$ rarely escape the initial metastable basin, these results show EntangledSBM effectively learns the target bridge dynamics $\mathbb{P}^\star$ even across substantial energy barriers. 

\section{Conclusion}
\label{sec:conclusion}
In this work, we present \textbf{Entangled Schrödinger Bridge Matching (EntangledSBM)}, a principled framework for learning the second-order dynamics of interacting multi-particle systems through entangled bias forces and an unconstrained cross-entropy objective. EntangledSBM captures dependencies between particle positions and velocities, enabling the modeling of complex dynamics across biological scales. For perturbation modeling, EntangledSBM reconstructs perturbed cell states while generalizing to divergent target states not seen during training, and for molecular dynamics (MD), it generates physically plausible transition paths for fast-folding proteins at an all-atom resolution. 

\paragraph{Limitations and Future Work} 
To conclude, we highlight some limitations and future directions of our work with an extended discussion in App~\ref{app:Limitations}. First, our cell-state results show in-manifold generalization to held-out clusters of perturbations seen during training, and we have not yet tested new perturbations. Conditioning the bias force (e.g. on perturbation identity) is a promising route toward generalizing predictions to unseen systems (e.g. unseen drugs and cell types). Second, our protein validation is structural and trajectory-based, so we do not recover rates, committors, or free-energy surfaces. Since EntangledSBM tracks path likelihoods during simulation, future work could reweight generated paths to estimate these kinetic and thermodynamic quantities. Finally, all-particle attention costs $\mathcal{O}(n^2)$, which limits scaling to larger systems. Sparse or local attention could extend EntangledSBM to model multi-domain complexes, functional interactions between proteins and ligands, and other interacting multi-particle systems.

\section*{Declarations}
\paragraph{Acknowledgments} We thank Mark III Systems for providing database and hardware support that has contributed to the research reported within this manuscript. 

\paragraph{Author Contributions} S.T. conceived the study. S.T. devised and developed model architectures and theoretical formulations, and trained and benchmarked models. Y.Z. advised on model design and theoretical framework. S.T. drafted the manuscript and designed the figures. P.C. supervised and directed the study, and reviewed and finalized the manuscript.

\paragraph{Data and Materials Availability} The codebase is freely accessible to the academic community at \url{https://github.com/sophtang/EntangledSBM} and \url{https://huggingface.co/ChatterjeeLab/EntangledSBM}.

\paragraph{Funding Statement} This research was supported by NIH grant R35GM155282 to the lab of P.C.

\paragraph{Competing Interests} P.C. is a co-founder of Gameto, Inc., UbiquiTx, Inc., AtomBioworks, Inc., and Kosha Sciences, Inc. and advises companies involved in therapeutics development. P.C.’s interests are reviewed and managed by the University of Pennsylvania in accordance with their conflict-of-interest policies. S.T. and Y.Z. have no conflicts of interest to declare.

\bibliographystyle{unsrtnat}
\bibliography{citation}  

\clearpage
\appendix
\section*{Appendix}
\startcontents[app]
\printcontents[app]{l}{1}{\setcounter{tocdepth}{2}}

\section{Broader Impact}
\label{app:impact-statement}
This work introduces a computational framework for modeling interacting multi-particle dynamics with applications to protein folding and cell state perturbations. The cell perturbation experiments (Section~\ref{experiment:Cell}) use the publicly available Tahoe-100M \citep{tahoe-100m} and pancreatic $\beta$-cell \citep{veres2019charting} datasets, cited with attribution and used under their original licenses. All biomolecular simulations (Section~\ref{experiment:TPS}) were performed on standard test proteins (Alanine Dipeptide, Chignolin, Trpcage, BBA) using publicly available force fields and simulation protocols, with no human or animal subjects involved. No proprietary or sensitive data were employed. The work does not provide experimental protocols or actionable therapeutic targets, and is restricted to methodological contributions in generative modeling. While such frameworks may be broadly useful for molecular or cellular design, the immediate scope of EntangledSBM is fundamental methodology, and the dual-use risk is minimal. 

\section{Related Works}
\label{app:Related Works}
\paragraph{Off-Policy Learning}
Importance-weighted cross-entropy (CE) objectives have previously been used to sample the target distribution $g^\star\propto hf_u$ by aligning a proposal distribution $f_v$ to the intractable target with $\text{KL}(g^\star\|f_v)$ \citep{kappen2016adaptive}. The main difference is that our approach aims to match a \textbf{controlled path measure} $\mathbb{P}^{b_\theta}$ with an energy-minimizing path measure $\mathbb{P}^\star$ rather than a static target distribution. Furthermore, we adapt our path CE objective to apply to discrete-time paths, similar to the approach taken by \citet{seong2025transition}, which discretizes the log-variance (LV) divergence \citep{nusken2021solving}. While the LV objective has theoretical optimality guarantees, it is non-convex in $\mathbb{P}^{b_\theta}$. Since the CE objective is convex in $\mathbb{P}^{b_\theta}$, it yields an ideal optimization landscape for $\theta$ and demonstrates improved performance as shown in Sec \ref{experiment:Cell}. We also highlight that our optimization scheme is a form of \textit{off-policy} learning, as we minimize the objective with respect to the path distribution generated from an auxiliary control $\boldsymbol{v}$, which we set to the non-gradient-tracking bias force $\boldsymbol{v}:=\bar{\boldsymbol{b}}=\texttt{stopgrad}(\boldsymbol{b}_\theta)$. This objective is inspired by recent work that leverages off-policy reinforcement learning in the discrete state space for discrete diffusion sampling \citep{zhu2025mdns} and fine-tuning \citep{tang2025tr2}.

\paragraph{Transition Path Sampling}
\label{related:TPS}
To overcome the challenge of generating feasible transition paths across high energy barriers for transition path sampling, several non-ML and ML-based approaches have been introduced \citep{bolhuis2002transition, dellago1998transition, vanden2010transition}. Non-ML approaches often rely on collective variables (CVs), which reduce the dimensionality of the molecular conformation to a function of coordinates that are known to be involved in transition states \citep{hooft2021discovering}. These include steered MD \citep{schlitter1994targeted, izrailev1999steered}, umbrella sampling \citep{torrie1977nonphysical, kastner2011umbrella}, meta-dynamics \citep{laio2002escaping, ensing2006metadynamics, branduardi2012metadynamics, bussi2015free}, adaptive biasing force \citep{comer2015adaptive}, and on-the-fly probability-enhanced sampling \citep{invernizzi2020rethinking}. To overcome the lack of well-understood CVs for large biomolecular systems, ML-based methods have been explored for determining or constructing CVs \citep{sultan2018automated, rogal2019neural, chen2018molecular, sun2022multitask, sipka2023constructing}, but modeling transition paths at all-atom resolution is still desirable due to systems of increased complexity, and CVs are uncertain. 

Several ML-based approaches for simulating transition paths have been developed \citep{singh2023variational, yan2022learning, sipka2023differentiable, das2021reinforcement}. Notably, Path Integral Path Sampling (PIPS; \citet{holdijk2023stochastic}) learns a bias force with stochastic optimal control theory, Doob's Lagrangian \citep{du2024doobs} defines an optimal transition path with a Lagrangian objective that solves the Doob's $h$-transform, Transition Path Sampling with Diffusion Path Samplers (TPS-DPS; \citet{seong2025transition}) trains a bias force with an off-policy log-variance objective, and Action Minimization with Onsager-Machlup (OM) Functional \citep{raja2025actionminimization} generates a discrete interpolation between endpoints by minimizing the OM functional. Furthermore, \citet{blessing2025trust} introduces a trust-region-constrained SOC optimization algorithm, which is applied to TPS. While these approaches demonstrate the promise of ML-based TPS, their applicability remains limited to small-scale biomolecular systems (Alanine Dipeptide, tetrapeptides, etc.) for all-atom simulation \citep{holdijk2023stochastic, du2024doobs} or require coarse-grained representations when scaling up to larger fast-folding proteins \citep{raja2025actionminimization}. We directly compare against PIPS \citep{holdijk2023stochastic} and TPS-DPS \citep{seong2025transition} and achieve the highest target hit percentage on all-atom simulations of larger fast-folding proteins. Furthermore, we highlight that our framework extends beyond molecular dynamics (MD) simulations, with applications to single-cell simulation and generalization to unseen target distributions.

\paragraph{Modeling Cell Dynamics Under Perturbation}
\label{related:Perturbation}
Predicting the dynamics of heterogeneous cell populations under perturbations such as treatment with a drug candidate, gene editing, and knockouts, or protein expression, has critical applications in drug discovery and predictive medicine \citep{Shalem2014, Kramme2021, Dixit2016, Gavriilidis2024, tahoe-100m, Kobayashi2022, PiersonSmela2023, PiersonSmela2025, yeo2021generative}. Generative modeling frameworks, including flow matching \citep{zhang2025cellflow, rohbeck2025modeling, atanackovic2024meta, tong24a, wang2025joint}, Schrödinger Bridge Matching \citep{alatkar2025artemis, tong24a, Kapusniak2024, tang2025branched}, and stochastic optimal control \citep{Zhang2024, zhang2025modeling}, have enabled significant advances in parameterizing velocities that evolve dynamically over time on non-linear energy manifolds by learning on static snapshots across the trajectory. However, these models often require \textit{explicit training on a specific target distribution} and fail to generalize to unseen target distributions at inference, especially if they diverge from those seen in the training data, which limits their scalability to cell types and perturbations that are not seen during training. Furthermore, many of these approaches simulate cells as populations of particles that evolve independently from each other, which fails to capture the diverse intercellular signaling and interactions that occur naturally and under perturbations. Although there have been some early developments in learning these interactions from data \citep{zhang2025modeling}, they operate under the mean-field assumption, where each cell takes the average effect of all surrounding cells rather than simulating individual pairwise interactions. This assumption does not account for the complex interactions between cells and limits its applicability to heterogeneous cell populations.

\section{Limitations and Future Directions}
\label{app:Limitations}
While our results demonstrate the effectiveness of EntangledSBM across both cellular and molecular regimes, several limitations should be noted to properly contextualize the scope of our claims.

First, the cell-state results reported here characterize in-manifold generalization: the model is evaluated on held-out cells drawn from the same perturbation distributions seen during training, rather than on entirely unseen perturbations. The bridge formulation can therefore be expected to interpolate trajectories within a perturbation manifold it has already observed, but our experiments do not establish that the model extrapolates to novel chemical or genetic perturbations whose endpoint distributions lie outside the training support. Extending EntangledSBM to a true zero-shot perturbation setting would require conditioning on perturbation identity in a way that supports compositional or out-of-distribution generalization, which we leave to future work.

Second, our protein-system validation is primarily structural and trajectory-based. We assess the physical plausibility of generated transitions (e.g., energy of transition state, smoothness, etc.) but do not conduct full kinetic or thermodynamic validation. We do not recover transition rates, committor probabilities, or free-energy surfaces at quantitative accuracy comparable to enhanced-sampling molecular dynamics, both because such validation lies outside the scope of the bridge formulation and because it would require substantially greater computational resources. The trajectories should therefore be interpreted as physically plausible interpolants rather than calibrated samples from the underlying Boltzmann transition density.

Third, the all-particle attention mechanism that underlies the entanglement step incurs $O(n^2)$ time and memory cost in the number of particles. We demonstrate that this remains tractable for systems of $500+$ atoms, which covers the protein and cellular regimes considered here, but the quadratic scaling presents a clear barrier to substantially larger systems such as full multi-domain complexes, membrane assemblies, or dense tissue-scale simulations. Scaling further will likely require sparse, hierarchical, or locality-aware attention variants -- for example, neighborhood-restricted attention, low-rank approximations, or message-passing schemes that exploit the geometric locality of physical interactions -- and integrating such architectures into the bridge framework without disrupting its probabilistic guarantees is left for future work.

\section{Extended Theoretical Preliminaries}
\label{app:Extended Preliminaries}
In this section, we provide relevant theoretical background in stochastic optimal control (SOC) theory for path measures. We use the terms \textit{bias force} and \textit{optimal control} interchangeably, denoted by $\boldsymbol{b}$, which is less common in general SOC literature where $\boldsymbol{u}$ is often used. For a more comprehensive analysis of SOC theory, we refer the reader to \citet{nusken2021solving} and \citet{tang2026foundations}. 

\subsection{Notation} 
In this work, we consider an $n$-particle system $\boldsymbol{X}_t=(\boldsymbol{R}_t, \boldsymbol{V}_t)$ where $\boldsymbol{R}_t\in \mathbb{R}^{n\times d}$ and $\boldsymbol{V}_t\in \mathbb{R}^{n\times d}$ denote the positions and velocities of the full system that evolves over the time horizon $t\in[0,T]$. The notation $\boldsymbol{X}$ indicates a random variable and $\boldsymbol{x}$ denotes a deterministic realization. Each particle $i$ is defined by its position $\boldsymbol{r}^i_t\in \mathbb{R}^d$ and velocity $\boldsymbol{v}_t^i\in \mathbb{R}^d$ which lie in $d$-dimensional Euclidean space. The base dynamics evolve according to a potential energy function denoted $U(\boldsymbol{R}_t): \mathbb{R}^{n\times d}\to\mathbb{R}$ and the biased dynamics evolve with an additional parameterized bias force $\boldsymbol{b}_\theta(\boldsymbol{R}_t, \boldsymbol{V}_t): \mathbb{R}^{n\times d}\times \mathbb{R}^{n\times d}\to \mathbb{R}^{n\times d}$ that captures the implicit dependencies within the system and controls the base dynamics towards a target distribution $\pi_\mathcal{B}(\boldsymbol{R}_T)$ from a state in the initial distribution $\boldsymbol{R}_\mathcal{A}\in \pi_\mathcal{A}(\boldsymbol{R}_0)$. The corresponding path measures over trajectories $\boldsymbol{X}_{0:T}:=(\boldsymbol{X}_t)_{t\in [0,T]}$ are denoted $\mathbb{P}^0$ for the base process and $\mathbb{P}^{b_\theta}$ for the biased process. The optimal path measure that solves the EntangledSB problem is denoted $\mathbb{P}^\star\equiv \mathbb{P}^{b^\star}$ with bias force $\boldsymbol{b}^\star(\boldsymbol{R}_t, \boldsymbol{V}_t)$.

In the SOC derivations, we write the controlled SDE as in (\ref{eq:control-SDE}), where the control enters the drift as $\boldsymbol{\Sigma}\boldsymbol{b}$, so that the running cost is $\frac{1}{2}\|\boldsymbol{b}\|^2$. The physical bias force maps to this form after rescaling by $\boldsymbol{\Sigma}_V^{-1}\boldsymbol{M}^{-1}$ in the second-order dynamics (\ref{eq:multi-langevin}), with $\boldsymbol{\Sigma}_V=\mathrm{diag}\big(\sqrt{2\gamma k_B\tau/m_i}\big)$, and by $(\gamma\boldsymbol{\Sigma})^{-1}$ in the first-order dynamics, with $\boldsymbol{\Sigma}=\sqrt{2k_B\tau/\gamma}\,\boldsymbol{I}$.

\subsection{Stochastic Optimal Control}
\label{app:ext-SOC-background}

\paragraph{Controlled Path Measures}
First, we consider the controlled path measure $\mathbb{P}^b$ defined by the SDE:
\begin{align}
    d\boldsymbol{X}_t=[\boldsymbol{f}(\boldsymbol{X}_t, t)+\boldsymbol{\Sigma} \boldsymbol{b}(\boldsymbol{X}_t,t)]dt+\boldsymbol{\Sigma} d\boldsymbol{W}_t\label{eq:control-SDE}
\end{align}
where $\boldsymbol{b}:[0,T]\times \mathcal{X}\to\mathcal{X}\in \mathcal{U}$ is known as the \textbf{control}\footnote{$\mathcal{U}$ is the set of admissible controls continuously differentiable in $(\boldsymbol{x}, t)$ with bounded linear growth in $\boldsymbol{x}$} that tilts the path measure from the unconditional dynamics $\mathbb{P}^0$. Now, we will define the \textbf{Radon-Nikodym derivative (RND)} of the path measures corresponding to the controlled and unconditional SDEs, which is crucial for defining our objective.

\begin{lemma}[Radon-Nikodym Derivative]\label{lemma:RND}
    Given a pair of SDEs with and without a control drift $\boldsymbol{b}(\boldsymbol{X}_t,t)$ on $t\in [0,T]$ with the same diffusion $\boldsymbol{\Sigma}$ defined as
    \begin{align}
        \mathbb{P}^0: d\boldsymbol{X}_t&=\boldsymbol{f}(\boldsymbol{X}_t, t)dt+\boldsymbol{\Sigma}d\boldsymbol{W}_t, \quad &\boldsymbol{X}_0=\boldsymbol{x}_0\\
        \mathbb{P}^b: d\boldsymbol{X}^b_t&=\left[\boldsymbol{f}(\boldsymbol{X}_t, t)+ \boldsymbol{\Sigma}\boldsymbol{b}(\boldsymbol{X}_t, t)\right]dt+\boldsymbol{\Sigma}d\boldsymbol{W}_t, \quad &\boldsymbol{X}^b_0=\boldsymbol{x}_0
    \end{align}
    Then, the Radon-Nikodym derivative satisfies
    \begin{small}
    \begin{align}
        \log\frac{\mathrm{d}\mathbb{P}^b}{\mathrm{d}\mathbb{P}^0}(\boldsymbol{X}_{0:T})=\int_0^T(\boldsymbol{b}^{\top}\boldsymbol{\Sigma}^{-1})(\boldsymbol{X}_t,t)d\boldsymbol{X}_t-\int_0^T(\boldsymbol{\Sigma}^{-1}\boldsymbol{f}\cdot \boldsymbol{b})(\boldsymbol{X}_t,t)dt-\frac{1}{2}\int_0^T\|\boldsymbol{b}(\boldsymbol{X}_t,t)\|^2dt\nonumber
    \end{align}
    \end{small}
    or equivalently:
    \begin{align}
        \log\frac{\mathrm{d}\mathbb{P}^b}{\mathrm{d}\mathbb{P}^0}(\boldsymbol{X}_{0:T})=\int_0^T\boldsymbol{b}(\boldsymbol{X}_t,t)^\top d\boldsymbol{W}_t-\frac{1}{2}\int_0^T\|\boldsymbol{b}(\boldsymbol{X}_t,t)\|^2dt
    \end{align}
\end{lemma}
\textit{Proof.} First, we use Girsanov's theorem to get the Radon-Nikodym derivative with respect to the zero-drift reference path measure $\mathbb{Q}$ defined by the SDE $d\boldsymbol{X}_t=\boldsymbol{\Sigma}(\boldsymbol{X}_t, t)d\boldsymbol{W}_t$, we have
\begin{small}
\begin{align}
    \log\frac{\mathrm{d}\mathbb{P}^0}{\mathrm{d}\mathbb{Q}}=\int_0^T\left(\boldsymbol{f}(\boldsymbol{X}_t, t)\cdot \boldsymbol{\Sigma}^{-2}(\boldsymbol{X}_t, t)\right)d\boldsymbol{X}_t-\frac{1}{2}\int_0^T\left(\boldsymbol{f}\cdot \boldsymbol{\Sigma}^{-2}\boldsymbol{f}\right)(\boldsymbol{X}_t, t)dt\label{app-eq:rnd-ref}
\end{align}
\end{small}
and for the controlled path measure $\mathbb{P}^b$, we have
\begin{small}
\begin{align}
    \log\frac{\mathrm{d}\mathbb{P}^b}{\mathrm{d}\mathbb{Q}}(\boldsymbol{X}_{0:T})&=\int_0^T(\boldsymbol{f}+\boldsymbol{\Sigma}\boldsymbol{b})(\boldsymbol{X}_t,t)\cdot \boldsymbol{\Sigma}^{-2}(\boldsymbol{X}_t,t)d\boldsymbol{X}_t\nonumber\\
    &-\frac{1}{2}\int_0^T\left((\boldsymbol{f}+\boldsymbol{\Sigma}\boldsymbol{b})\cdot \boldsymbol{\Sigma}^{-2}(\boldsymbol{f}+\boldsymbol{\Sigma}\boldsymbol{b})\right)(\boldsymbol{X}_t,t)dt\label{app-eq:rnd-control}
\end{align}
\end{small}
Then, using the identity
\begin{align}
\log \frac{\mathrm{d}\mathbb{P}^b}{\mathrm{d}\mathbb{P}^0}= \log \frac{\mathrm{d}\mathbb{P}^b}{\mathrm{d}\mathbb{Q}}\frac{\mathrm{d}\mathbb{Q}}{\mathrm{d}\mathbb{P}^0}=\log \frac{\mathrm{d}\mathbb{P}^b}{\mathrm{d}\mathbb{Q}}+\log \frac{\mathrm{d}\mathbb{Q}}{\mathrm{d}\mathbb{P}^0}=\log \frac{\mathrm{d}\mathbb{P}^b}{\mathrm{d}\mathbb{Q}}-\log \frac{\mathrm{d}\mathbb{P}^0}{\mathrm{d}\mathbb{Q}}
\end{align}

Substituting in (\ref{app-eq:rnd-ref}) and (\ref{app-eq:rnd-control}) and canceling terms, we get
\begin{small}
\begin{align}
    \log\frac{\mathrm{d}\mathbb{P}^b}{\mathrm{d}\mathbb{P}^0}(\boldsymbol{X}_{0:T})=\int_0^T(\boldsymbol{b}^{\top}\boldsymbol{\Sigma}^{-1})(\boldsymbol{X}_t,t)d\boldsymbol{X}_t-\int_0^T(\boldsymbol{\Sigma}^{-1}\boldsymbol{f}\cdot \boldsymbol{b})(\boldsymbol{X}_t,t)dt-\frac{1}{2}\int_0^T\|\boldsymbol{b}(\boldsymbol{X}_t,t)\|^2dt
\end{align}
\end{small}

Equivalently, since we can write $d\boldsymbol{W}_t=\boldsymbol{\Sigma}^{-1}(\boldsymbol{X}_t,t)(d\boldsymbol{X}_t-\boldsymbol{f}(\boldsymbol{X}_t,t)dt)$ (the Brownian motion under $\mathbb{P}^0$), we can write
\begin{align}
    \log\frac{\mathrm{d}\mathbb{P}^b}{\mathrm{d}\mathbb{P}^0}(\boldsymbol{X}_{0:T})=\int_0^T\boldsymbol{b}(\boldsymbol{X}_t,t)^\top d\boldsymbol{W}_t-\frac{1}{2}\int_0^T\|\boldsymbol{b}(\boldsymbol{X}_t,t)\|^2dt
\end{align}
which concludes the proof. \hfill $\square$

\begin{remark}
    In the under-damped Langevin model, the diffusion acts only on the velocity coordinates, so the full diffusion matrix on $\boldsymbol{X}_t=(\boldsymbol{R}_t, \boldsymbol{V}_t)$ is degenerate, i.e., $\boldsymbol{\Sigma}=\begin{bmatrix}
        0&0\\
        0&\boldsymbol{\Sigma}_V
    \end{bmatrix}$. We therefore apply Girsanov’s theorem to the velocity SDE only, where the diffusion $\boldsymbol{\Sigma}_V$ is invertible, and treat positions as deterministic integrals of velocity (i.e., $\boldsymbol{R}_t=\boldsymbol{R}_0+\int_0^t\boldsymbol{V}_sds$). Therefore, Girsanov's theorem is first defined on velocity paths as
    \begin{align}
        \log\frac{\mathrm{d}\mathbb{P}_V^b}{\mathrm{d}\mathbb{P}_V^0}(\boldsymbol{X}_{0:T})=\int_0^T\boldsymbol{b}(\boldsymbol{X}_t,t)^\top d\boldsymbol{W}_t-\frac{1}{2}\int_0^T\|\boldsymbol{b}(\boldsymbol{X}_t,t)\|^2dt
    \end{align}
    and pushed forward to the full path measure via $(\boldsymbol{R}_t, \boldsymbol{V}_t)=(\boldsymbol{R}_0+\int_0^t\boldsymbol{V}_sds, \boldsymbol{V}_t)$. For simplicity in notation, we define $\boldsymbol{\Sigma}^{-1}\equiv\boldsymbol{\Sigma}^{-1}_V$ in the theoretical proofs. 
\end{remark}

\paragraph{Stochastic Optimal Control}
The reward-guided \textbf{stochastic optimal control (SOC)} problem aims to determine the \textbf{optimal control} $\boldsymbol{b}^\star=\arg\min _{\boldsymbol{b}\in \mathcal{U}}J(\boldsymbol{b}; \boldsymbol{x}, t)$ that minimizes the \textbf{cost functional} $J(\boldsymbol{b}_\theta; \boldsymbol{x}, t)$ with reward $r:\mathcal{X}\to \mathbb{R}$ given by
\begin{align}
    J(\boldsymbol{b}; \boldsymbol{x}, t)&=\mathbb{E}_{p_t}\left[\int_t^T\left(\boldsymbol{f}(\boldsymbol{X}_s,s)+\frac{1}{2}\|\boldsymbol{b}(\boldsymbol{X}_s, s)\|^2\right)ds-r(\boldsymbol{X}_T)\bigg|\boldsymbol{X}_t=\boldsymbol{x}\right]\label{eq:cost-functional}
\end{align}
which measures the \textit{cost-to-go} from the current state $\boldsymbol{X}_t=\boldsymbol{x}$ at time $t$ under the controlled dynamics defined in (\ref{eq:control-SDE}) to the terminal state $\boldsymbol{X}_T$. The infimum of the cost functional is defined as
\begin{align}
    J(\boldsymbol{b}^\star; \boldsymbol{x}, t)=\inf_{\boldsymbol{b}\in \mathcal{U}}J(\boldsymbol{b};\boldsymbol{x}, t)=: V_t(\boldsymbol{x})
\end{align}
which is also referred to as the \textit{value function} $V_t(\boldsymbol{x})$. To derive the path integral representation of the optimal control $\boldsymbol{b}^\star$, we introduce the \textit{work functional}, which computes the running cost subtracted from the terminal reward of a state path $\boldsymbol{X}_{0:T}$,
\begin{align}
    \mathcal{W}(\boldsymbol{X}_{t:T}, t):=\int_t^T\boldsymbol{f}(\boldsymbol{X}_s, s)ds-r(\boldsymbol{X}_T)
\end{align}

We can now define a series of statements that connect the optimal drift $\boldsymbol{b}^\star$ and the value function $V_t$.
\begin{theorem}\label{theorem:soc}
    Let $V_t$ be the value function and $\boldsymbol{b}^\star$ be the optimal control. Then, the following are true:
    \begin{enumerate}
        \item[(a)] The optimal control satisfies $\boldsymbol{b}^\star(\boldsymbol{x})=-\boldsymbol{\Sigma}^\top \nabla_{\boldsymbol{x}}V_t(\boldsymbol{x})$
        \item[(b)] The Radon-Nikodym derivative of the optimal path measure $\mathbb{P}^\star$ can be defined by taking state paths from the unconditional path measure $\mathbb{P}^0$ and reweighting them with the output of the work functional $\mathcal{W}(\boldsymbol{X}_{0:T}, 0)$.
        \begin{align}
            \frac{\mathrm{d}\mathbb{P}^\star}{\mathrm{d}\mathbb{P}^0}(\boldsymbol{X}_{0:T})=\frac{1}{Z}e^{-\mathcal{W}(\boldsymbol{X}_{0:T}, 0)}, \quad Z=\mathbb{E}_{\boldsymbol{X}_{0:T}\sim \mathbb{P}^0}\left[e^{-\mathcal{W}(\boldsymbol{X}_{0:T}, 0)}\right]
        \end{align}
        \item[(c)] For any $(\boldsymbol{x}, t)\in \mathcal{X}\times [0,T]$, the value function can be written in path integral form by taking the expectation over the uncontrolled path measure $\mathbb{P}^0$ as 
        \begin{align}
            V_t(\boldsymbol{x})=-\log \mathbb{E}\left[e^{-\mathcal{W}(\boldsymbol{X}_{t:T}, t)}\big|\boldsymbol{X}_t=\boldsymbol{x}\right]
        \end{align}
        \item[(d)] Combining (a) and (c), we can write the optimal control as
        \begin{align}
            \boldsymbol{b}^\star(\boldsymbol{x})=\boldsymbol{\Sigma}^\top \nabla_{\boldsymbol{x}}\log \mathbb{E}_{\boldsymbol{X}_{t:T}\sim \mathbb{P}^0}\left[e^{-\mathcal{W}(\boldsymbol{X}_{t:T}, t )}\big|\boldsymbol{X}_t=\boldsymbol{x}\right]
        \end{align}
    \end{enumerate}
    
\end{theorem}
While Theorem \ref{theorem:soc} provides a closed-form solution to the optimal control $\boldsymbol{b}^\star$, it remains impractical to compute as $\mathbb{P}^0$ contains an infinite number of paths from any point $\boldsymbol{x}$. This problem motivates parameterizing $\boldsymbol{b}^\star$ with a neural network $\boldsymbol{b}_\theta$. The natural objective for obtaining an accurate approximation of $\boldsymbol{b}^\star$ that induces the path measure $\mathbb{P}^\star$ is the KL-divergence $\mathcal{L}(\mathbb{P}^\star, \mathbb{P}^{b_\theta}):=D_{\text{KL}}(\mathbb{P}^{b_\theta}\|\mathbb{P}^\star)$ as its minimizer is exactly $\mathbb{P}^\star$. 

However, taking the gradient $\nabla_\theta D_{\text{KL}}(\mathbb{P}^{b_\theta}\|\mathbb{P}^\star)$ requires differentiating through the full stochastic trajectories $\boldsymbol{X}_{0:T}\sim \mathbb{P}^{b_\theta}$ generated by the Euler-Maruyama SDE solver due to the expectation over $\mathbb{P}^{b_\theta}$, resulting in a significant computational bottleneck. To overcome this bottleneck, previous work \citep{nusken2021solving, seong2025transition} leverages alternative path-measure objectives that do not involve expectations over $\mathbb{P}^{b_\theta}$ while preserving optimality guarantees. We focus on two objectives: the \textbf{log-variance divergence} used in \citet{seong2025transition} and the \textbf{cross-entropy objective}, which we formalize in the present work for solving the EntangledSB problem. 

\paragraph{Log-Variance Objective}
Here, we describe the \textbf{log-variance (LV) divergence}, which is introduced in \citet{nusken2021solving} and applied to transition path sampling (TPS) in TPS-DPS \citep{seong2025transition}. Given the path measure $\mathbb{P}^{b_\theta}$ generated with the parameterized bias force $\boldsymbol{b}_\theta$ and the target path measure $\mathbb{P}^\star$, the log-variance divergence $\mathcal{L}_{\text{LV}}$ is defined as
\begin{align}
    \mathcal{L}_{\text{LV}}(\mathbb{P}^\star, \mathbb{P}^{b_\theta}): =\text{Var}_{\mathbb{P}^v}\left[\log \frac{\mathrm{d}\mathbb{P}^\star}{\mathrm{d}\mathbb{P}^{b_\theta}}\right]=\mathbb{E}_{\mathbb{P}^v}\left[\left(\log \frac{\mathrm{d}\mathbb{P}^\star}{\mathrm{d}\mathbb{P}^{b_\theta}}-\mathbb{E}_{\mathbb{P}^v}\left[\log \frac{\mathrm{d}\mathbb{P}^\star}{\mathrm{d}\mathbb{P}^{b_\theta}}\right]\right)^2\right]
\end{align}
where $\mathbb{P}^v$ can be defined as any arbitrary path measure generated from a control $\boldsymbol{v}$. To allow simulated trajectories to be used across multiple training iterations, it is common to define $\boldsymbol{v}$ as the bias force with frozen parameters $\boldsymbol{v}=\bar{\boldsymbol{b}}:=\texttt{stopgrad}(\boldsymbol{b}_\theta)$. Using Lemma \ref{lemma:RND}, we define $\mathcal{L}_{\text{LV}}(\mathbb{P}^\star, \mathbb{P}^{b_\theta})=\mathbb{E}_{\mathbb{P}^v}[\left(\mathcal{F}_{\boldsymbol{b}_\theta, \boldsymbol{v}}-\mathbb{E}_{\mathbb{P}^v}\left[\mathcal{F}_{\boldsymbol{b}_\theta, \boldsymbol{v}}\right]\right)^2]$, where $\mathcal{F}_{\boldsymbol{b}_\theta, \boldsymbol{v}}$ is defined as
\begin{small}
\begin{align}
    \mathcal{F}_{\boldsymbol{b}_\theta, \boldsymbol{v}}&:=\log\frac{\mathrm{d}\mathbb{P}^\star}{\mathrm{d}\mathbb{P}^{b_\theta}}=\log \frac{\mathrm{d}\mathbb{P}^\star}{\mathrm{d}\mathbb{P}^0}\frac{\mathrm{d}\mathbb{P}^0}{\mathrm{d}\mathbb{P}^{b_\theta}}\nonumber\\
    &=r(\boldsymbol{X}_T)-\log Z+\frac{1}{2}\int_0^T\|\boldsymbol{b}_\theta(\boldsymbol{X}_t)\|^2dt-\int_0^T(\boldsymbol{b}_\theta^\top\boldsymbol{v})(\boldsymbol{X}_t)dt- \int_0^T\boldsymbol{b}_\theta^\top d \boldsymbol{W}_t
\end{align}
\end{small}
using similar justification as in App \ref{appendix-prop:Cross-Entropy-2}. Since the expectation $\mathbb{E}_{\mathbb{P}^v}\left[\mathcal{F}_{\boldsymbol{b}_\theta, \boldsymbol{v}}\right]$ is computationally intractable, \citet{seong2025transition} introduces a scalar parameter $w$ that is jointly optimized with $\theta$ such that $\arg\min_w\mathcal{L}_{\text{LV}}(\theta, w)=\mathbb{E}_{\mathbb{P}^v}\left[\mathcal{F}_{\boldsymbol{b}_\theta, \boldsymbol{v}}\right]$. Substituting in $w$ for $\mathbb{E}_{\mathbb{P}^v}\left[\mathcal{F}_{\boldsymbol{b}_\theta, \boldsymbol{v}}\right]$ and taking the discretization over $K$ steps, the LV objective becomes
\begin{small}
\begin{align}
    \mathcal{L}_{\text{LV}}(\theta, w)=\mathbb{E}_{\boldsymbol{X}_{0:K}\sim \mathbb{P}^v}\left[\left(\mathcal{F}_{\boldsymbol{b}_\theta, \boldsymbol{v}}(\boldsymbol{X}_{0:K})-w\right)^2\right]=\mathbb{E}_{\boldsymbol{x}_{0:K}\sim \mathbb{P}^v}\left[\left(\log \frac{p^0(\boldsymbol{x}_{0:K})\exp(r(\boldsymbol{X}_K))}{p^{b_\theta}(\boldsymbol{x}_{0:K})}-w\right)^2\right]\nonumber
\end{align}
\end{small}
where $\boldsymbol{v}=\bar{\boldsymbol{b}}:=\texttt{stopgrad}(\boldsymbol{b}_\theta)$.

\paragraph{Regression Form and Why We Do Not Use It}
From Prop.~\ref{prop:Cross-Entropy-2}, $\mathcal{L}_{\text{CE}}$ differs only by a $\theta$-independent term from the regression objective
\begin{align}
    \mathcal{L}_{\text{L2}}(\theta)=\mathbb{E}_{\mathbb{P}^v}\left[w^\star(\boldsymbol{X}_{0:T})\frac{1}{2}\int_0^T\big\|\boldsymbol{b}_\theta(\boldsymbol{R}_t,\boldsymbol{V}_t)-\boldsymbol{b}^\star(\boldsymbol{R}_t,\boldsymbol{V}_t)\big\|^2dt\right]
\end{align}
which typically has lower variance. It is unusable here because $\boldsymbol{b}^\star=\boldsymbol{\Sigma}^\top\nabla\log h(\boldsymbol{x},t)$, where $h$ solves the backward Kolmogorov equation of the underdamped reference with terminal condition $\pi_\mathcal{B}$. Under a nonlinear force field, computing $h$ is as hard as the original problem. In first-order settings with Brownian or Ornstein-Uhlenbeck references, the analogous target is a closed-form Gaussian bridge drift, which is what makes the L2 form attractive there. $\mathcal{L}_{\text{CE}}$ needs no pointwise regression target and evaluates $r$ only at the endpoint. Our variance-reduced baseline is instead the log-variance divergence (App~\ref{app:LV-comparison}).

\subsection{Connection Between the Schr\"odinger Bridge Objective and Stochastic Optimal Control}
\label{app:sb_soc_connection}

The Schr\"{o}dinger Bridge (SB) problem stated in the main text is not written in its most standard form, but is an equivalent reformulation that is better suited to our setting. Similar formulations have appeared in prior work \citep{holdijk2023stochastic}. In this section, we make the connection between the standard SB problem, our reformulation, and the stochastic optimal control (SOC) objective explicit. For a broader treatment of the SB problem and its links to optimal transport and SOC, see \citet{tang2026foundations}.

\paragraph{Standard SB Problem to a Tilted Path Measure}
The standard SB problem is defined as
\begin{align}
\min_{\mathbb{P}} D_{\mathrm{KL}}\left(\mathbb{P}\| \mathbb{P}^0\right)
\quad \text{s.t.} \quad
\mathbb{P}_0 = \pi_{\mathcal{A}}, \quad \mathbb{P}_T = \pi_{\mathcal{B}}
\label{eq:standard_sb}
\end{align}
which seeks the path measure closest in KL to the reference $\mathbb{P}^0$ subject to \textit{both} marginal constraints. Its solution has the form $\frac{\mathrm{d}\mathbb{P}^{\mathrm{SB}}}{\mathrm{d}\mathbb{P}^0}(\boldsymbol{X}_{0:T})=f(\boldsymbol{X}_0)\,g(\boldsymbol{X}_T)$, where the Schr\"odinger potentials $(f,g)$ solve a coupled system that is intractable for nonlinear references. Following \citet{holdijk2023stochastic} and the SOC view of \citet{liu2025adjoint}, we instead fix $\boldsymbol{X}_0\sim\pi_\mathcal{A}$ and replace the hard terminal constraint by a terminal reward $r=\log\pi_\mathcal{B}$. The resulting optimal path measure is the conditional tilt
\begin{align}
\mathbb{P}^\star(\boldsymbol{X}_{0:T}) = \pi_\mathcal{A}(\boldsymbol{X}_0)\,\mathbb{P}^0(\boldsymbol{X}_{0:T}\mid\boldsymbol{X}_0)\,\frac{\pi_{\mathcal{B}}(\boldsymbol{R}_T)}{Z(\boldsymbol{X}_0)},\qquad Z(\boldsymbol{X}_0)=\mathbb{E}_{\mathbb{P}^0}\left[\pi_\mathcal{B}(\boldsymbol{R}_T)\mid\boldsymbol{X}_0\right]
\label{eq:optimal_path_measure}
\end{align}
which satisfies $\mathbb{P}^\star_0=\pi_\mathcal{A}$ exactly. Its terminal marginal equals $\pi_\mathcal{B}$ reweighted by the reference transition density, not $\pi_\mathcal{B}$ itself. The two coincide in special cases, e.g.\ for memoryless reference processes with a corrected terminal cost \citep{domingo2024adjoint, liu2025adjoint}, and they are close when $\pi_\mathcal{B}$ is sharply concentrated relative to the reference transition density, as in TPS where $\pi_\mathcal{B}$ is a narrow Gaussian around $\boldsymbol{R}_\mathcal{B}$. In general, (\ref{eq:optimal_path_measure}) is a soft-constrained Schr\"odinger bridge \citep{garg2024soft}, and this is the problem EntangledSBM solves.

\paragraph{Reformulation as an Unconstrained KL Minimization}
Let $\mathbb{P}^{b_\theta}$ denote the path measure induced by the controlled Langevin SDEs in (\ref{eq:multi-langevin}) of the main text. Substituting (\ref{eq:optimal_path_measure}) into the KL divergence between $\mathbb{P}^{b_\theta}$ and the optimal path measure $\mathbb{P}^\star$ yields:
\begin{align}
D_{\mathrm{KL}}\left(\mathbb{P}^{b_\theta} \| \mathbb{P}^\star\right)
= D_{\mathrm{KL}}\left(\mathbb{P}^{b_\theta} \| \mathbb{P}^0\right)
- \mathbb{E}_{\mathbb{P}^{b_\theta}}\left[\log \pi_{\mathcal{B}}(\boldsymbol{R}_T)\right] + \mathbb{E}_{\pi_\mathcal{A}}\left[\log Z(\boldsymbol{X}_0)\right]
\label{eq:tilted_kl}
\end{align}

Because the base path measure (induced by the uncontrolled SDE in (\ref{eq:langevin-dynamics})) and the controlled path measure $\mathbb{P}^{b_\theta}$ (\ref{eq:multi-langevin}) differ only by the learned control $\boldsymbol{b}_\theta$, we may apply Girsanov's theorem to express $D_{\mathrm{KL}}(\mathbb{P}^{b_\theta} \| \mathbb{P}^0)$ as a quadratic running cost in $\boldsymbol{b}_\theta$, recovering the form stated in Proposition \ref{prop:EntangledSOC}.

\paragraph{Connection to Stochastic Optimal Control (SOC)}
(\ref{eq:tilted_kl}) has a direct interpretation as a SOC problem. Defining the terminal reward as:

\begin{align}
r(\boldsymbol{X}_T) := \log \pi_{\mathcal{B}}(\boldsymbol{R}_T)
\end{align}

with the log-density of the target distribution, the objective decomposes into a running control cost and a terminal reward, and its optimal solution coincides exactly with the tilted SB measure: $\mathbb{P}^{b^\star} = \mathbb{P}^\star$. Equivalently, the optimal control $\boldsymbol{b}^\star$ steers the base process toward high-probability regions of $\pi_{\mathcal{B}}$ while incurring minimal deviation, in the KL sense, from the reference dynamics.

Although (\ref{eq:tilted_kl}) is not the textbook statement of the SB problem, this SOC-style formulation has been used in prior work \citep[see, e.g.,][]{holdijk2023stochastic} and recasts the constrained problem in (\ref{eq:standard_sb}) as an \textit{unconstrained} optimization over per-trajectory path KL and terminal log-probability. This unconstrained form is the foundation of our practical algorithm: it admits a tractable trajectory-level Monte Carlo estimator and avoids the need to enforce the terminal marginal constraint explicitly during training.

\paragraph{Requirements on $\pi_\mathcal{B}$} EntangledSBM never samples the terminal marginal of the bridge. It only needs to evaluate $\log\pi_\mathcal{B}$ and its gradient pointwise, up to an additive constant. The constant cancels both in the score $\nabla_{\boldsymbol{r}^i_t}\log\pi_\mathcal{B}$ (Sec~\ref{sec:Sampling Direction}) and in the self-normalized importance weights (Sec~\ref{sec:Cross-Entropy Objective}). This covers \textbf{(i)} transition path sampling, where $\pi_\mathcal{B}$ is a Gaussian around a target structure (App~\ref{app:tps-experiment}), \textbf{(ii)} bridges between data distributions, where $\pi_\mathcal{B}$ is a Gaussian smoothing of target samples (App~\ref{app:cell-experiment}), and \textbf{(iii)} Boltzmann or reward-tilted targets $\pi_\mathcal{B}\propto e^{-U}$ or $\pi_\mathcal{B}\propto e^{r}$. When the target is known only through samples, as in our cell experiments, we construct $\pi_\mathcal{B}$ by Gaussian smoothing of the samples. The learned bridge then matches this smoothed density rather than the true target distribution, and the approximation degrades in high dimensions or with few samples.

\subsection{Justification for Entangled Bias Forces}
\begin{proposition}[Monotone Optimality of Entangled Bias Forces]
    Let $\mathcal{F}_{\text{ind}}$ be the hypothesis class of bias forces that depend only on particle positions $\boldsymbol{b}^i_\theta:=\boldsymbol{b}_\theta^i(\boldsymbol{R}_t)$
    and $\mathcal{F}_{\text{ent}}$ be the hypothesis class of entangled bias forces $\boldsymbol{b}^i_\theta:=\boldsymbol{b}_\theta^i(\boldsymbol{R}_t, \boldsymbol{V}_t)$. Then, under the cross-entropy objective,
    \begin{align}
        \inf_{\boldsymbol{b}_\theta\in \mathcal{F}_{\text{ent}}}\mathcal{L}_{\text{CE}}(\boldsymbol{b}_\theta)\leq \inf_{\boldsymbol{b}_\theta\in \mathcal{F}_{\text{ind}}}\mathcal{L}_{\text{CE}}(\boldsymbol{b}_\theta)
    \end{align}
    with strict improvement when no position-only bias force can realize the optimal path measure.
\end{proposition}
\textit{Proof.} Let $\mathbb{P}^\star$ denote the optimal path measure that solves the EntangledSB problem from Definition~\ref{def:EntangledSB} and $\mathbb{P}^{b_\theta}$ be the path measure induced by the bias force $\boldsymbol{b}_\theta$. Let $\mathcal{F}_{\text{ind}}$ be the hypothesis class of bias forces independent of the full system velocities $\boldsymbol{b}^i_\theta:=\boldsymbol{b}_\theta^i(\boldsymbol{R}_t)$ and $\mathcal{F}_{\text{ent}}$ be the hypothesis class of entangled bias forces $\boldsymbol{b}^i_\theta:=\boldsymbol{b}_\theta^i(\boldsymbol{R}_t, \boldsymbol{V}_t)$. Clearly, we have $\mathcal{F}_{\text{ind}}\subseteq\mathcal{F}_{\text{ent}}$. It follows that
\begin{align}
    \{\mathbb{P}^{b_\theta}:\boldsymbol{b}_\theta\in\mathcal{F }_{\text{ind}}\}\subseteq \{\mathbb{P}^{b_\theta}:\boldsymbol{b}_\theta\in\mathcal{F }_{\text{ent}}\}
\end{align}

Taking the infimum over the KL-divergence functional $D_{\text{KL}}(\mathbb{P}^\star\|\cdot)\geq 0$ over a larger set of functions cannot increase the value.
\begin{align}
    \inf_{\boldsymbol{b}_\theta\in \mathcal{F}_{\text{ent}}}D_{\text{KL}}(\mathbb{P}^\star\|\mathbb{P}^{b_\theta})\leq \inf_{\boldsymbol{b}_\theta\in \mathcal{F}_{\text{ind}}}D_{\text{KL}}(\mathbb{P}^\star\|\mathbb{P}^{b_\theta})
\end{align}

Furthermore, if $\mathbb{P}^\star\notin\overline{\{\mathbb{P}^{b_\theta}:\boldsymbol{b}_\theta\in\mathcal{F }_{\text{ind}}\}}$ which denotes the closure of the set of path measures induced by $\mathcal{F }_{\text{ind}}$ and there exists $\boldsymbol{b}^\star\in \overline{\mathcal{F}_{\text{ent}}}$ where $\mathbb{P}^{\boldsymbol{b}^\star} =\mathbb{P}^\star$, then we have
\begin{align}
    \inf_{\boldsymbol{b}_\theta\in \mathcal{F}_{\text{ent}}}D_{\text{KL}}(\mathbb{P}^\star\|\mathbb{P}^{b_\theta})=0 \quad \text{while}\quad \inf_{\boldsymbol{b}_\theta\in \mathcal{F}_{\text{ind}}}D_{\text{KL}}(\mathbb{P}^\star\|\mathbb{P}^{b_\theta})>0
\end{align}
yielding strict improvement. By Proposition \ref{prop:Cross-Entropy}, we have that $\mathbb{P}^{b^\star}=\mathbb{P}^\star$ is the unique minimizer of $D_{\text{KL}}(\mathbb{P}^\star\|\cdot)$.\hfill $\square$

\section{Theoretical Proofs}
\label{app:Theoretical Proofs}
For notational simplicity, we will drop the explicit dependence on $(\boldsymbol{R}_t, \boldsymbol{V}_t)$ and simply denote the stochastic path of the system as $(\boldsymbol{X}_t)_{t\in [0,T]}$.

\subsection{Proof of Proposition \ref{prop:EntangledSOC}}
\label{appendix-prop:EntangledSOC}

\begin{lemma}\label{lemma:RND-optimal}
     The bias force $\boldsymbol{b}^\star$ that minimizes the SOC objective $J(\boldsymbol{b}_\theta; \boldsymbol{x}, t)$ generates the path measure $\mathbb{P}^{b}$ that minimizes the KL-divergence with the optimal tilted path measure $\mathbb{P}^\star$ that satisfies
     \begin{align}
         \mathbb{P}^\star=\frac{1}{Z}\mathbb{P}^0e^{r(\boldsymbol{X}_T)}, \quad Z=\mathbb{E}_{\mathbb{P}^0_T}\left[e^{r(\boldsymbol{X}_T)}\right]
     \end{align}
     where $\mathbb{P}^0$ is the base path measure.
\end{lemma}

\textit{Proof.} Let the dynamics of $\boldsymbol{X}_t=(\boldsymbol{R}_t, \boldsymbol{V}_t)$ under the base path measure $\mathbb{P}^0$ be defined as
\begin{align}
    d\boldsymbol{X}_t=\boldsymbol{f}(\boldsymbol{X}_t)dt+\boldsymbol{\Sigma} d\boldsymbol{W}_t
\end{align}
and the dynamics under the biased path measure $\mathbb{P}^{b_\theta}$ be defined as
\begin{align}
    d\boldsymbol{X}_t=\left[\boldsymbol{f}(\boldsymbol{X}_t)+ \boldsymbol{\Sigma} \boldsymbol{b}_\theta(\boldsymbol{X}_t)\right]dt+\boldsymbol{\Sigma} d\boldsymbol{W}_t
\end{align}
By Lemma \ref{lemma:RND}, we have that the logarithm of the Radon-Nikodym derivative of the biased measure $\mathbb{P}^{b_\theta}$ with respect to the base measure $\mathbb{P}^0$ is given by
\begin{align}
    \log \frac{\mathrm{d}\mathbb{P}^{b_\theta}}{\mathrm{d}\mathbb{P}^0}(\boldsymbol{X}_{0:T})=\int_0^T\boldsymbol{b}_\theta(\boldsymbol{X}_t)^\top  d\boldsymbol{W}_t-\frac{1}{2}\int_0^T\|\boldsymbol{b}_\theta(\boldsymbol{X}_t)\|^2dt
\end{align}
Now, taking the expectation with respect to $\mathbb{P}^{b_\theta}$, we derive the expression for the KL-divergence
\begin{align}
    D_{\text{KL}}(\mathbb{P}^{b_\theta}\|\mathbb{P}^0)&=\mathbb{E}_{\mathbb{P}^{b_\theta}}\left[\log \frac{\mathrm{d}\mathbb{P}^{b_\theta}}{\mathrm{d}\mathbb{P}^0}\right]\nonumber\\
    &=\mathbb{E}_{\mathbb{P}^{b_\theta}}\left[\int_0^T\boldsymbol{b}_\theta(\boldsymbol{X}_t)^\top d\boldsymbol{W}_t-\frac{1}{2}\int_0^T\|\boldsymbol{b}_\theta(\boldsymbol{X}_t)\|^2dt\right]\nonumber\\
    &=\mathbb{E}_{\mathbb{P}^{b_\theta}}\left[\int_0^T\boldsymbol{b}_\theta(\boldsymbol{X}_t)^\top (d\boldsymbol{W}_t^b+\boldsymbol{b}_\theta(\boldsymbol{X}_t)dt)-\frac{1}{2}\int_0^T\|\boldsymbol{b}_\theta(\boldsymbol{X}_t)\|^2dt\right]\nonumber\\
    &=\underbrace{\mathbb{E}_{\mathbb{P}^{b_\theta}}\left[\int_0^T\boldsymbol{b}_\theta(\boldsymbol{X}_t)^\top d\boldsymbol{W}_t^b\right]}_{=0}+\mathbb{E}_{\mathbb{P}^{b_\theta}}\left[\int_0^T\|\boldsymbol{b}_\theta(\boldsymbol{X}_t)\|^2dt-\frac{1}{2}\int_0^T\|\boldsymbol{b}_\theta(\boldsymbol{X}_t)\|^2dt\right]\nonumber\\
    &=\frac{1}{2}\mathbb{E}_{\mathbb{P}^{b_\theta}}\int _0^T\|\boldsymbol{b}_\theta(\boldsymbol{X}_t)\|^2dt
\end{align}
Substituting this into the SOC objective in Proposition~\ref{prop:EntangledSOC}, we get
\begin{align}
    J(\boldsymbol{b}_\theta)&=\mathbb{E}_{\mathbb{P}^{b_\theta}}\left[\int_0^T\frac{1}{2}\|\boldsymbol{b}_\theta(\boldsymbol{X}_t)\|^2dt-r(\boldsymbol{X}_T)\right]\nonumber\\
    &=D_{\text{KL}}(\mathbb{P}^{b_\theta}\|\mathbb{P}^0)-\mathbb{E}_{\mathbb{P}^{b_\theta}}[r(\boldsymbol{X}_T)]
\end{align}
By adding $\log Z$ to both sides, we do not change the minimizer of $J(\boldsymbol{b}_\theta)$, and the SOC objective becomes the KL-divergence between the controlled path measure $\mathbb{P}^{b_\theta}$ and the optimal path measure $\mathbb{P}^\star$:
\begin{align}
    J(\boldsymbol{b}_\theta)+\log Z&=D_{\text{KL}}(\mathbb{P}^{b_\theta}\|\mathbb{P}^0)-\mathbb{E}_{\mathbb{P}^{b_\theta}}[r(\boldsymbol{X}_T)]+\log Z\nonumber\\
    &=\underbrace{\mathbb{E}_{\mathbb{P}^{b_\theta}}\left[\log \frac{\mathrm{d}\mathbb{P}^{b_\theta}}{\mathrm{d}\mathbb{P}^0}-r(\boldsymbol{X}_T)+\log Z\right]}_{D_{\text{KL}}(\mathbb{P}^{b_\theta}\|\mathbb{P}^\star)}
\end{align}
This shows that the minimizer $\boldsymbol{b}^\star$ to the SOC objective also generates the path measure $\mathbb{P}^{b^\star}$ that is closest in KL-divergence to the optimal path measure $\mathbb{P}^\star$. Therefore, the optimal control solution satisfies
\begin{align}
    \log\frac{\mathrm{d}\mathbb{P}^\star}{\mathrm{d}\mathbb{P}^0}(\boldsymbol{X}_{0:T})=r(\boldsymbol{X}_T)-\log Z\iff \mathbb{P}^\star=\frac{1}{Z}\mathbb{P}^0e^{r(\boldsymbol{X}_T)}
\end{align}
which concludes our proof. \hfill $\square$

\begin{tcolorbox}[sharp corners, colback=gray!10, boxrule=0pt]
  \entangledsoc*  
\end{tcolorbox}

We aim to recover the optimal path measure $\mathbb{P}^\star$ that minimizes the KL divergence from the base dynamics $\mathbb{P}^0$ while matching the terminal distribution $\pi_\mathcal{B}$ defined as
\begin{align}
    \mathbb{P}^\star(\boldsymbol{X}_{0:T})=\frac{1}{Z}\mathbb{P}^0\pi_\mathcal{B}(\boldsymbol{R}_{T})
\end{align}  
From Lemma \ref{lemma:RND-optimal}, we have that the bias force $\boldsymbol{b}_\theta$ that minimizes the SOC objective defined as
\begin{align}
    \boldsymbol{b}^\star=\arg\min_{\boldsymbol{b}_\theta}\mathbb{E}_{\boldsymbol{X}_{0:T}\sim \mathbb{P}^{b_\theta}}\left[\int_0^T\frac{1}{2}\|\boldsymbol{b}_\theta(\boldsymbol{X}_t)\|^2dt-r(\boldsymbol{X}_T)\right]
\end{align}
also generates the path measure $\mathbb{P}^{b^\star}$ that minimizes the KL-divergence $D_{\text{KL}}(\mathbb{P}^{b_\theta}\|\mathbb{P}^\star)$ with the optimal path measure $\mathbb{P}^\star$ defined as
\begin{align}
    \mathbb{P}^\star=\frac{1}{Z}\mathbb{P}^0e^{r(\boldsymbol{X}_T)}
\end{align}

Therefore, setting $r(\boldsymbol{X}_T):=\log \pi_\mathcal{B}(\boldsymbol{R}_T)$ recovers the solution to the EntangledSB problem. \hfill $\square$

\subsection{Proof of Proposition \ref{prop:Non-Increasing Distance}}
\label{appendix-prop:Non-Increasing Distance}

\begin{tcolorbox}[sharp corners, colback=gray!10, boxrule=0pt]
    \nonincreasing*
\end{tcolorbox}

\textit{Proof.} We aim to show that our parameterization of the bias force in (\ref{eq:bias-force}) ensures that the distance between the current position $\boldsymbol{R}_t$ and some target state $\boldsymbol{R}_T\sim \pi_\mathcal{B}$ in the target distribution is \textit{non-increasing}. For each atom $i\in \{1, \dots, n\}$, we define the bias force as
\begin{align}
    \boldsymbol{b}_\theta^i:=\alpha_\theta^i\hat{\boldsymbol{s}}_i +\left(\boldsymbol{I}-\hat{\boldsymbol{s}}_i\hat{\boldsymbol{s}}_i^\top\right)\boldsymbol{h}_\theta^i, 
    \qquad \alpha^i_\theta\geq 0,\;\; \hat{\boldsymbol{s}}_i=\frac{\nabla_{\boldsymbol{r}_t^i}\log \pi_{\mathcal{B}}}{\|\nabla_{\boldsymbol{r}_t^i}\log \pi_{\mathcal{B}}\|}
\end{align}

After the update in the direction of the potential energy $-\nabla_{\boldsymbol{r}_t^i}U(\boldsymbol{R}_t)$, the Euler update to the position with time step $\Delta t$ using the bias force is
\begin{align}
    \boldsymbol{r}^i_{t+\Delta t}
    =\boldsymbol{r}^i_t+\frac{\Delta t}{m_i}\left(\alpha^i_\theta\hat{\boldsymbol{s}}_i +\left(\boldsymbol{I}-\hat{\boldsymbol{s}}_i\hat{\boldsymbol{s}}_i^\top\right)\boldsymbol{h}^i_\theta\right)= \boldsymbol{r}^i_t+\frac{\Delta t}{m_i}\boldsymbol{b}_\theta^i
\end{align}
Let $\boldsymbol{r}^i_\mathcal{B}\in\mathrm{supp}(\pi_\mathcal{B})$ be a target point chosen on the ascent ray of $\hat{\boldsymbol{s}}_i$ given by
\begin{align}
    \boldsymbol{r}^i_\mathcal{B}\in\Big\{\boldsymbol{r}^i_t+\lambda\,\hat{\boldsymbol{s}}_i:\;\lambda\ge 0\Big\}\cap \mathrm{supp}(\pi_\mathcal{B})
\end{align}

We denote displacement to this target as $\boldsymbol{d}_i:=\boldsymbol{r}^i_\mathcal{B}-\boldsymbol{r}^i_t=\rho_i\,\hat{\boldsymbol{d}}_i$ with $\hat{\boldsymbol{d}}_i:=\frac{\boldsymbol{d}_i}{\|\boldsymbol{d}_i\|}$ and $\rho_i:=\|\boldsymbol{d}_i\|\ge 0$. By construction of the ray, $\hat{\boldsymbol{d}}_i=\hat{\boldsymbol{s}}_i$ (i.e., $\boldsymbol{d}_i$ is parallel with $\hat{\boldsymbol{s}}_i$). After a single step, the squared distance is given by
\begin{align}
    \|\boldsymbol{r}^i_\mathcal{B}-\boldsymbol{r}^i_{t+\Delta t}\|^2
    &=\left\|\boldsymbol{r}^i_\mathcal{B}-\left(\boldsymbol{r}^i_t+\frac{\Delta t}{m_i}\boldsymbol{b}_\theta^i\right)\right\|^2\nonumber\\
    &=\left\|\boldsymbol{d}_i-\frac{\Delta t}{m_i}\boldsymbol{b}_\theta^i\right\|^2\nonumber\\
    &=\|\boldsymbol{d}_i\|^2-\frac{2\Delta t}{m_i}\big\langle \boldsymbol{d}_i,\boldsymbol{b}^i_\theta\big\rangle+\frac{\Delta t^2}{m_i^2}\|\boldsymbol{b}^i_\theta\|^2 \label{eq:dist-expansion}
\end{align}
Using the decomposition of $\boldsymbol{b}^i_\theta$ and the fact that $\boldsymbol{d}_i\parallel \hat{\boldsymbol{s}}_i$ and $(\boldsymbol{I}-\hat{\boldsymbol{s}}_i\hat{\boldsymbol{s}}_i^\top)\boldsymbol{h}^i_\theta\perp \hat{\boldsymbol{s}}_i$, we obtain
\begin{align}
\big\langle \boldsymbol{d}_i,\boldsymbol{b}^i_\theta\big\rangle
&=\alpha^i_\theta\,\big\langle \boldsymbol{d}_i,\hat{\boldsymbol{s}}_i\big\rangle
+\big\langle \boldsymbol{d}_i,\underbrace{(\boldsymbol{I}-\hat{\boldsymbol{s}}_i\hat{\boldsymbol{s}}_i^\top)\boldsymbol{h}^i_\theta}_{\perp\,\hat{\boldsymbol{s}}_i}\big\rangle
=\alpha^i_\theta\,\big\langle \rho_i\hat{\boldsymbol{s}}_i,\hat{\boldsymbol{s}}_i\big\rangle
=\alpha^i_\theta\rho_i\geq  0\label{eq:first-order-term}
\end{align}
Thus, the first-order term in (\ref{eq:dist-expansion}) always decreases (or preserves) the distance, and the orthogonal correction cannot increase it at first order because it is orthogonal to $\boldsymbol{d}_i$. When $\Delta t\to 0$, the second-order term becomes negligible, and the first-order term dominates. We determine the exact threshold for $\Delta t$ that guarantees a non-increasing distance to be
\begin{align}
    \frac{2\Delta t}{m_i}\alpha_\theta^i\rho_i\geq \frac{\Delta t^2}{m_i^2}\|\boldsymbol{b}^i_\theta\|^2\implies \Delta t\leq  \frac{2m_i\alpha_\theta^i\rho_i}{\|\boldsymbol{b}_\theta^i\|^2}
\end{align}
under which the decrease due to the first-order term dominates the quadratic increase, yielding
\begin{align}
    \|\boldsymbol{r}^i_{t+\Delta t}-\boldsymbol{r}^i_\mathcal{B}\|\leq \|\boldsymbol{r}^i_{t}-\boldsymbol{r}^i_\mathcal{B}\|
\end{align}

In the continuous-time limit as $\Delta t\to 0$, the non-increasing guarantee follows immediately from (\ref{eq:first-order-term}):
\begin{align}
    \frac{\partial}{\partial t}\|\boldsymbol{r}^i_\mathcal{B}-\boldsymbol{r}^i_t\|^2
=\frac{\partial}{\partial t}\|\boldsymbol{d}_i\|^2=-\frac{2}{m_i}\big\langle \boldsymbol{d}_i,\boldsymbol{b}^i_\theta\big\rangle
=-\frac{2}{m_i}\alpha^i_\theta\rho_i\leq 0
\end{align}
Summing over $i=1,\dots,n$ (or equivalently using the concatenated coordinates), we conclude that there exists a target configuration $\boldsymbol{R}_\mathcal{B}\in\mathrm{supp}(\pi_\mathcal{B})$ such that the distance to $\boldsymbol{R}_\mathcal{B}$ is \emph{non-increasing} under the bias force update, which proves the claim.
\hfill$\square$

\subsection{Proof of Proposition \ref{prop:Cross-Entropy}}
\label{appendix-prop:Cross-Entropy}

\begin{tcolorbox}[sharp corners, colback=gray!10, boxrule=0pt]
\crossentropy*
\end{tcolorbox}

\textit{Proof.} First, we show that the cross-entropy objective is convex in the path measure $\mathbb{P}^{b_\theta}$. The functional $\mathbb{Q}\mapsto D_{\text{KL}}(\mathbb{P}^\star\|\mathbb{Q})$ is convex on the space of probability measures, since $x\mapsto -\log x$ is convex. The set of controlled path measures $\{\mathbb{P}^b\}$ need not be convex, so this statement concerns the objective as a functional of the measure. Since $D_{\text{KL}}(\mathbb{P}^\star\|\mathbb{Q})\geq 0$, with equality if and only if $\mathbb{Q}=\mathbb{P}^\star$, the minimizer of $\mathcal{L}_{\text{CE}}$ is unique and equals $\mathbb{P}^{b^\star}\equiv\mathbb{P}^\star$. From Proposition \ref{prop:EntangledSOC}, we have that the unique minimizer of $\mathcal{L}_{\text{CE}}$ is exactly the solution to the EntangledSB problem. While this does not necessarily imply convexity in the neural network parameters $\theta$, it yields a more favorable optimization objective. \hfill $\square$

\subsection{Proof of Proposition \ref{prop:Cross-Entropy-2}}
\label{appendix-prop:Cross-Entropy-2}

\begin{tcolorbox}[sharp corners, colback=gray!10, boxrule=0pt]
\crossentropytwo*
\end{tcolorbox}

From the cross-entropy objective, we have
\begin{align}
    \mathcal{L}_{\text{CE}}(\mathbb{P}^\star, \mathbb{P}^{b_\theta})&:=D_{\text{KL}}(\mathbb{P}^\star\|\mathbb{P}^{b_\theta})=\mathbb{E}_{\mathbb{P}^\star}\left[\log \frac{\mathrm{d}\mathbb{P}^\star}{\mathrm{d}\mathbb{P}^{b_\theta}}\right]=\mathbb{E}_{\mathbb{P}^{\bar{b}}}\left[\frac{\mathrm{d}\mathbb{P}^\star}{\mathrm{d}\mathbb{P}^{\bar{b}}}\log \frac{\mathrm{d}\mathbb{P}^\star}{\mathrm{d}\mathbb{P}^{b_\theta}}\right]
\end{align}
Setting $w^\star:=\frac{\mathrm{d}\mathbb{P}^\star}{\mathrm{d}\mathbb{P}^{\bar{b}}}$, we can derive 
\begin{align}
    w^\star(\boldsymbol{X}_{0:T})=\frac{\mathrm{d}\mathbb{P}^\star}{\mathrm{d}\mathbb{P}^{\bar{b}}}(\boldsymbol{X}_{0:T})=\frac{\mathrm{d}\mathbb{P}^\star}{\mathrm{d}\mathbb{P}^0}\frac{\mathrm{d}\mathbb{P}^0}{\mathrm{d}\mathbb{P}^{\bar{b}}}(\boldsymbol{X}_{0:T})
\end{align}
From Lemma \ref{lemma:RND-optimal}, we have
\begin{align}
    w^\star(\boldsymbol{X}_{0:T})=\frac{e^{r(\boldsymbol{X}_T)}}{Z}\frac{\mathrm{d}\mathbb{P}^0}{\mathrm{d}\mathbb{P}^{\bar{b}}}(\boldsymbol{X}_{0:T})
\end{align}
which is our definition for $w^\star$. Now, expanding $\log \frac{\mathrm{d}\mathbb{P}^\star}{\mathrm{d}\mathbb{P}^{b_\theta}}$, we have 
\begin{align}
    \log \frac{\mathrm{d}\mathbb{P}^\star}{\mathrm{d}\mathbb{P}^{b_\theta}}&=\log \frac{\mathrm{d}\mathbb{P}^\star }{\mathrm{d}\mathbb{P}^{0}}+\log \frac{\mathrm{d}\mathbb{P}^0 }{\mathrm{d}\mathbb{P}^{b_\theta}}\nonumber\\
    &=\log \frac{e^{r(\boldsymbol{X}_T)}}{Z}-\log \frac{\mathrm{d}\mathbb{P}^{b_\theta}}{\mathrm{d}\mathbb{P}^{0}}\nonumber\\
    &=r(\boldsymbol{X}_T)-\log Z -\log \frac{\mathrm{d}\mathbb{P}^{b_\theta}}{\mathrm{d}\mathbb{P}^{0}}
\end{align}
Applying Girsanov's theorem from Lemma \ref{lemma:RND}, we expand the second term $\log \frac{\mathrm{d}\mathbb{P}^{b_\theta}}{\mathrm{d}\mathbb{P}^0}$ as
\begin{small}
\begin{align}
    \log \frac{\mathrm{d}\mathbb{P}^{b_\theta}}{\mathrm{d}\mathbb{P}^0}(\boldsymbol{X}_{0:T})&=\int_0^T(\boldsymbol{b}_\theta(\boldsymbol{X}_t)^\top\boldsymbol{\Sigma}^{-1})d\boldsymbol{X}_t-\int_0^T(\boldsymbol{\Sigma}^{-1}\boldsymbol{f}(\boldsymbol{X}_t))^\top \boldsymbol{b}_\theta(\boldsymbol{X}_t)dt-\frac{1}{2}\int_0^T\|\boldsymbol{b}_\theta(\boldsymbol{X}_t)\|^2dt\label{eq:prop-girsanov}
\end{align}
\end{small}
Given that $d\boldsymbol{X}_t$ evolves via the SDE
\begin{align}
    d\boldsymbol{X}_t&=(\boldsymbol{f}(\boldsymbol{X}_t)+ \boldsymbol{\Sigma}\bar{\boldsymbol{b}}(\boldsymbol{X}_t))dt+ \boldsymbol{\Sigma}d\boldsymbol{W}_t
\end{align}
Applying $(\boldsymbol{b}_\theta^\top\boldsymbol{\Sigma}^{-1})$ to both sides of the equation, we have
\begin{align}
    (\boldsymbol{b}_\theta^\top\boldsymbol{\Sigma}^{-1})d\boldsymbol{X}_t&=(\boldsymbol{b}_\theta^\top\boldsymbol{\Sigma}^{-1})(\boldsymbol{f}(\boldsymbol{X}_t)+ \boldsymbol{\Sigma}\bar{\boldsymbol{b}}(\boldsymbol{X}_t))dt+(\boldsymbol{b}_\theta^\top\boldsymbol{\Sigma}^{-1})\boldsymbol{\Sigma}d\boldsymbol{W}_t\nonumber\\
    &=\boldsymbol{b}_\theta^\top(\boldsymbol{\Sigma}^{-1}\boldsymbol{f})(\boldsymbol{X}_t)dt+ (\boldsymbol{b}_\theta^\top\boldsymbol{\Sigma}^{-1}\boldsymbol{\Sigma}\bar{\boldsymbol{b}})(\boldsymbol{X}_t)dt+ (\boldsymbol{b}_\theta^\top\boldsymbol{\Sigma}^{-1}\boldsymbol{\Sigma})d\boldsymbol{W}_t\nonumber\\
    &=(\boldsymbol{\Sigma}^{-1}\boldsymbol{f}(\boldsymbol{X}_t))^\top\boldsymbol{b}_\theta(\boldsymbol{X}_t)dt+(\boldsymbol{b}_\theta^\top\bar{\boldsymbol{b}})(\boldsymbol{X}_t)dt+ \boldsymbol{b}_\theta^\top d \boldsymbol{W}_t
\end{align}
Substituting this into (\ref{eq:prop-girsanov}), we get
\begin{small}
\begin{align}
    \log \frac{\mathrm{d}\mathbb{P}^{b_\theta}}{\mathrm{d}\mathbb{P}^0}(\boldsymbol{X}_{0:T})&=\int_0^T(\boldsymbol{b}_\theta(\boldsymbol{X}_t)^\top\boldsymbol{\Sigma}^{-1})d\boldsymbol{X}_t-\int_0^T(\boldsymbol{\Sigma}^{-1}\boldsymbol{f}(\boldsymbol{X}_t))^\top \boldsymbol{b}_\theta(\boldsymbol{X}_t)dt-\frac{1}{2}\int_0^T\|\boldsymbol{b}_\theta(\boldsymbol{X}_t)\|^2dt\nonumber\\
    &=\int_0^T\bigg((\boldsymbol{\Sigma}^{-1}\boldsymbol{f}(\boldsymbol{X}_t))^\top\boldsymbol{b}_\theta(\boldsymbol{X}_t)dt+(\boldsymbol{b}_\theta^\top\bar{\boldsymbol{b}})(\boldsymbol{X}_t)dt+ \boldsymbol{b}_\theta^\top d \boldsymbol{W}_t\bigg)\nonumber\\
    &-\int_0^T(\boldsymbol{\Sigma}^{-1}\boldsymbol{f}(\boldsymbol{X}_t))^\top \boldsymbol{b}_\theta(\boldsymbol{X}_t)dt-\frac{1}{2}\int_0^T\|\boldsymbol{b}_\theta(\boldsymbol{X}_t)\|^2dt\nonumber\\
    &=\int _0^T(\boldsymbol{\Sigma}^{-1}\boldsymbol{f}(\boldsymbol{X}_t))^\top\boldsymbol{b}_\theta(\boldsymbol{X}_t)dt+ \int_0^T(\boldsymbol{b}_\theta^\top\bar{\boldsymbol{b}})(\boldsymbol{X}_t)dt+ \int_0^T\boldsymbol{b}_\theta^\top d \boldsymbol{W}_t\nonumber\\
    &-\int_0^T(\boldsymbol{\Sigma}^{-1}\boldsymbol{f}(\boldsymbol{X}_t))^\top \boldsymbol{b}_\theta(\boldsymbol{X}_t)dt-\frac{1}{2}\int_0^T\|\boldsymbol{b}_\theta(\boldsymbol{X}_t)\|^2dt\nonumber\\
    &=\int_0^T(\boldsymbol{b}_\theta^\top\bar{\boldsymbol{b}})(\boldsymbol{X}_t)dt+ \int_0^T\boldsymbol{b}_\theta^\top d \boldsymbol{W}_t-\frac{1}{2}\int_0^T\|\boldsymbol{b}_\theta(\boldsymbol{X}_t)\|^2dt
\end{align}
\end{small}
Putting everything together into the expression for the cross-entropy objective, we get
\begin{small}
\begin{align}
    &\mathcal{L}_{\text{CE}}(\boldsymbol{b}_\theta)=\mathbb{E}_{\mathbb{P}^{v}}\left[w^\star(\boldsymbol{X}_{0:T})\left( r(\boldsymbol{X}_T)-\log Z-\log \frac{\mathrm{d}\mathbb{P}^{b_\theta}}{\mathrm{d}\mathbb{P}^0}(\boldsymbol{X}_{0:T})\right)\right]\nonumber\\
    &=\mathbb{E}_{\mathbb{P}^v}\left[w^\star(\boldsymbol{X}_{0:T})\left(r(\boldsymbol{X}_T)-\log Z-\left[\int_0^T(\boldsymbol{b}_\theta^\top\bar{\boldsymbol{b}})(\boldsymbol{X}_t)dt+ \int_0^T\boldsymbol{b}_\theta^\top d \boldsymbol{W}_t-\frac{1}{2}\int_0^T\|\boldsymbol{b}_\theta(\boldsymbol{X}_t)\|^2dt\right]\right)\right]\nonumber\\
    &=\mathbb{E}_{\mathbb{P}^v}\left[w^\star(\boldsymbol{X}_{0:T})\left(r(\boldsymbol{X}_T)-\log Z\right)-w^\star(\boldsymbol{X}_{0:T})\left(\int_0^T(\boldsymbol{b}_\theta^\top\bar{\boldsymbol{b}})(\boldsymbol{X}_t)dt+ \int_0^T\boldsymbol{b}_\theta^\top d \boldsymbol{W}_t-\frac{1}{2}\int_0^T\|\boldsymbol{b}_\theta(\boldsymbol{X}_t)\|^2dt\right)\right]\nonumber\\
    &=\mathbb{E}_{\mathbb{P}^v}\bigg[\underbrace{w^\star(\boldsymbol{X}_{0:T})\left(r(\boldsymbol{X}_T)-\log Z\right)}_{\text{independent of }\theta}+w^\star(\boldsymbol{X}_{0:T})\underbrace{\left(\frac{1}{2}\int_0^T\|\boldsymbol{b}_\theta(\boldsymbol{X}_t)\|^2dt-\int_0^T(\boldsymbol{b}_\theta^\top\bar{\boldsymbol{b}})(\boldsymbol{X}_t)dt- \int_0^T\boldsymbol{b}_\theta^\top d \boldsymbol{W}_t\right)}_{:=\mathcal{F}_{\boldsymbol{b}_\theta, \bar{\boldsymbol{b}}}(\boldsymbol{X}_{0:T})}\bigg]\nonumber
\end{align}
\end{small}

Since $w^\star(\boldsymbol{X}_{0:T})\left(r(\boldsymbol{X}_T)-\log Z\right)$ is independent of $\theta$, we can drop it and write the cross-entropy objective as
\begin{align}
    \mathcal{L}_{\text{CE}}(\theta)=\mathbb{E}_{\mathbb{P}^{\bar{b}}}\left[w^\star(\boldsymbol{X}_{0:T})\mathcal{F}_{\boldsymbol{b}_\theta, \bar{\boldsymbol{b}}}(\boldsymbol{X}_{0:T})\right] 
\end{align}
where we define $\bar{\boldsymbol{b}}:=\texttt{stopgrad}(\boldsymbol{b}_\theta)$. \hfill$\square{}$

\subsection{Proof of Proposition \ref{prop:discretized}}\label{appendix-prop:discretized}
\begin{tcolorbox}[sharp corners, colback=gray!10, boxrule=0pt]
\discretized*
\end{tcolorbox}
We will show that $\hat{\mathcal{F}}_{\boldsymbol{b}_\theta, \bar{\boldsymbol{b}}}(\boldsymbol{X}_{0:K})$, together with the terminal reward, can be written as
\begin{align}
    \hat{\mathcal{F}}_{\boldsymbol{b}_\theta, \bar{\boldsymbol{b}}}(\boldsymbol{X}_{0:K})+r(\boldsymbol{X}_K)=\log \left(\frac{p^0(\boldsymbol{X}_{0:K})\exp(r(\boldsymbol{X}_K))}{p^{b_\theta}(\boldsymbol{X}_{0:K})}\right)
\end{align}
First, we expand the expression as
\begin{small}
\begin{align}
    \log \left(\frac{p^0(\boldsymbol{X}_{0:K})\exp(r(\boldsymbol{X}_K))}{p^{b_\theta}(\boldsymbol{X}_{0:K})}\right)&=\log p^0(\boldsymbol{X}_{0:K})+r(\boldsymbol{X}_K)-\log p^{b_\theta}(\boldsymbol{X}_{0:K})\nonumber\\
    &=\sum_{k=0}^{K-1}\bigg[\log p^0(\boldsymbol{X}_{k+1}|\boldsymbol{X}_k)-\log p^{b_\theta}(\boldsymbol{X}_{k+1}|\boldsymbol{X}_k)\bigg]+r(\boldsymbol{X}_K)
\end{align}
\end{small}

Throughout this proof, $\boldsymbol{X}_{k+1}-\boldsymbol{X}_k$ denotes the increment after subtracting the base drift $\boldsymbol{f}(\boldsymbol{X}_k)\Delta t$. This drift is common to both densities and cancels in their log-ratio. The Gaussian normalizing constants cancel for the same reason, and $\boldsymbol{\Sigma}$ is diagonal, so $\boldsymbol{\Sigma}^\top\boldsymbol{\Sigma}=\boldsymbol{\Sigma}\boldsymbol{\Sigma}^\top$. 

For second-order dynamics, these densities apply to the velocity increments, as in the remark after Lemma~\ref{lemma:RND}. Under the uncontrolled dynamics, $\boldsymbol{X}_{k+1}$ is Gaussian with mean $\boldsymbol{X}_k$ and covariance $\boldsymbol{\Sigma}^\top\boldsymbol{\Sigma}\Delta t$. The log-density becomes
\begin{small}
\begin{align}
    \log p^0(\boldsymbol{X}_{k+1}|\boldsymbol{X}_k)&=\log \mathcal{N}\left(\boldsymbol{X}_{k+1}\big|\boldsymbol{X}_k, \boldsymbol{\Sigma}^\top\boldsymbol{\Sigma}\Delta t\right)\nonumber\\
    &=-\frac{1}{2}(\boldsymbol{X}_{k+1}-\boldsymbol{X}_k)^\top(\boldsymbol{\Sigma}^\top\boldsymbol{\Sigma}\Delta t)^{-1}(\boldsymbol{X}_{k+1}-\boldsymbol{X}_k)\label{eq:density-increment}
\end{align}
\end{small}
Similarly, the log-density under the bias force $\boldsymbol{b}_\theta$ is given by
\begin{small}
\begin{align}
    \log &p^{b_\theta}(\boldsymbol{X}_{k+1}|\boldsymbol{X}_k)=\log \mathcal{N}\left(\boldsymbol{X}_{k+1}\big|\boldsymbol{X}_k+\boldsymbol{\Sigma}\boldsymbol{b}_\theta(\boldsymbol{X}_k)\Delta t, \boldsymbol{\Sigma}^\top\boldsymbol{\Sigma}\Delta t\right)\nonumber\\
    &=-\frac{1}{2}(\boldsymbol{X}_{k+1}-\boldsymbol{X}_k-\boldsymbol{\Sigma}\boldsymbol{b}_\theta(\boldsymbol{X}_k)\Delta t)^\top(\boldsymbol{\Sigma}^\top\boldsymbol{\Sigma}\Delta t)^{-1}(\boldsymbol{X}_{k+1}-\boldsymbol{X}_k-\boldsymbol{\Sigma}\boldsymbol{b}_\theta(\boldsymbol{X}_k)\Delta t )\label{eq:density-increment-bar}
\end{align}
\end{small}
Now, we write the increment from time $k$ to $k+1$ as
\begin{align}
    \boldsymbol{X}_{k+1}&=\boldsymbol{X}_k+\boldsymbol{\Sigma}\bar{\boldsymbol{b}}(\boldsymbol{X}_k)\Delta t+\boldsymbol{\Sigma}\Delta \boldsymbol{W}_k\nonumber\\
    \boldsymbol{X}_{k+1}-\boldsymbol{X}_k&=\boldsymbol{\Sigma}\bar{\boldsymbol{b}}(\boldsymbol{X}_k)\Delta t+\boldsymbol{\Sigma}\Delta \boldsymbol{W}_k
\end{align}
and substitute into (\ref{eq:density-increment}) and (\ref{eq:density-increment-bar}) to get
\begin{small}
\begin{align}
    &\log p^0(\boldsymbol{X}_{k+1}|\boldsymbol{X}_k)-\log  p^{b_\theta}(\boldsymbol{X}_{k+1}|\boldsymbol{X}_k)\nonumber\\
    &=-\frac{1}{2}\big(\boldsymbol{\Sigma}\bar{\boldsymbol{b}}(\boldsymbol{X}_k)\Delta t+\boldsymbol{\Sigma}\Delta \boldsymbol{W}_k\big)^\top\big(\boldsymbol{\Sigma}^\top\boldsymbol{\Sigma}\Delta t\big)^{-1}\big(\boldsymbol{\Sigma}\bar{\boldsymbol{b}}(\boldsymbol{X}_k)\Delta t+\boldsymbol{\Sigma}\Delta \boldsymbol{W}_k\big)\nonumber\\
    &+\frac{1}{2}\big(\boldsymbol{\Sigma}\bar{\boldsymbol{b}}(\boldsymbol{X}_k)\Delta t+\boldsymbol{\Sigma}\Delta \boldsymbol{W}_k-\boldsymbol{\Sigma}\boldsymbol{b}_\theta(\boldsymbol{X}_k)\Delta t\big)^\top\big(\boldsymbol{\Sigma}^\top\boldsymbol{\Sigma}\Delta t\big)^{-1}\big(\boldsymbol{\Sigma}\bar{\boldsymbol{b}}(\boldsymbol{X}_k)\Delta t+\boldsymbol{\Sigma}\Delta \boldsymbol{W}_k-\boldsymbol{\Sigma}\boldsymbol{b}_\theta(\boldsymbol{X}_k)\Delta t \big)\nonumber
\end{align}
\end{small}

Given that $\big(\boldsymbol{\Sigma}^\top\boldsymbol{\Sigma}\Delta t\big)^{-1}=\frac{1}{\Delta t}(\boldsymbol{\Sigma}^\top\boldsymbol{\Sigma})^{-1}=\frac{1}{\Delta t }(\boldsymbol{\Sigma}^{-1}\boldsymbol{\Sigma}^{-\top})$, and rewriting $\boldsymbol{a}=\bar{\boldsymbol{b}}\Delta t+\Delta \boldsymbol{W}_k-\boldsymbol{b}_\theta\Delta t$ and $\bar{\boldsymbol{a}}=\bar{\boldsymbol{b}}\Delta t+\Delta \boldsymbol{W}_k$, we have
\begin{small}
\begin{align}
    &\log p^0(\boldsymbol{X}_{k+1}|\boldsymbol{X}_k)-\log  p^{b_\theta}(\boldsymbol{X}_{k+1}|\boldsymbol{X}_k)\nonumber\\
    &=\frac{1}{2\Delta t}\bigg[-\big(\boldsymbol{\Sigma}\bar{\boldsymbol{a}}\big)^\top\big(\boldsymbol{\Sigma}^{-1}\boldsymbol{\Sigma}^{-\top}\big)\big(\boldsymbol{\Sigma}\bar{\boldsymbol{a}}\big)+\big(\boldsymbol{\Sigma}\boldsymbol{a}\big)^\top\big(\boldsymbol{\Sigma}^{-1}\boldsymbol{\Sigma}^{-\top}\big)\big(\boldsymbol{\Sigma}\boldsymbol{a}\big)\bigg]\nonumber\\
    &=\frac{1}{2\Delta t}\bigg[-\bar{\boldsymbol{a}}^\top\left(\boldsymbol{\Sigma}^\top\boldsymbol{\Sigma}^{-1}\boldsymbol{\Sigma}^{-\top}\boldsymbol{\Sigma}\right)\bar{\boldsymbol{a}}+\boldsymbol{a}^\top\left(\boldsymbol{\Sigma}^\top\boldsymbol{\Sigma}^{-1}\boldsymbol{\Sigma}^{-\top}\boldsymbol{\Sigma}\right)\boldsymbol{a}\bigg]\nonumber\\
    &=\frac{1}{2\Delta t}\big[-\|\bar{\boldsymbol{a}}\|^2+\|\boldsymbol{a}\|^2\big]\nonumber\\
    &=\frac{1}{2\Delta t}\big[-\|\bar{\boldsymbol{b}}\Delta t+\Delta \boldsymbol{W}_k\|^2+\|\bar{\boldsymbol{b}}\Delta t+\Delta \boldsymbol{W}_k-\boldsymbol{b}_\theta\Delta t\|^2\big]\nonumber\\
    &=\frac{1}{2\Delta t}\left[-\|\bar{\boldsymbol{b}}\Delta t+\Delta \boldsymbol{W}_k\|^2+\|\bar{\boldsymbol{b}}\Delta t+\Delta \boldsymbol{W}_k\|^2-2(\boldsymbol{b}_\theta\Delta t ) \left(\bar{\boldsymbol{b}}\Delta t+\Delta \boldsymbol{W}_k\right)+\|\boldsymbol{b}_\theta\Delta t\|^2\right]\nonumber\\
    &=\frac{1}{2\Delta t }\left[-2(\boldsymbol{b}_\theta\cdot \bar{\boldsymbol{b}})(\Delta t)^2-2(\boldsymbol{b}_\theta\cdot \Delta \boldsymbol{W}_k)\Delta t+\|\boldsymbol{b}_\theta\|^2(\Delta t)^2\right]\nonumber\\
    &=-(\boldsymbol{b}_\theta\cdot \bar{\boldsymbol{b}})\Delta t-(\boldsymbol{b}_\theta\cdot \Delta \boldsymbol{W}_k)+\frac{1}{2}\|\boldsymbol{b}_\theta\|^2\Delta t
\end{align}
\end{small}
Summing over all $K$ steps and adding $r(\boldsymbol{X}_K)$, we have
\begin{small}
\begin{align}
    \sum_{k=0}^{K-1}&\bigg[\log p^0(\boldsymbol{X}_{k+1}|\boldsymbol{X}_k)-\log  p^{b_\theta}(\boldsymbol{X}_{k+1}|\boldsymbol{X}_k)\bigg]+r(\boldsymbol{X}_K)\nonumber\\
    &=\underbrace{\frac{1}{2}\sum_{k=0}^{K-1}\|\boldsymbol{b}_\theta(\boldsymbol{X}_k)\|^2\Delta t-\sum_{k=0}^{K-1}(\boldsymbol{b}_\theta\cdot \bar{\boldsymbol{b}})(\boldsymbol{X}_k)\Delta t-\sum_{k=0}^{K-1}\boldsymbol{b}_\theta(\boldsymbol{X}_k)\cdot \Delta \cdot \Delta \boldsymbol{W}_k}_{:=\hat{\mathcal{F}}_{\boldsymbol{b}_\theta, \bar{\boldsymbol{b}}}(\boldsymbol{X}_{0:K})}+r(\boldsymbol{X}_K)
\end{align}
\end{small}
Therefore, we can write $\hat{\mathcal{F}}_{\boldsymbol{b}_\theta, \bar{\boldsymbol{b}}}(\boldsymbol{X}_{0:K})+r(\boldsymbol{X}_K)$ as
\begin{align}
    \hat{\mathcal{F}}_{\boldsymbol{b}_\theta, \bar{\boldsymbol{b}}}(\boldsymbol{X}_{0:K})+r(\boldsymbol{X}_K)=\sum_{k=0}^{K-1}\log\frac{p^0(\boldsymbol{X}_{k+1}|\boldsymbol{X}_k)}{p^{b_\theta}(\boldsymbol{X}_{k+1}|\boldsymbol{X}_k)}+r(\boldsymbol{X}_K)=\log\left(\frac{p^0(\boldsymbol{X}_{0:K})\exp(r(\boldsymbol{X}_K))}{p^{b_\theta}(\boldsymbol{X}_{0:K})}\right)
\end{align}
where $r(\boldsymbol{X}_K):=\log\pi_\mathcal{B}(\boldsymbol{R}_K)$. Since $r(\boldsymbol{X}_K)$ does not depend on $\theta$, adding it to $\hat{\mathcal{F}}_{\boldsymbol{b}_\theta, \bar{\boldsymbol{b}}}$ leaves the gradient of the cross-entropy objective unchanged, which gives the stated loss.\hfill $\square$

\section{Cell Perturbation Experiment Details}
\label{app:cell-experiment}
\subsection{Experiment Setup}
\paragraph{Data Processing}
For this experiment, we extract the cell perturbation data from the Tahoe-100M dataset, which consists of 50 cell lines and over $1000$ different drug-dose conditions \citep{tahoe-100m}. Specifically, we use data on a single cell line (A-549) under two drug perturbation conditions: Clonidine at $5 \mu M$ and Trametinib at $5 \mu M$. The initial distribution is defined as the DMSO-treated control cells. Using centroid-based sampling, we balance each experiment by keeping, from every cluster, the cells closest to its centroid until it matches the smallest cluster in that experiment, giving 1033 cells per cluster for Clonidine and 381 for Trametinib (original counts in Table~\ref{table:perturbation cell counts}). We split the cells into training and validation sets with a 0.9/0.1 ratio. All final visualizations are plotted with the first two principal components, but the cell state trajectories are simulated for $d\in \{50, 100, 150\}$ for Clonidine and $d=50$ for Trametinib.
\begin{figure*}
    \centering
    \includegraphics[width=\linewidth]{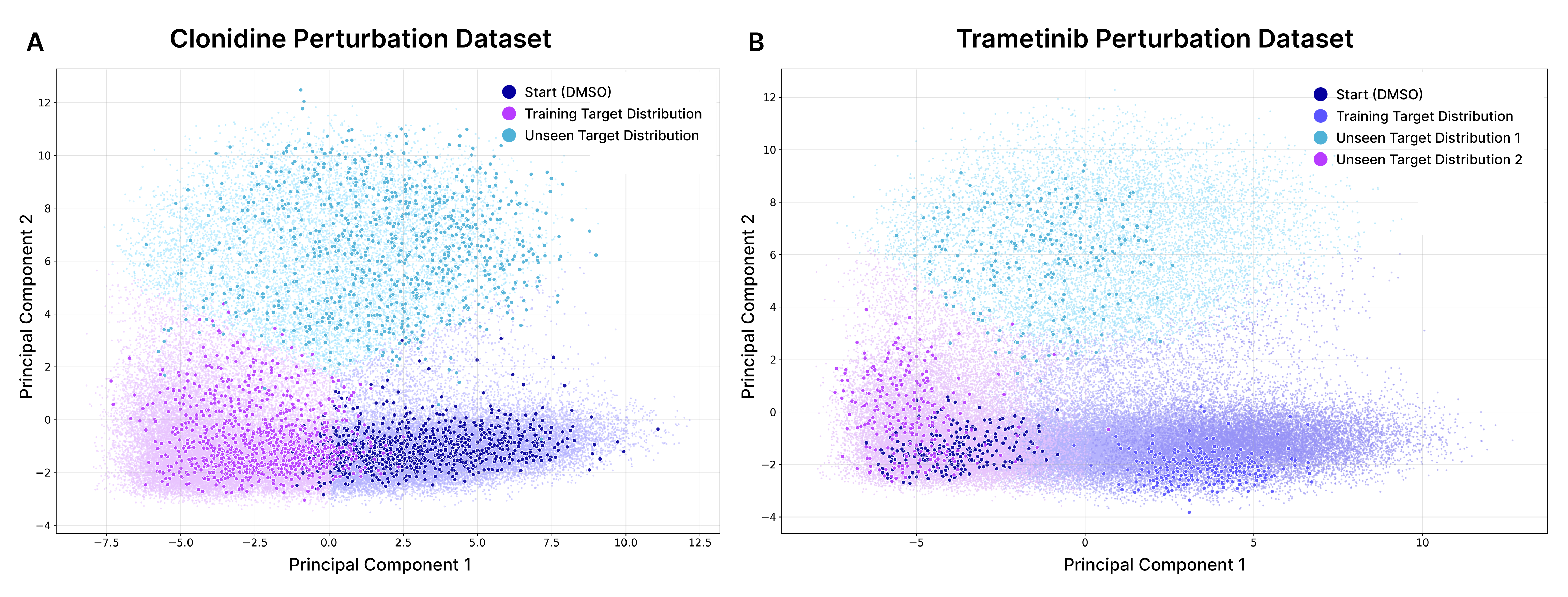}
    \caption{\textbf{Training and held-out cell clusters for the cell perturbation experiment.} All cells are plotted with the top 2 PCs. \textbf{(A)} Clonidine ($5\mu  M$) perturbation data containing the initial DMSO-treated control cells (dark blue), perturbed cluster for training (magenta), and held-out validation cluster (turquoise). \textbf{(B)} Trametinib ($5\mu M$) perturbation data containing the initial DMSO-treated control cells (navy blue), the perturbed cluster for training (purple), and two held-out perturbed clusters (magenta and turquoise). }
    \label{fig:tahoe-datasets}
\end{figure*}

\begin{table*}[h!]
\caption{\textbf{Cell counts per perturbed cluster for perturbation experiments.}}
\label{table:perturbation cell counts}
\begin{center}
\resizebox{0.6\linewidth}{!}{
\begin{small}
\begin{tabular}{@{}lccccc@{}}
\toprule
& \multicolumn{2}{c}{\textbf{Clonidine}}  & \multicolumn{3}{c}{\textbf{Trametinib}} \\
& Cluster 1 & Cluster 2 & Cluster 1 & Cluster 2 & Cluster 3 \\ 
\midrule
Original Cell Count & $1675$ & $1033$ & $1622$ & $686$ & $381$ \\
\bottomrule
\end{tabular}
\end{small}
}
\end{center}
\end{table*}

\paragraph{Model Architecture}
\label{app:Model Architecture}
To learn the dependencies across the $n$ particles, we use Transformer blocks to learn dependencies between the positions $\boldsymbol{R}_t\in \mathbb{R}^{n\times d}$ and velocities $\boldsymbol{V}_t\in \mathbb{R}^{n\times d}$ in the system. The input to the model is the concatenated position and velocity vectors \texttt{Cat}$(\boldsymbol{R}_t, \boldsymbol{V}_t, \boldsymbol{R}_\mathcal{B}-\boldsymbol{R}_t, \|\boldsymbol{R}_\mathcal{B}-\boldsymbol{R}_t\|)\in \mathbb{R}^{n\times(3d+1)}$. The architecture consists of an input projection layer that projects the input to $d_{\text{hidden}}=256$, a 4-layer Transformer Encoder (\texttt{gelu} activation, feedforward dimension $d_{\text{ff}}=512$, $8$ attention heads, and dropout $0.1$), and two MLP decoder heads that predict the scaling factor $\alpha_t(\boldsymbol{R}_t, \boldsymbol{V}_t)$ and correction vector $\boldsymbol{h}^i_t(\boldsymbol{R}_t, \boldsymbol{V}_t)$ per atom. To ensure positive scaling, the \texttt{softplus} activation is applied to the output of the scalar MLP head. The correction vector is projected onto the orthogonal plane of the vector pointing in the direction of the target $\hat{\boldsymbol{s}}_i$, and the output bias force is the sum of the scaled directional unit vector $\alpha_\theta^i (\boldsymbol{R}_t, \boldsymbol{V}_t)\hat{\boldsymbol{s}}_i$ and the projected correction vector $(\boldsymbol{I}-\hat{\boldsymbol{s}}_i \hat{\boldsymbol{s}}_i^\top)\boldsymbol{h}^i_\theta(\boldsymbol{R}_t, \boldsymbol{V}_t)$.

\paragraph{Terminal Reward}
To solve the EntangledSB problem, we defined the terminal reward to be the log-probability under the target density $\pi_\mathcal{B}$ given by $r(\boldsymbol{X}_T)=\log \pi_\mathcal{B}(\boldsymbol{R}_T)$. We set the target density as the Gaussian centered around $\boldsymbol{R}_\mathcal{B}$ with radius $\sigma$.
\begin{align}
    \pi_\mathcal{B}(\boldsymbol{R}_T)=\frac{1}{(2\pi\sigma^2)^{\frac{d}{2}}}\exp\left(\frac{-\|\boldsymbol{R}_T-\boldsymbol{R}_\mathcal{B}\|^2_2}{2\sigma^2}\right)
\end{align}

Therefore, to train the bias force to reconstruct the input target state of a cell cluster $\boldsymbol{R}_\mathcal{B}\in \mathbb{R}^{n\times d}$, we define $r:\mathcal{X}\to\mathbb{R}$ as:
\begin{align}
    r(\boldsymbol{X}_T):=\frac{-\|\boldsymbol{R}_T-\boldsymbol{R}_\mathcal{B}\|^2_2}{2\sigma^2}
\end{align}

The specific values for $\sigma$ are provided in Table \ref{table:cell-hyperparameters}.

\paragraph{Simulating Trajectories on the Data Manifold}
\label{app:data-manifold}
To define an energy landscape on the data manifold, we use the RBF metric introduced in \citet{Kapusniak2024}, which is lower in magnitude when $\boldsymbol{x}$ is within the support of the data manifold and larger in magnitude when $\boldsymbol{x}$ moves away from the support of the dataset into sparse regions. First, we define the elements of a function $h^{\text{RBF}}_j$ that satisfies $h^{\text{RBF}}_j(\boldsymbol{x})\approx 1$ on the data manifold as
\begin{align}
    h^{\text{RBF}}_{j}(\boldsymbol{x})&=\sum_{m=1}^{N_c }\omega_{m, j}(\boldsymbol{x})\exp\left(-\frac{\lambda_{m}}{2}\|\boldsymbol{x}-\hat{\boldsymbol{x}}_m\|^2\right)\\
    \lambda_m&=\frac{1}{2}\left(\frac{\kappa^2}{d\,|C_m|}\sum_{\boldsymbol{x}\in C_m}\|\boldsymbol{x}-\hat{\boldsymbol{x}}_m\|^2\right)^{-1}\label{appendix-eq:rbf}
\end{align}
where $\{\omega_{m, j}\}_{m=1}^{N_c}$ for each cluster and each coordinate are learned given a dataset $\mathcal{D}$ to enforce $h^{\text{RBF}}_{j}(\boldsymbol{x})\approx 1$ for all $\boldsymbol{x}\in \mathcal{D}$ with the following loss function
\begin{align}
    \mathcal{L}_{\text{RBF}}(\{\omega_{m, j}\})=\sum_{\boldsymbol{x}_i\in \mathcal{D}}(1- h^{\text{RBF}}_j (\boldsymbol{x}_i))^2
\end{align}
Then, we define the potential energy function of the base dynamics as
\begin{align}
    U(\boldsymbol{x})=\sum_{j=1}^d\log (M_j(\boldsymbol{x})+\varepsilon), \quad \text{where}\quad M_j(\boldsymbol{x})=\frac{1}{(h_j^{\text{RBF}}(\boldsymbol{x}) +\varepsilon )^{\alpha}}
\end{align}
where $M_j(\boldsymbol{x})$ is small in regions of high data density and large in regions of low data density. Since $U$ increases with each $M_j$, the potential is low on the data manifold and high in sparse regions. The simulator uses the natural-gradient force $-\nabla_{\boldsymbol{x}}U(\boldsymbol{x})/(M(\boldsymbol{x})+\varepsilon)$, where the division acts on each coordinate. This force points toward regions of higher data density.

\paragraph{Hyperparameters}
We present the hyperparameters used for the cell perturbation modeling experiment in Table \ref{table:cell-hyperparameters}. For each perturbation, we conducted ablations while keeping all other parameters constant on: \textbf{(1)} using the log-variance divergence with the learnable control variate $\mathcal{L}_{\text{LV}}$ described in App~\ref{app:ext-SOC-background} instead of the cross-entropy objective $\mathcal{L}_{\text{CE}}$ and \textbf{(2)} removing the dependency on velocities as a feature input to the bias force model $\boldsymbol{b}_\theta(\boldsymbol{R}_t)$. The hyperparameters of the Transformer architecture are given in App \ref{app:Model Architecture} and are kept constant across all experiments. All models are trained with the Adam optimizer \citep{kingma2014adam} with learning rate $\eta =0.0001$. Due to the small batch sizes used for training, we leverage the importance weights $w^\star(\boldsymbol{X}_{0:T})$ for categorical sampling from the replay buffer $\mathcal{R}$ as $\boldsymbol{X}_{0:T}\sim \texttt{Cat}(\texttt{softmax}_{\mathcal{R}}(w^\star(\boldsymbol{X}_{0:T})))$ to mimic the effect of the reweighting in the cross-entropy objective $\mathcal{L}_{\text{CE}}$ over a larger batch size. 

\begin{table*}[h]
\caption{\textbf{Hyperparameter settings for cell perturbation experiment.} The Clonidine perturbation experiment is split into three columns for each of the three dimensions of principal components (PCs) used, $d\in \{50, 100, 150\}$.}
\label{table:cell-hyperparameters}
\begin{center}
\begin{small}
\resizebox{0.7\linewidth}{!}{
\begin{tabular}{@{}lccccc@{}}
\toprule
\textbf{Parameter} &\multicolumn{3}{c}{\textbf{Clonidine}} & \textbf{Trametinib} & \textbf{Pancreatic $\beta$-Cell} \\ \midrule
 & $50$ PCs & $100$ PCs & $150$ PCs & $50$ PCs & $30$ PCs \\
\midrule
number of rollouts $N_{\text{rollouts}}$ & $100$ & $100$ & $100$ & $100$ & $100$ \\
trains per rollout $N_{\text{epochs}}$ & $1000$ & $1000$ & $1000$ & $1000$ & $1000$ \\
step size $\Delta t$ & $0.01$ & $0.01$ & $0.01$ & $0.01$ & $0.01$ \\
total time steps $T$ & $100$ & $100$ & $100$ & $100$ & $100$ \\
number of samples $M$ & $64$ & $64$ & $64$ & $64$ & $64$ \\
number of particles $n$ & $16$ & $16$ & $16$ & $16$ & $16$ \\
batch size $N_{\text{batch}}$ & $64$ & $64$ & $64$ & $64$ & $64$ \\
buffer size $|\mathcal{R}|$ & $1000$ & $1000$ & $1000$ & $1000$ & $1000$ \\
radius $\sigma$ & $0.1$ & $0.1$ & $0.1$ & $0.1$ & $0.1$ \\
friction $\gamma$ & $2.0$ & $2.0$ & $2.0$ & $2.0$ & $1.0$ \\
learning rate & $0.0001$ & $0.0001$ & $0.0001$ & $0.0001$ & $0.0001$ \\
RBF $N_c$ & $150$ & $300$ & $300$ & $150$ & $500$ \\
RBF $\kappa$ & $1.5$ & $2.0$ & $3.0$ & $1.5$ & $2.5$ \\
\bottomrule
\end{tabular}
}
\end{small}
\end{center}
\end{table*}

\subsection{Evaluation Metrics}
\label{app:cell-eval}
\paragraph{Maximum Mean Discrepancy (RBF-MMD)} We evaluate the reconstruction accuracy of the target distribution and the distribution simulated with EntangledSBM using MMD with the RBF kernel (RBF-MMD) on all $d$ principal components used during training. For Clonidine, we evaluate $d\in \{50, 100, 150\}$ and for Trametinib, we evaluate $d=50$. Given the simulated endpoints of $M$ paths for $M$ cell clusters $\{\boldsymbol{R}_T\in \mathbb{R}^{n\times d}\}_{j=1}^M$ and the target states $\{\boldsymbol{R}_\mathcal{B}\in \mathbb{R}^{n\times d}\}_{\ell=1}^M$, the RBF-MMD is calculated as 
\begin{small}
    \begin{align}
        \text{RBF-MMD}=\frac{1}{ M^2}\sum_{j=1}^M\sum_{\ell=1}^Mk_{\text{mix}}(\boldsymbol{R}^j_T, \boldsymbol{R}^\ell_T)+\frac{1}{M^2}\sum_{j=1}^M\sum_{\ell=1}^Mk_{\text{mix}}(\boldsymbol{R}^j_\mathcal{B}, \boldsymbol{R}^\ell_\mathcal{B})-\frac{2}{M^2}\sum_{j=1}^M\sum_{\ell=1}^Mk_{\text{mix}}(\boldsymbol{R}^j_T, \boldsymbol{R}^\ell_\mathcal{B})\label{app-eq:rbf-mmd}
    \end{align}
\end{small}
where we define the mixture of RBF kernel functions $k_{\text{mix}}(\cdot, \cdot)$ as
\begin{align}
    k_{\text{mix}}(\boldsymbol{R}, \boldsymbol{R}')=\frac{1}{|\Sigma|}\sum_{\sigma \in \Sigma}\exp\left(-\frac{\|\boldsymbol{R}-\boldsymbol{R}'\|^2}{2\sigma^2}\right)
\end{align}
for $\Sigma=\{0.01, 0.1, 1, 10, 100\}$.

\paragraph{1-Wasserstein ($\mathcal{W}_1$) and 2-Wasserstein ($
\mathcal{W}_2$) Distances} 
We further compute the $\mathcal{W}_1$ and $\mathcal{W}_2$ distances for the top two PCs of the simulated endpoints of $M$ paths for $M$ cell clusters $\{\boldsymbol{R}_T\in \mathbb{R}^{n\times d}\}_{j=1}^M$ which form the predicted distribution $\pi_T$ and the \textit{full distribution} that the target states are sampled from $\pi_\mathcal{B}$ since the $\mathcal{W}_1$ and $\mathcal{W}_2$ distances can be calculated for a pair of distributions with different sizes. Concretely, the $\mathcal{W}_1$ and $\mathcal{W}_2$ distances are calculated as 
\begin{align}
    \mathcal{W}_1&=\left(\min_{\pi\in \Pi(\pi_T, \pi_\mathcal{B})}\int\|\boldsymbol{R}_T-\boldsymbol{R}_\mathcal{B}\|_2d\pi(\boldsymbol{R}_T, \boldsymbol{R}_\mathcal{B})\right)\label{appendix-eq:W1}\\
    \mathcal{W}_2&=\left(\min_{\pi\in \Pi(\pi_T, \pi_\mathcal{B})}\int\|\boldsymbol{R}_T-\boldsymbol{R}_\mathcal{B}\|^2_2d\pi(\boldsymbol{R}_T, \boldsymbol{R}_\mathcal{B})\right)^{1/2}\label{appendix-eq:W2}
\end{align}
which quantify the minimal effort required to transform the simulated endpoint distribution into the true target distribution, demonstrating the ability of the model to capture the true perturbation dynamics.

\section{Pancreatic $\beta$-Cell Experiment Details}
\label{app:pancreatic-cell}
\subsection{Experiment Setup}
\paragraph{Dataset} The pancreatic $\beta$-cell dataset from \citet{veres2019charting} contains $51{,}274$ cells over eight time points that evolve from human pluripotent stem cells to pancreatic $\beta$-like cells. Following \citet{zhang2025modeling}, the gene expression representations are projected to a $30$-dimensional PCA space. 

\paragraph{Model Architecture and Hyperparameters} 
We leverage the same model architecture as the cell perturbation experiment described in App \ref{app:cell-experiment}, with hyperparameters provided in Table \ref{table:cell-hyperparameters}. 

\paragraph{Evaluation Metrics} 
We compute the 1-Wasserstein distance ($\mathcal{W}_1$) following \citet{zhang2025modeling}.

\section{Transition Path Sampling Experiment Details}
\label{app:tps-experiment}

\begin{figure*}
    \centering
    \includegraphics[width=\linewidth]{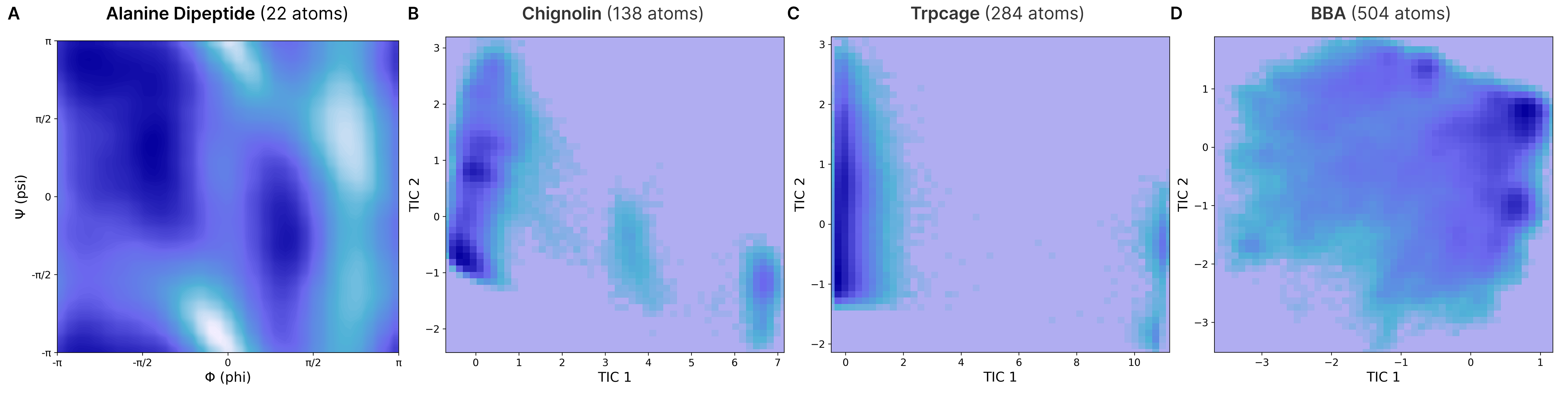}
    \caption{\textbf{Energy landscapes for transition path sampling experiments.} Potential energy plotted on dihedral angles $(\phi, \psi)$ for Alanine Dipeptide and the top two TICA components for fast-folding proteins.}
    \label{fig:tps-energy}
\end{figure*}

\subsection{Experiment Setup}
\paragraph{Model Architecture}
We use the same model architecture as the cell perturbation experiment described in App \ref{app:cell-experiment}, with the addition of Kabsch alignment \citep{kabsch1976solution}. We define the \textbf{aligned frame} as the frame of the target coordinates and align the input positions and velocities of the \textit{heavy atoms} (non-hydrogen atoms) to the aligned frame. The model then predicts the optimal bias force in the aligned frame, which we transform back to the original frame of the input positions.

\paragraph{Terminal Reward}
Following the cell perturbation experiment, given the coordinates of a single target state $\boldsymbol{R}_\mathcal{B}$, we define $r$ as the log-probability under a Gaussian of radius $\sigma$ centered at $\boldsymbol{R}_\mathcal{B}$, computed over all heavy atoms after Kabsch alignment:
\begin{align}
    r(\boldsymbol{X}_T):=\frac{-\|\boldsymbol{R}_T-\boldsymbol{R}_\mathcal{B}\|^2_2}{2\sigma^2}
\end{align}
No collective variable enters the reward or the model, and the backbone dihedrals (Alanine Dipeptide) and TICA components (fast-folding proteins) are used only to define a hit for the THP metric, following \citet{seong2025transition}.

The specific values for $\sigma$ are provided in Table \ref{table:tps-hyperparameters}.

\paragraph{Molecular Dynamics Setup}
We closely follow the setup of \citet{seong2025transition}. To simulate the position and velocity via the Langevin SDEs, we use the Velocity-Verlet with Velocity Randomization (VVVR) integrator \citep{sivak2014time} in OpenMM \citep{eastman2017openmm} that updates the state with a velocity-Verlet step for the deterministic forces and a stochastic Ornstein-Uhlenbeck step that randomizes velocities. We set the friction parameter to $\gamma = 1\text{ps}^{-1}$ and the time step size to $\Delta t=1\text{fs}$ for a total time horizon of $T=1000\text{fs}$ for Alanine Dipeptide and $T=5000\text{fs}$ for the fast-folding proteins. Following \citet{seong2025transition}, we leverage temperature annealing with a starting temperature of $\tau_{\textbf{start}}=600\text{K}$ and a final temperature of $\tau_{\textbf{end}}=300\text{K}$ for Alanine Dipeptide and Chignolin and $\tau_{\textbf{end}}=400\text{K}$ for the remaining fast-folding proteins. 

To simulate the unconditional dynamics following the potential energy landscape $U(\boldsymbol{R}_t)$ visualized in Fig. \ref{fig:tps-energy} and the corresponding force field $F=-\nabla_{\boldsymbol{R}_t}U(\boldsymbol{R}_t)$, we use the AMBER99SB force field with the ILDN side-chain torsion corrections (\texttt{amber99sbildn}) \citep{lindorff2010improved} for Alanine Dipeptide in a vacuum and the \texttt{ff14SBonlysc} force field \citep{maier2015ff14sb} paired with the \texttt{gbn2} solvation model \citep{nguyen2013improved} for the fast-folding proteins. 

\paragraph{Hyperparameters}
We present the hyperparameters used for the TPS experiment in Table \ref{table:tps-hyperparameters}. The hyperparameters of the Transformer architecture are given in App \ref{app:Model Architecture} and are kept constant across all experiments. We run the benchmark against TPS-DPS \citep{seong2025transition} with the hyperparameters below to ensure fair comparison. All models are trained with the Adam optimizer \citep{kingma2014adam} with learning rate $\eta = 1\times 10^{-4}$.

\begin{table*}[h]
\caption{\textbf{Hyperparameter settings for transition path sampling experiment.} The setup follows that of \citet{seong2025transition} to ensure fair comparison.}
\label{table:tps-hyperparameters}
\begin{center}
\begin{small}
\resizebox{0.7\linewidth}{!}{
\begin{tabular}{@{}lcccc@{}}
\toprule
\textbf{Parameter} &\multicolumn{4}{c}{\textbf{Task}} \\ \midrule
 & Alanine Dipeptide & Chignolin & Trpcage & BBA \\
\midrule
number of rollouts $N_{\text{rollouts}}$ & $100$ & $100$ & $100$ & $100$ \\
trains per rollout $N_{\text{epochs}}$ & $1000$ & $1000$ & $1000$ & $1000$ \\
step size $\Delta t$ & $1$\,fs & $1$\,fs & $1$\,fs & $1$\,fs\\
total time steps $T$ & $1000$ & $5000$ & $5000$ & $5000$ \\
number of samples $M$ & $64$ & $64$ & $64$ & $64$ \\
number of particles $n$ & $22$ & $138$ & $284$ & $504$ \\
batch size $N_{\text{batch}}$ & $16$ & $1$ & $1$ & $1$  \\
buffer size $|\mathcal{R}|$ & $1000$ & $100$ & $100$ & $100$ \\
starting temperature (Kelvin) & $600$ & $600$ & $600$ & $600$ \\
ending temperature (Kelvin) & $300$ & $300$ & $400$ & $400$ \\
radius $\sigma$ & $0.1$ & $0.5$ & $0.5$ & $0.5$ \\
friction $\gamma$ (fs$^{-1}$) & $0.001$ & $0.001$ & $0.001$ & $0.001$\\
learning rate & $0.0001$ & $0.0001$ & $0.0001$ & $0.0001$ \\
\bottomrule
\end{tabular}
}
\end{small}
\end{center}
\end{table*}

\subsection{Evaluation Metrics} 
\label{app:tps-eval}
We follow the evaluations of \citet{seong2025transition, holdijk2023stochastic} and report three metrics: Root Mean Square Distance (RMSD), Target Hit Percentage (THP), and Energy of Transition State (ETS).

\paragraph{Root Mean Square Distance (RMSD)} 
RMSD ($\downarrow$) measures the distance between the \textit{heavy atoms} (non-hydrogen atoms) of the final position of the system $\boldsymbol{R}_T$ and the target position $\boldsymbol{R}_{\mathcal{B}}$. To align the coordinate frames, we use the Kabsch algorithm \citep{kabsch1976solution}, which determines the optimal rotation and translation that aligns the two sets of heavy atoms. The mean and standard deviation across all $M$ paths are reported.

\paragraph{Target Hit Percentage (THP)} 
THP ($\uparrow$) measures the percentage of the total trajectories where the final position $\boldsymbol{R}_T$ reaches the vicinity of the target state $\boldsymbol{R}_\mathcal{B}$. Specifically, we consider a hit when the two backbone dihedral angles $(\phi, \psi)$ for Alanine Dipeptide or the first two TICA components for the fast-folding proteins (Chignolin, Trpcage, BBA), denoted $\xi(\boldsymbol{R})$, are within the $0.75$-radius sphere around the target defined as $\pi_{\mathcal{B}}=\{\boldsymbol{R}\;|\;\|\xi(\boldsymbol{R})-\xi(\boldsymbol{R}_{\mathcal{B}})\|<0.75\}$. For $M$ total paths, the THP is calculated as

\begin{align}
    \text{THP}=\frac{\sum_{i=1}^M\boldsymbol{1}[\boldsymbol{R}_T^i\in \pi_\mathcal{B}]}{M}
\end{align}

\paragraph{Energy of Transition State (ETS)} 
ETS ($\downarrow$) measures the maximum potential energy returned by $U:\mathcal{X}\to \mathbb{R}$ along the discrete transition path $\boldsymbol{R}_{0:K}$ in $\text{kJmol}^{-1}$. It is calculated only for trajectories that reach the vicinity of the target state ($\boldsymbol{R}_T\in \pi_\mathcal{B}$), which the THP metric classifies as hits. 
\begin{align}
    \text{ETS}(\boldsymbol{R}_{0:K}) =\max_{k\in \{1,\dots, K\}}U(\boldsymbol{R}_k)
\end{align}
The mean and standard deviation across the hits for $M$ paths are reported.

\subsection{Compute Time}
To assess the computational efficiency of our framework, we report the average rollout times of EntangledSBM alongside the closest baseline, TPS-DPS \citep{seong2025transition}, across molecular dynamics systems of increasing atomic complexity in Table \ref{tab:runtime_comparison}. EntangledSBM achieves runtimes that are comparable to TPS-DPS, with only modest overhead attributable to its Transformer-based interaction modeling. Rollout times for both methods increase with the number of atoms. The systems also differ in simulation length, so these times do not isolate the scaling with system size. For smaller systems such as Alanine Dipeptide (22 atoms), rollouts complete in well under a minute, while larger fast-folding proteins such as BBA (504 atoms) remain tractable within a few minutes per rollout. These results confirm that EntangledSBM preserves the practical scalability of transition path sampling methods while introducing the expressive capacity of Transformer-based interactions, making it well-suited for all-atom molecular simulations across diverse system sizes.

\begin{table*}[h]
\centering
\caption{\textbf{Runtime comparison between EntangledSBM and TPS-DPS for molecular dynamics experiments on a single NVIDIA B200 GPU}. We report rollout times against the closest baseline TPS-DPS \citep{seong2025transition} and show that EntangledSBM is comparable, with modest overhead due to Transformer-based interactions. Times are reported in seconds (mean $\pm$ standard deviation).}
\label{tab:runtime_comparison}
\resizebox{0.6\linewidth}{!}{
\begin{tabular}{llc}
\toprule
\textbf{System} & \textbf{Method} & \textbf{Rollout Time (s)} \\
\midrule
\multirow{2}{*}{Alanine Dipeptide (22 atoms)}  & TPS-DPS & $24.59_{\pm 0.23}$ \\
& EntangledSBM  & $31.47_{\pm 0.16}$ \\
\midrule
\multirow{2}{*}{Chignolin (138 atoms)} & TPS-DPS & $132.93_{\pm 1.05}$ \\
& EntangledSBM  & $160.84_{\pm 0.38}$ \\
\midrule
\multirow{2}{*}{Trpcage (284 atoms)} & TPS-DPS & $184.35_{\pm 2.01}$ \\
& EntangledSBM  & $196.62_{\pm 7.35}$ \\
\midrule
\multirow{2}{*}{BBA (504 atoms)} & TPS-DPS & $267.01_{\pm 1.85}$ \\
& EntangledSBM  & $291.68_{\pm 2.64}$ \\
\bottomrule
\end{tabular}
}
\end{table*}

\section{Comparison to Log-Variance Divergence}
\label{app:LV-comparison}
\begin{figure}[h!]
    \centering
    \includegraphics[width=\linewidth]{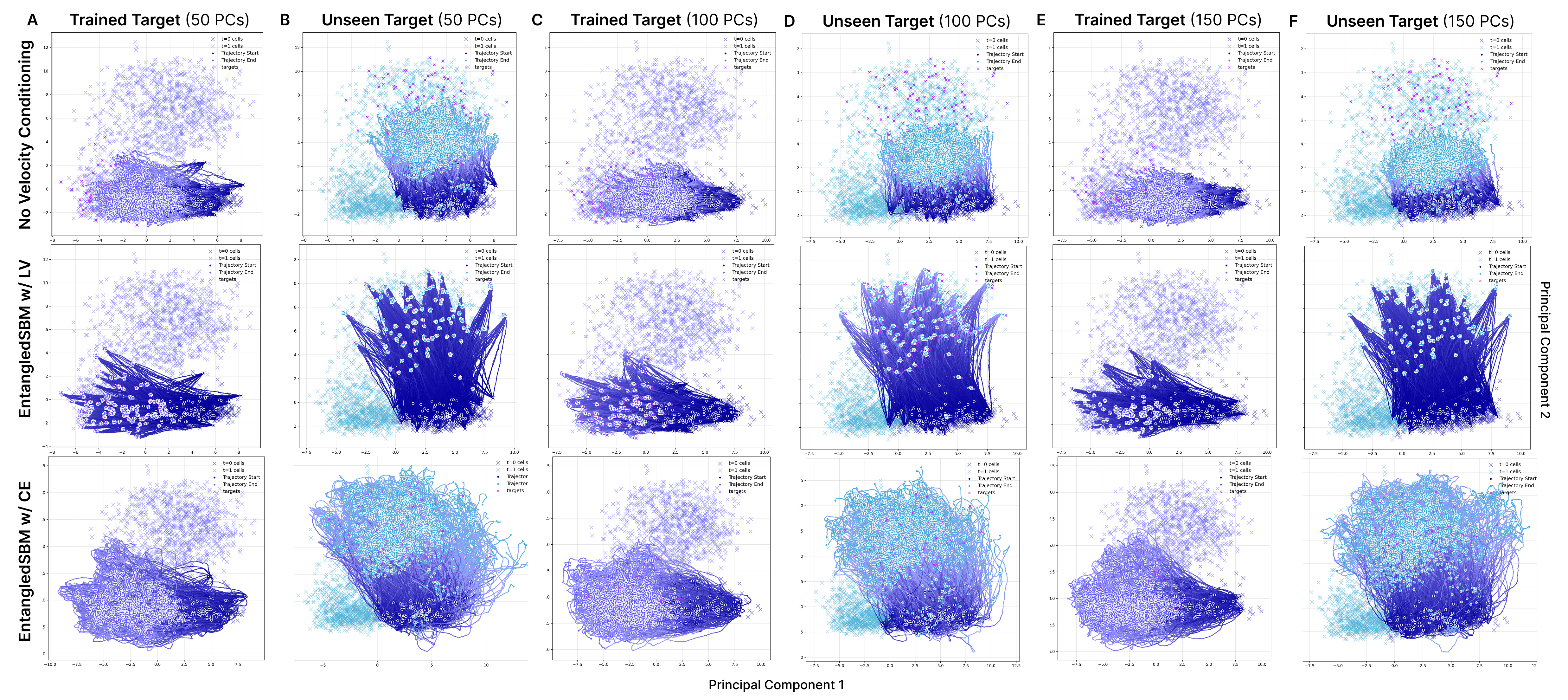}
    \caption{\textbf{Full visualization of simulated cell cluster dynamics with EntangledSBM under Clonidine perturbation.} Nearest neighbor cell clusters with $n=16$ cells are simulated over $100$ time steps. The gradient indicates the evolution of timesteps from the initial time $t=0$ (navy) to the final time $t=T$ (purple or turquoise). Rows compare no velocity conditioning, the log-variance (LV) objective, and the cross-entropy (CE) objective. 50 PCs simulated to \textbf{(A)} trained target distribution and \textbf{(B)} unseen target. 100 PCs simulated to \textbf{(C)} trained target distribution and \textbf{(D)} unseen target. 150 PCs simulated to \textbf{(E)} trained target distribution and \textbf{(F)} unseen target.}
    \label{fig:clonidine-full}
\end{figure}

\begin{figure*}[h]
    \centering
    \includegraphics[width=\linewidth]{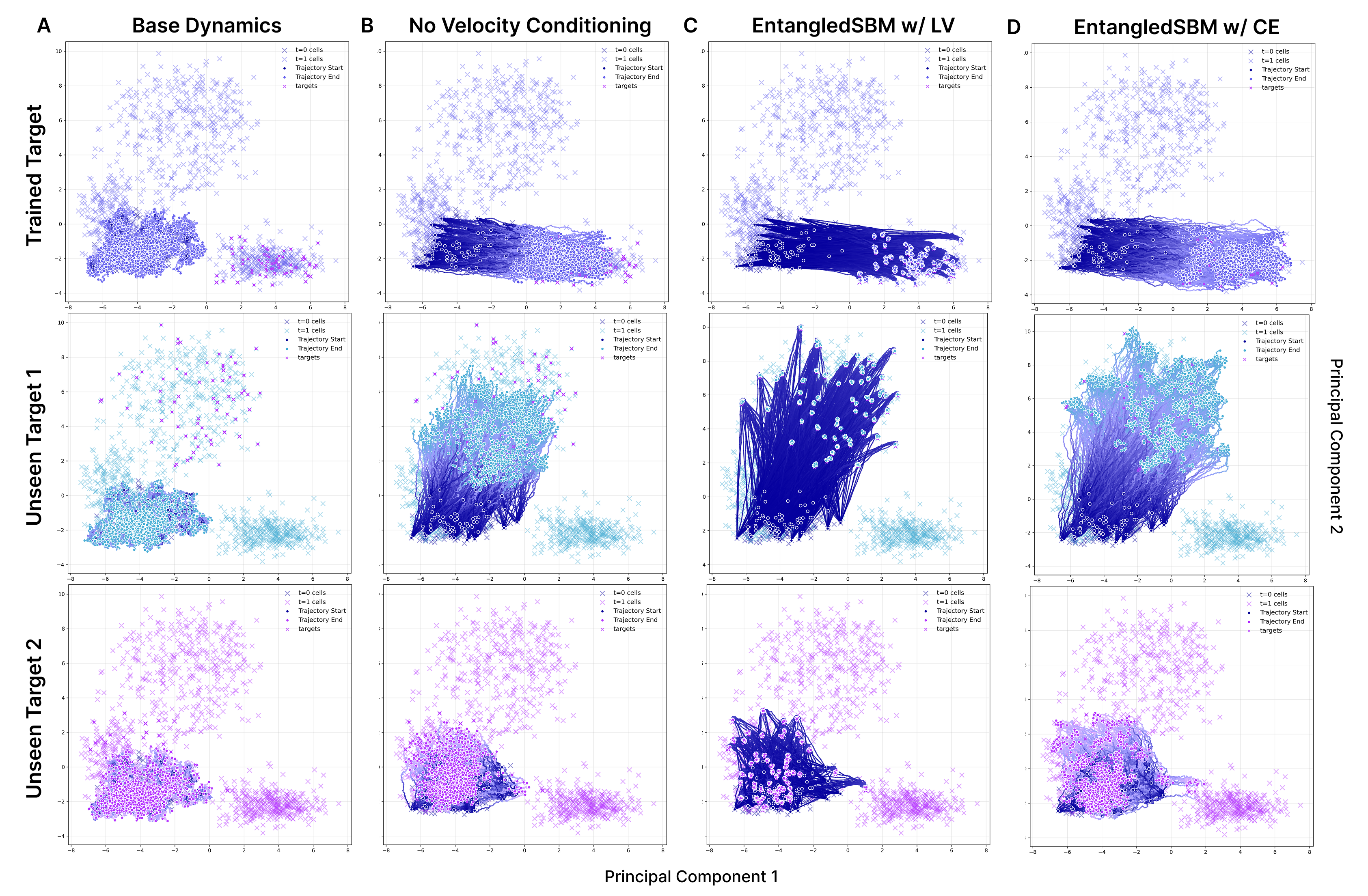}
    \caption{\textbf{Full visualization of simulated cell cluster dynamics with EntangledSBM under Trametinib perturbation.} All trajectories generated for $d=50$ PCs. Nearest neighbor cell clusters with $n=16$ cells are simulated over $100$ time steps. The gradient indicates the evolution of timesteps from the initial time $t=0$ (navy) to the final time $t=T$ (purple, turquoise, or magenta). \textbf{(A)} Base dynamics with no bias force. Dynamics with base and learned bias force trained with \textbf{(B)} no velocity conditioning, \textbf{(C)} log-variance (LV) objective, and \textbf{(D)} cross-entropy objective to the trained target distribution and two unseen target distributions.}
    \label{fig:trametinib-full}
\end{figure*}

In this section, we discuss the intuition behind the differences in cell state trajectories generated with the cross-entropy (CE) objective and log-variance (LV) objective for Clonidine, as shown in Fig. \ref{fig:clonidine-full} and Table \ref{table:clonidine-metrics-full}, and Trametinib, as shown in Fig. \ref{fig:trametinib-full} and Table \ref{table:trametinib-metrics-full}.

We observe that the log-variance (LV) objective fails to accurately simulate the intermediate dynamics of cells following perturbation and generates nearly straight, abrupt trajectories connecting the initial and terminal populations. The trajectories generated from the LV-trained bias forces ignore the gradual, curved progression of cell states through intermediate data manifolds, in contrast to the trajectories generated from the cross-entropy (CE)-trained bias forces (Fig. \ref{fig:clonidine-full}, \ref{fig:trametinib-full}). While the LV objective accurately reconstructs the target cell distribution, it fails to reconstruct the intermediate dynamics that arise from nonlinear coupling between position and velocity fields after perturbation (Fig.~\ref{fig:clonidine-full}, \ref{fig:trametinib-full}).

While the LV objective aims to make the log–likelihood ratio $\log \frac{\mathrm{d}\mathbb{P}^\star}{\mathrm{d}\mathbb{P}^{b_\theta}}(\boldsymbol{X}_{0:T})$ constant \textit{in expectation} to minimize the variance, it is easiest to only maximize the terminal reward $r(\boldsymbol{X}_T)$ rather than intermediate path alignment with $\mathbb{P}^0$. In contrast, the CE objective is defined as the KL-divergence $D_{\text{KL}}(\mathbb{P}^\star\|\mathbb{P}^{b_\theta})$ which expands into a path-integral action term $\frac{1}{2}\int_0^T\|\boldsymbol{b}_\theta(\boldsymbol{R}_t, \boldsymbol{V}_t)\|^2dt$ using Girsanov's theorem. This explicitly regularizes the whole trajectory, rewarding gradual, smooth transport through intermediate states.

\begin{table*}[h!]
\caption{\textbf{Full comparisons for simulating cell cluster dynamics under Clonidine perturbation with EntangledSBM.} We report RBF-MMD for all $d$ PCs with the target cluster and $\mathcal{W}_1$ and $\mathcal{W}_2$ distances of top 2 PCs against the full distribution of the perturbed cells for the seen and unseen populations after $100$ simulation steps and cluster size set to $n=16$. The mean and standard deviation of metrics from 5 independent simulations are reported. Comparisons include the base dynamics, without velocity conditioning, and EntangledSBM with the log-variance (LV) objective instead of cross-entropy (CE) for increasing principal component dimensions $d=\{50, 100, 150\}$.}
\label{table:clonidine-metrics-full}
\begin{center}
\begin{small}
\resizebox{\linewidth}{!}{
\begin{tabular}{@{}lcccccc@{}}
\toprule
& \multicolumn{3}{c}{\textbf{Seen Target Distribution}} &\multicolumn{3}{c}{\textbf{Unseen Target Distribution}} \\
\midrule
\textbf{Model} & RBF-MMD ($\downarrow$) & $\mathcal{W}_1$ ($\downarrow$) & $\mathcal{W}_2$ ($\downarrow$)  & RBF-MMD ($\downarrow$) & $\mathcal{W}_1$ ($\downarrow$) & $\mathcal{W}_2$ ($\downarrow$) \\
\midrule
\textbf{Base Dynamics} (50 PCs) & $0.677_{\pm 0.001}$ & $5.947_{\pm 0.005}$ & $6.015_{\pm 0.005}$ & $0.784_{\pm 0.001}$ & $8.217_{\pm 0.005}$ & $8.384_{\pm 0.005}$ \\
\midrule
\textbf{EntangledSBM w/o Velocity Conditioning}  & & &  & & & \\
50 PCs  & $0.440_{\pm 0.000}$ & $1.741_{\pm 0.003}$ & $1.857_{\pm 0.004}$ & $0.478_{\pm 0.000}$ & $2.907_{\pm 0.006}$ & $3.022_{\pm 0.006}$ \\
100 PCs & $0.494_{\pm 0.000}$ & $2.315_{\pm 0.004}$ & $2.423_{\pm 0.004}$ & $0.539_{\pm 0.000}$ & $4.110_{\pm 0.004}$ & $4.249_{\pm 0.003}$ \\
150 PCs & $0.510_{\pm 0.000}$ & $2.497_{\pm 0.006}$ & $2.620_{\pm 0.006}$ & $0.560_{\pm 0.000}$ & $4.573_{\pm 0.006}$ & $4.716_{\pm 0.007}$ \\
\midrule
\textbf{EntangledSBM w/ LV} & & &  & & & \\
50 PCs  & $0.294_{\pm 0.000}$ & $0.205_{\pm 0.001}$ & $0.285_{\pm 0.001}$ & $0.290_{\pm 0.000}$ & $0.230_{\pm 0.001}$ & $0.382_{\pm 0.001}$ \\
100 PCs & $0.330_{\pm 0.000}$ & $0.277_{\pm 0.001}$ & $0.332_{\pm 0.001}$ & $0.346_{\pm 0.000}$ & $0.358_{\pm 0.001}$ & $0.492_{\pm 0.001}$ \\
150 PCs & $0.323_{\pm 0.000}$ & $0.128_{\pm 0.001}$ & $0.278_{\pm 0.000}$ & $0.327_{\pm 0.000}$ & $0.193_{\pm 0.001}$ & $0.425_{\pm 0.001}$ \\
\midrule
\textbf{EntangledSBM w/ CE} &  & &   & & & \\
50 PCs  & $0.401_{\pm 0.000}$ & $0.342_{\pm 0.002}$ & $0.400_{\pm 0.001}$ & $0.419_{\pm 0.000}$ & $0.538_{\pm 0.013}$ & $0.705_{\pm 0.030}$ \\
100 PCs & $0.455_{\pm 0.000}$ & $0.953_{\pm 0.025}$ & $1.015_{\pm 0.025}$ & $0.500_{\pm 0.001}$ & $0.899_{\pm 0.006}$ & $1.055_{\pm 0.008}$ \\
150 PCs & $0.478_{\pm 0.000}$ & $0.753_{\pm 0.008}$ & $0.826_{\pm 0.007}$ & $0.506_{\pm 0.000}$ & $0.700_{\pm 0.009}$ & $0.811_{\pm 0.011}$  \\
\bottomrule
\end{tabular}
}
\end{small}
\end{center}
\end{table*}

\begin{table*}[h!]
\caption{\textbf{Full comparisons for simulating cell cluster dynamics under Trametinib perturbation with EntangledSBM.} We report RBF-MMD for all $d=50$ PCs with the target cluster and $\mathcal{W}_1$ and $\mathcal{W}_2$ distances of the top 2 PCs against the full distribution of the perturbed cells for the seen and unseen populations after $100$ simulation steps and cluster size set to $n=16$. Mean and standard deviation of metrics from 5 independent simulations are reported. Comparisons include the base dynamics, without velocity conditioning, and EntangledSBM with the log-variance (LV) objective instead of cross-entropy (CE).}
\label{table:trametinib-metrics-full}
\begin{center}
\begin{small}
\resizebox{\linewidth}{!}{
\begin{tabular}{@{}lccccccccc@{}}
\toprule
& \multicolumn{3}{c}{\textbf{Seen Target Distribution}} &\multicolumn{3}{c}{\textbf{Unseen Target Distribution 1}} &\multicolumn{3}{c}{\textbf{Unseen Target Distribution 2}} \\
\midrule
\textbf{Method} & RBF-MMD ($\downarrow$) & $\mathcal{W}_1$ ($\downarrow$) & $\mathcal{W}_2$ ($\downarrow$)  & RBF-MMD ($\downarrow$) & $\mathcal{W}_1$ ($\downarrow$) & $\mathcal{W}_2$ ($\downarrow$) & RBF-MMD ($\downarrow$) & $\mathcal{W}_1$ ($\downarrow$) & $\mathcal{W}_2$ ($\downarrow$)\\
\midrule
\textbf{Base Dynamics} & $0.938_{\pm 0.001}$ & $7.637_{\pm 0.005}$ & $7.653_{\pm 0.006}$ & $0.900_{\pm 0.000}$ & $7.766_{\pm 0.009}$ & $7.877_{\pm 0.009}$ & $0.754_{\pm 0.001}$ & $1.201_{\pm 0.009}$ & $1.455_{\pm 0.012}$ \\
\midrule
\textbf{EntangledSBM w/o Velocity Conditioning} & $0.449_{\pm 0.000}$ & $1.506_{\pm 0.005}$ & $1.544_{\pm 0.005}$ & $0.476_{\pm 0.000}$ & $2.116_{\pm 0.005}$ & $2.197_{\pm 0.005}$ & $0.480_{\pm 0.000}$ & $0.505_{\pm 0.004}$ & $0.627_{\pm 0.005}$ \\
\midrule
\textbf{EntangledSBM w/ LV} & $0.308_{\pm 0.000}$ & $0.175_{\pm 0.002}$ & $0.256_{\pm 0.001}$ & $0.302_{\pm 0.000}$ & $0.340_{\pm 0.001}$ & $0.565_{\pm 0.000}$ & $0.312_{\pm 0.000}$ & $0.198_{\pm 0.001}$ & $0.321_{\pm 0.001}$ \\
\textbf{EntangledSBM w/ CE} & $0.428_{\pm 0.000}$ & $0.392_{\pm 0.005}$ & $0.434_{\pm 0.006}$ & $0.409_{\pm 0.000}$ & $0.453_{\pm 0.008}$ & $0.561_{\pm 0.009}$ & $0.451_{\pm 0.000}$ & $0.394_{\pm 0.003}$ & $0.469_{\pm 0.004}$ \\
\bottomrule
\end{tabular}
}
\end{small}
\end{center}
\end{table*}

\section{Trajectory Diversity Analysis}
\label{app:diversity}

A natural concern when interpreting Energy of Transition State (ETS) values is whether higher ETS reflects \emph{meaningful diversity} of folding pathways or simply a mismatch with the underlying energy landscape. To disentangle these two possibilities, we directly compare EntangledSBM against TPS-DPS along a suite of trajectory-level diversity metrics that probe energetic, mechanistic, and geometric variation across target-hitting trajectories.

\subsection{Metrics}
For each method, we restrict attention to trajectories that successfully reach the target state and compute the following quantities:

\paragraph{ETS range} (kJ/mol) Difference between the maximum and minimum ETS values among target-hitting trajectories. A wider range indicates that the method explores transition states spanning a broader region of the energy landscape.

\paragraph{Contact correlation mean} (lower is more diverse) For each pair of target-hitting trajectories, we extract the contact formation order vector (which native contacts form first, second, and so on along the path) and compute the Pearson correlation between the two vectors. We report the mean correlation across all pairs. Low mean correlation indicates that trajectories fold via mechanistically distinct routes, rather than reordering minor variants of the same pathway.

\paragraph{Energy effective dimensionality} (higher is more diverse) Effective number of principal components required to describe the ensemble of per-trajectory energy profiles. Higher values indicate that energy profiles vary along many independent axes rather than along a single dominant mode.

\paragraph{$R_g$ midpoint standard deviation} (nm, higher is more diverse) Standard deviation of the radius of gyration evaluated at the midpoint of each trajectory. A higher value indicates that some trajectories pass through compact intermediates while others pass through expanded ones, providing a coarse geometric signature of distinct folding mechanisms.

\subsection{Results}
Table~\ref{tab:diversity_eval} reports the four diversity metrics on Chignolin and BBA, the two systems with the highest ETS values.

\begin{table*}[h]
    \centering
    \caption{\textbf{Trajectory diversity comparison between EntangledSBM and TPS-DPS on protein folding experiments.} Bold indicates the more diverse method on each metric. Contact correlation is reported as the mean $\pm$ standard deviation across trajectory pairs.}
    \label{tab:diversity_eval}
    \small
    \resizebox{\linewidth}{!}{
    \begin{tabular}{llcccc}
        \toprule
        \textbf{Protein} & Method & ETS range (kJ/mol) & Contact corr. mean ($\downarrow$) & Energy eff. dimensionality ($\uparrow$) & $R_g$ midpoint std ($\uparrow$; nm) \\
        \midrule
        \multirow{2}{*}{\textbf{Chignolin}}
            & TPS-DPS & $433.45$ & $0.775_{\pm 0.024}$ & $2.17$ & $0.02$ \\
            & \textbf{EntangledSBM} & $\mathbf{535.58}$  & $\mathbf{0.698_{\pm 0.123}}$ & $\mathbf{4.40}$ & $\mathbf{0.03}$ \\
        \midrule
        \multirow{2}{*}{\textbf{BBA}}
            & TPS-DPS & $539.52$ & $\mathbf{0.720_{\pm 0.018}}$ & $4.07$ & $0.01$ \\
            & \textbf{EntangledSBM} & $\mathbf{1097.26}$ & $0.726_{\pm 0.029}$ & $\mathbf{8.35}$ & $\mathbf{0.02}$ \\
        \bottomrule
    \end{tabular}
    }
\end{table*}

\paragraph{Energetic Diversity}
EntangledSBM achieves a substantially wider ETS range than TPS-DPS on both systems, with $535.58$ vs. $433.45$ kJ/mol on Chignolin and $1097.26$ vs. $539.52$ kJ/mol on BBA, the latter representing roughly a twofold increase. This indicates that EntangledSBM samples transition states spanning a broader region of the energy landscape, rather than concentrating on a single low-energy channel.

\paragraph{Mechanistic Diversity}
On Chignolin, EntangledSBM achieves a mean pairwise correlation of $0.698$ compared with $0.775$ for TPS-DPS, showing that pairs of EntangledSBM trajectories differ more in the \textit{ordering} of native contact formation. On BBA, the two methods are essentially tied on this metric ($0.726$ vs. $0.720$), but EntangledSBM substantially outperforms TPS-DPS on energy-effective dimensionality ($8.35$ vs. $4.07$) and $R_g$ midpoint variability, indicating that even when contact ordering is comparable, the underlying energetic and geometric trajectories explore a richer set of independent modes.

These metrics show that the higher ETS values produced by EntangledSBM correspond to diverse and structured folding mechanisms rather than to a mismatch with the energy landscape. The combination of a wider ETS range, lower contact ordering correlation, and higher energy effective dimensionality indicates that the model produces interpretable variation in how the protein folds (i.e., different contact orders, different intermediate geometries, and different energetic profiles) while reaching the target state with higher THP and lower RMSD than TPS-DPS. 

\section{Physical Feasibility of Generated Trajectories}
\label{app:physics_eval}

To assess whether the trajectories generated by EntangledSBM are physically plausible, we evaluate each sampled path along three structural and dynamical metrics that together characterize local smoothness, steric feasibility, and structural consistency with the target fold.

\subsection{Metrics}
We compute the following quantities along generated trajectories:

\paragraph{Smoothness} (nm/step, $\downarrow$) Mean per-step atomic displacement, averaged over all heavy atoms and integration steps. This quantifies the local continuity of the trajectory and detects abrupt discontinuities that would be incompatible with physical evolution.

\paragraph{Steric Clashes} (count, $\downarrow$) Number of heavy-atom pairs whose pairwise distance falls below a clash threshold of $1.2\,\text{\AA}$ ($0.12\,\text{nm}$), computed over all unordered heavy-atom pairs in each frame and summed across the trajectory. The threshold is chosen below typical heavy-atom bond lengths, so bonded neighbors are not counted as clashes.

\paragraph{Contact Fraction} ($\uparrow$) Fraction of native residue-residue contacts recovered along the path, where a native contact is defined by a C$\alpha$-C$\alpha$ distance below the standard $8\,\text{\AA}$ cutoff in the reference structure. This captures structural correctness with respect to the target fold.

\subsection{Results}
Table~\ref{tab:physics_eval} summarizes the three metrics for both \textbf{Chignolin} and \textbf{BBA}. Per-step displacements are on the order of $3 \times 10^{-3}\,\text{nm}$, indicating smooth evolution along the path. Clash counts represent about $0.13\%$ (Chignolin) and $0.05\%$ (BBA) of all heavy-atom pairs per frame, indicating that intermediate conformations remain largely free of non-physical atomic overlap, and contact fractions exceed $0.96$, with BBA recovering essentially all native contacts ($0.994$). Together, these results indicate that EntangledSBM trajectories evolve continuously, remain sterically feasible, and are structurally consistent with the target fold throughout the transition.

\begin{table*}[h]
    \caption{\textbf{Physical feasibility of intermediate trajectories for Chignolin and BBA.} Arrows indicate the preferred direction.}
    \label{tab:physics_eval}
    \centering
    \resizebox{0.7\linewidth}{!}{
    \begin{tabular}{lccc}
        \toprule
        \textbf{Protein} & Smoothness ($\downarrow$ nm/step) & Mean clashes ($\downarrow$) & Contact fraction ($\uparrow$) \\
        \midrule
        \textbf{Chignolin} & $0.00306$ & $3.9$  & $0.965$ \\
        \textbf{BBA} & $0.00296$ & $14.8$ & $0.994$ \\
        \bottomrule
    \end{tabular}
    }
\end{table*}

\clearpage
\section{Algorithms}
\label{app:Algorithms}
Here, we provide the pseudocode for training (Alg \ref{alg:EntangledSBM Training}) and inference (Alg \ref{alg:EntangledSBM Inference}) with EntangledSBM.
\begin{algorithm}[]
\caption{EntangledSBM \textbf{Training}}\label{alg:EntangledSBM Training}
    \begin{algorithmic}[1]
        \State \textbf{Input:} Parameterized networks $\alpha_\theta(\boldsymbol{R}_t, \boldsymbol{V}_t):\mathbb{R}^{n\times d}\times \mathbb{R}^{n\times d}\to \mathbb{R}^{n}$ and $\boldsymbol{h}_\theta(\boldsymbol{R}_t, \boldsymbol{V}_t):\mathbb{R}^{n\times d}\times \mathbb{R}^{n\times d}\to \mathbb{R}^{n\times d}$, potential energy function $U(\boldsymbol{R}_t): \mathbb{R}^{n\times d}\to \mathbb{R}$, initial distribution $\pi_\mathcal{A}$, distribution of target states $\pi_\mathcal{B}$, buffer size $|\mathcal{R}|$, batch size $N_{\text{batch}}$, number of samples $M$, number of rollouts $N_{\text{rollouts}}$, training steps per rollout $N_{\text{epochs}}$,  timestep size $\Delta t$
        \State $\mathcal{R}\gets\{\}$\Comment{Initialize empty replay buffer}
        \For {rollout in $1, \dots, N_{\text{rollouts}}$}
        \For{$j$ in $1, \dots, M$}         
            \State Sample initial state $\boldsymbol{X}_0=(\boldsymbol{R}_0, \boldsymbol{V}_0)$ with $\boldsymbol{R}_0\sim\pi_\mathcal{A}$         
            \For{$t$ in $0, \dots, T-1$}
            \State Predict $\alpha_\theta^i(\boldsymbol{R}_t, \boldsymbol{V}_t)\in \mathbb{R}$ and $\boldsymbol{h}_\theta^i(\boldsymbol{R}_t, \boldsymbol{V}_t)\in \mathbb{R}^d$ with parameterized neural network where $\alpha_\theta^i(\boldsymbol{R}_t, \boldsymbol{V}_t)\geq 0$ is enforced with \texttt{softplus} activation
            \State $\boldsymbol{s}_i\gets\nabla_{\boldsymbol{r}_t^i}\log \pi_\mathcal{B}$, $\hat{\boldsymbol{s}}_i\gets \boldsymbol{s}_i/\|\boldsymbol{s}_i\|$
            \State Compute bias force 
            \begin{align}
                \boldsymbol{b}_\theta^i(\boldsymbol{R}_t,\boldsymbol{V}_t)\gets\alpha^i_\theta(\boldsymbol{R}_t, \boldsymbol{V}_t)\hat{\boldsymbol{s}}_i+\left(\boldsymbol{I}-\hat{\boldsymbol{s}}_i \hat{\boldsymbol{s}}_i^\top\right)\boldsymbol{h}^i_\theta(\boldsymbol{R}_t, \boldsymbol{V}_t)\nonumber
            \end{align}
            \State Update each particle $i$ following (\ref{eq:multi-langevin}), shown here with Euler-Maruyama integration (the TPS experiments use the VVVR integrator, and the cell experiments use the first-order form)
            \begin{align}
                \boldsymbol{r}^i_{t+1}&=\boldsymbol{r}^i_t+\boldsymbol{v}^i_t\Delta t\nonumber\\
                \boldsymbol{v}^i_{t+1}&=\boldsymbol{v}^i_t+\left(\frac{-\nabla_{\boldsymbol{r}^i_t}U(\boldsymbol{R}_t)+\boldsymbol{b}^i_\theta(\boldsymbol{R}_t, \boldsymbol{V}_t)}{m_i}-\gamma\boldsymbol{v}^i_t\right)\Delta t+\sqrt{\frac{2\gamma k_B\tau\Delta t}{m_i}}\,\boldsymbol{\epsilon}^i_t\nonumber
            \end{align}
        \EndFor
        \EndFor
        
        \State $\mathcal{R}\gets \mathcal{R}\cup\{\boldsymbol{X}_{0:T}\}_{j=1}^{M}$\Comment{update buffer with discrete trajectories}
        \For{step in $1, \dots, N_{\text{epochs}}$}
            \State Sample batch $\{\boldsymbol{X}_{0:T}\}_{j=1}^{N_{\text{batch}}}$ from buffer $\mathcal{R}$
            \State Compute cross-entropy objective $\mathcal{L}_{\text{CE}}$ with Prop.~\ref{prop:discretized}
            \State Update $\theta$ with $\nabla_{\theta}\mathcal{L}_{\text{CE}}$
        \EndFor
        \EndFor
        \State \textbf{return} parameterized $\alpha_\theta^i(\boldsymbol{R}_t, \boldsymbol{V}_t)$ and $\boldsymbol{h}_\theta^i(\boldsymbol{R}_t, \boldsymbol{V}_t)$
    \end{algorithmic}
\end{algorithm}
\begin{algorithm}[h]
\caption{EntangledSBM \textbf{Inference}}\label{alg:EntangledSBM Inference}
    \begin{algorithmic}[1]
        \State \textbf{Input:} Trained networks $\alpha_\theta^i(\boldsymbol{R}_t, \boldsymbol{V}_t)$ and $\boldsymbol{h}_\theta^i(\boldsymbol{R}_t, \boldsymbol{V}_t)$,  potential energy function $U(\boldsymbol{R}_t)$, initial state $\boldsymbol{X}_0=(\boldsymbol{R}_0, \boldsymbol{V}_0)$, target state $\boldsymbol{R}_\mathcal{B}$, time steps $T$, step size $\Delta t$, friction $\gamma$
        \State $\boldsymbol{R}_t\gets \boldsymbol{R}_0$, $\boldsymbol{V}_t\gets \boldsymbol{V}_0$
        \State $\mathcal{P}\gets \{\boldsymbol{R}_0\}$\Comment{initialize path}
        \For{$t$ in $0, \dots, T-1$}
            \State Predict $\alpha_\theta^i(\boldsymbol{R}_t, \boldsymbol{V}_t)$ and $\boldsymbol{h}_\theta^i(\boldsymbol{R}_t, \boldsymbol{V}_t)$ with parameterized neural network where $\alpha_\theta^i(\boldsymbol{R}_t, \boldsymbol{V}_t)\geq 0$ is enforced with \texttt{softplus} activation
            \State $\boldsymbol{s}_i\gets\nabla_{\boldsymbol{r}_t^i}\log \pi_\mathcal{B}$, $\hat{\boldsymbol{s}}_i\gets \boldsymbol{s}_i/\|\boldsymbol{s}_i\|$
            \State Compute bias force 
            \begin{align}
                \boldsymbol{b}_\theta^i(\boldsymbol{R}_t,\boldsymbol{V}_t)\gets\alpha^i_\theta(\boldsymbol{R}_t, \boldsymbol{V}_t)\hat{\boldsymbol{s}}_i+\left(\boldsymbol{I}-\hat{\boldsymbol{s}}_i \hat{\boldsymbol{s}}_i^\top\right)\boldsymbol{h}^i_\theta(\boldsymbol{R}_t, \boldsymbol{V}_t)\nonumber
            \end{align}
            \State Update each particle $i$ following (\ref{eq:multi-langevin}), shown here with Euler-Maruyama integration (the TPS experiments use the VVVR integrator, and the cell experiments use the first-order form)
            \begin{align}
                \boldsymbol{r}^i_{t+1}&=\boldsymbol{r}^i_t+\boldsymbol{v}^i_t\Delta t\nonumber\\
                \boldsymbol{v}^i_{t+1}&=\boldsymbol{v}^i_t+\left(\frac{-\nabla_{\boldsymbol{r}^i_t}U(\boldsymbol{R}_t)+\boldsymbol{b}^i_\theta(\boldsymbol{R}_t, \boldsymbol{V}_t)}{m_i}-\gamma\boldsymbol{v}^i_t\right)\Delta t+\sqrt{\frac{2\gamma k_B\tau\Delta t}{m_i}}\,\boldsymbol{\epsilon}^i_t\nonumber
            \end{align}
            \State Append to path $\mathcal{P}\gets \mathcal{P}\cup \{\boldsymbol{R}_{t+1}\}$
        \EndFor
        \State \textbf{return} path $\mathcal{P}$
    \end{algorithmic}
\end{algorithm}


\clearpage
\section*{NeurIPS Paper Checklist}

\begin{enumerate}

\item {\bf Claims}
    \item[] Question: Do the main claims made in the abstract and introduction accurately reflect the paper's contributions and scope?
    \item[] Answer: \answerYes{} 
    \item[] Justification: We clearly state the contributions and scope in the abstract and the \textbf{main contributions} list in the introduction.

\item {\bf Limitations}
    \item[] Question: Does the paper discuss the limitations of the work performed by the authors?
    \item[] Answer: \answerYes{} 
    \item[] Justification: We discuss limitations in the Limitations paragraph of Sec~\ref{sec:conclusion} and in more detail in Appendix~\ref{app:Limitations}.

\item {\bf Theory assumptions and proofs}
    \item[] Question: For each theoretical result, does the paper provide the full set of assumptions and a complete (and correct) proof?
    \item[] Answer: \answerYes{} 
    \item[] Justification: Rigorous proofs for our theoretical results are provided in Appendix \ref{app:Theoretical Proofs}.

    \item {\bf Experimental result reproducibility}
    \item[] Question: Does the paper fully disclose all the information needed to reproduce the main experimental results of the paper to the extent that it affects the main claims and/or conclusions of the paper (regardless of whether the code and data are provided or not)?
    \item[] Answer: \answerYes{} 
    \item[] Justification: We provide complete details to enable reproduction of our results. Experiment details for cell perturbation modeling, pancreatic $\beta$-cell differentiation modeling, and transition path sampling are provided in App \ref{app:cell-experiment}, \ref{app:pancreatic-cell}, and \ref{app:tps-experiment}. Algorithms \ref{alg:EntangledSBM Training} and~\ref{alg:EntangledSBM Inference} provide pseudocode for training and inference. Code and data are provided in the repository: \href{https://github.com/sophtang/EntangledSBM}{https://github.com/sophtang/EntangledSBM}. Together, these resources ensure full reproducibility.

\item {\bf Open access to data and code}
    \item[] Question: Does the paper provide open access to the data and code, with sufficient instructions to faithfully reproduce the main experimental results, as described in supplemental material?
    \item[] Answer: \answerYes{} 
    \item[] Justification: Code and data are provided in the repository \href{https://github.com/sophtang/EntangledSBM}{https://github.com/sophtang/EntangledSBM} and on Hugging Face at \url{https://huggingface.co/ChatterjeeLab/EntangledSBM}. 

\item {\bf Experimental setting/details}
    \item[] Question: Does the paper specify all the training and test details (e.g., data splits, hyperparameters, how they were chosen, type of optimizer) necessary to understand the results?
    \item[] Answer: \answerYes{} 
    \item[] Justification: Full training and test details are provided in App~\ref{app:cell-experiment} (cell perturbation), App~\ref{app:pancreatic-cell} (pancreatic $\beta$-cell differentiation), and App~\ref{app:tps-experiment} (transition path sampling). This includes data processing, model architecture (App~\ref{app:Model Architecture}), terminal reward formulation, data manifold construction (App~\ref{app:data-manifold}), and the Adam optimizer configuration with learning rate $\eta = 1\times10^{-4}$. Full hyperparameter values are tabulated in Tables~\ref{table:cell-hyperparameters} and~\ref{table:tps-hyperparameters}, and evaluation metric definitions are provided in App~\ref{app:cell-eval} and~\ref{app:tps-eval}.

\item {\bf Experiment statistical significance}
    \item[] Question: Does the paper report error bars suitably and correctly defined or other appropriate information about the statistical significance of the experiments?
    \item[] Answer: \answerYes{} 
    \item[] Justification: All quantitative results in Tables~\ref{table:cell-metrics},~\ref{tab:beta-cell},~\ref{table:tps-benchmark},~\ref{table:clonidine-metrics-full}, and~\ref{table:trametinib-metrics-full} report mean and standard deviation, except THP, which is a percentage of paths and has no error bar. Cell perturbation metrics (RBF-MMD, $\mathcal{W}_1$, $\mathcal{W}_2$) are computed across 5 independent simulations as described in App~\ref{app:cell-experiment} and App~\ref{app:cell-eval}. Transition path sampling metrics are computed from 64 paths per protein following App~\ref{app:tps-eval}: RMSD and THP use all paths, and ETS uses target-hitting paths. Runtime measurements in Table~\ref{tab:runtime_comparison} also report mean and standard deviation across runs.

\item {\bf Experiments compute resources}
    \item[] Question: For each experiment, does the paper provide sufficient information on the computer resources (type of compute workers, memory, time of execution) needed to reproduce the experiments?
    \item[] Answer: \answerYes{} 
    \item[] Justification: Compute resources are reported in App~\ref{app:tps-experiment} for the transition path sampling experiments, with detailed rollout times on a single NVIDIA B200 GPU across all four protein systems (Alanine Dipeptide, Chignolin, Trp-cage, BBA) in Table~\ref{tab:runtime_comparison}. Hyperparameters that determine compute usage (number of rollouts, training steps per rollout, total time steps, batch size, buffer size) are documented for the cell perturbation experiments in Table~\ref{table:cell-hyperparameters} and for the transition path sampling experiments in Table~\ref{table:tps-hyperparameters}.
    
\item {\bf Code of ethics}
    \item[] Question: Does the research conducted in the paper conform, in every respect, with the NeurIPS Code of Ethics \url{https://neurips.cc/public/EthicsGuidelines}?
    \item[] Answer: \answerYes{} 
    \item[] Justification: The research conforms with the NeurIPS Code of Ethics. The work does not involve human subjects or sensitive personal data, all datasets and pre-trained force fields used are publicly available and properly cited.

\item {\bf Broader impacts}
    \item[] Question: Does the paper discuss both potential positive societal impacts and negative societal impacts of the work performed?
    \item[] Answer: \answerYes{} 
    \item[] Justification: The paper includes a Broader Impact section in Appendix \ref{app:impact-statement} that describes applications to protein folding and cell state perturbation modeling, and the limited dual-use risk of the framework given its methodological scope. The statement also notes the use of publicly available datasets (Tahoe-100M and the pancreatic $\beta$-cell dataset) under their original licenses, and the use of standard test proteins and publicly available force fields with no human or animal subjects.
    
\item {\bf Safeguards}
    \item[] Question: Does the paper describe safeguards that have been put in place for responsible release of data or models that have a high risk for misuse (e.g., pre-trained language models, image generators, or scraped datasets)?
    \item[] Answer: \answerNA{} 
    \item[] Justification: The paper does not release new pre-trained foundation models, datasets, or generators that pose a high risk of misuse. The released code provides a simulation framework for cellular and molecular dynamics that operates on publicly available data and standard force fields.

\item {\bf Licenses for existing assets}
    \item[] Question: Are the creators or original owners of assets (e.g., code, data, models), used in the paper, properly credited and are the license and terms of use explicitly mentioned and properly respected?
    \item[] Answer: \answerYes{} 
    \item[] Justification: All existing assets used are properly credited via citations to the original works, including the Tahoe-100M dataset~\citep{tahoe-100m} used in App~\ref{app:cell-experiment}, the pancreatic $\beta$-cell dataset cited in App~\ref{app:pancreatic-cell}, the AMBER99SB-ILDN and ff14SBonlysc force fields with the gbn2 solvation model used in App~\ref{app:tps-experiment}, the OpenMM molecular dynamics engine, the VVVR integrator, and the TPS-DPS baseline whose hyperparameters and protocol we follow.

\item {\bf New assets}
    \item[] Question: Are new assets introduced in the paper well documented and is the documentation provided alongside the assets?
    \item[] Answer: \answerYes{} 
    \item[] Justification: The new asset introduced is the EntangledSBM framework, which is documented through the methodological descriptions in Sec~\ref{sec:Problem Formulation} and~\ref{sec:EntangledSBM}, the pseudo-code in Algorithms~\ref{alg:Framework},~\ref{alg:EntangledSBM Training}, and~\ref{alg:EntangledSBM Inference} (App~\ref{app:Algorithms}), the model architecture details in App~\ref{app:Model Architecture}, the hyperparameter tables in Tables~\ref{table:cell-hyperparameters} and~\ref{table:tps-hyperparameters}, and the code repository at \href{https://github.com/sophtang/EntangledSBM}{https://github.com/sophtang/EntangledSBM}.

\item {\bf Crowdsourcing and research with human subjects}
    \item[] Question: For crowdsourcing experiments and research with human subjects, does the paper include the full text of instructions given to participants and screenshots, if applicable, as well as details about compensation (if any)? 
    \item[] Answer: \answerNA{} 
    \item[] Justification: The paper does not involve crowdsourcing or research with human subjects. All experiments use publicly available datasets and in silico simulations.

\item {\bf Institutional review board (IRB) approvals or equivalent for research with human subjects}
    \item[] Question: Does the paper describe potential risks incurred by study participants, whether such risks were disclosed to the subjects, and whether Institutional Review Board (IRB) approvals (or an equivalent approval/review based on the requirements of your country or institution) were obtained?
    \item[] Answer: \answerNA{} 
    \item[] Justification: The paper does not involve human subjects research and therefore no IRB approval is required.

\item {\bf Declaration of LLM usage}
    \item[] Question: Does the paper describe the usage of LLMs if it is an important, original, or non-standard component of the core methods in this research? Note that if the LLM is used only for writing, editing, or formatting purposes and does \emph{not} impact the core methodology, scientific rigor, or originality of the research, declaration is not required.
    \item[] Answer: \answerNA{} 
    \item[] Justification: LLMs are not used as an important, original, or non-standard component of the core methodology. The framework concerns Schr\"odinger bridge matching for interacting multi-particle dynamics with a Transformer-based bias-force parameterization, and does not employ language models in the methodology.

\end{enumerate}

\end{document}